\documentclass[runningheads]{llncs}
\usepackage{subfiles}

\usepackage{graphicx}
\usepackage{tikz}
\usepackage{comment} 
\usepackage{amsmath,amssymb} % define this before the line numbering.
\usepackage{color}

\usepackage{booktabs}
\usepackage{multirow}
\usepackage{array}
\usepackage{xspace}
\usepackage{ulem}
\usepackage{hyperref}

\makeatletter
\DeclareRobustCommand\onedot{\futurelet\@let@token\@onedot}
\def\@onedot{\ifx\@let@token.\else.\null\fi\xspace}

\renewcommand{\emph}[1]{\textit{#1}}

\def\ie{\emph{i.e}\onedot}

\def\etal{\emph{et al}\onedot}
\makeatother

\usepackage[width=122mm,left=12mm,paperwidth=146mm,height=193mm,top=12mm,paperheight=217mm]{geometry}

\begin{document}
% \linenumbers
\pagestyle{headings}
\mainmatter
\def\ECCVSubNumber{4617}  % Insert your submission number here

\title{Learning Camera-Aware Noise Models} % Replace with your title

\begin{comment}
\titlerunning{ECCV-20 submission ID \ECCVSubNumber} 
\authorrunning{ECCV-20 submission ID \ECCVSubNumber} 
\author{Anonymous ECCV submission}
\institute{Paper ID \ECCVSubNumber}
\end{comment}
%******************

%\begin{comment}
\titlerunning{Learning Camera-Aware Noise Models}
\begin{comment}
\renewcommand{\lastandname}{\unskip,}
\author{Ke-Chi Chang\inst{1, 2}\orcidID{0000-0002-5500-1479} \and
        Ren Wang\inst{1}\orcidID{0000-0001-5168-2337} \and
        Hung-Jin Lin\inst{1}\orcidID{0000-0002-5077-6320} \and
        Yu-Lun Liu \inst{1}\orcidID{0000-0002-7561-6884} \and
        Chia-Ping Chen \inst{1}\orcidID{0000-0002-7022-3061} \and
        Yu-Lin Chang \inst{1}\orcidID{0000-0002-2840-0052} \and
        Hwann-Tzong Chen \inst{2}\orcidID{0000-0003-2806-7090}
}
\end{comment}
\author{Ke-Chi Chang\inst{1, 2} \and
        Ren Wang\inst{1} \and
        Hung-Jin Lin\inst{1} \and 
        Yu-Lun Liu \inst{1} \and
        Chia-Ping Chen \inst{1} \and 
        Yu-Lin Chang \inst{1} \and
        Hwann-Tzong Chen \inst{2}
}
\authorrunning{K.-C. Chang \etal}
\institute{ MediaTek Inc., Hsinchu, Taiwan \and
            National Tsing Hua University, Hsinchu, Taiwan 
            % \url{http://www.springer.com/gp/computer-science/lncs} 
}

%\end{comment}
%******************
\maketitle

\newcommand{\cleanImg}[2]{\ensuremath{{\mathbf{I}_C}^#1_#2}}
\newcommand{\noisyImg}[2]{\ensuremath{{\mathbf{I}_N}^#1_#2}}
\newcommand{\realNoise}[2]{\ensuremath{\mathbf{n}}^#1_#2}

\newcommand{\cpchen}[1]{{\color{red}(cpchen: #1)}}
\newcommand{\yulunliu}[1]{{\color{red}\textbf{yulunliu: }#1}}
\newcommand{\salas}[1]{{\color{blue}\emph{salas: }#1}}
\definecolor{MTKGold}{RGB}{241, 154, 33}
\newcommand{\ren}[1]{{\color{MTKGold}\textbf{ren: }#1}}
\definecolor{NTHUPurple}{RGB}{130, 13, 214}
\newcommand{\arc}[1]{{\color{NTHUPurple}\emph{arc: }#1}}
\definecolor{DoraemonBlue}{RGB}{26,225,248}
\newcommand{\dora}[1]{{\color{DoraemonBlue}\emph{dora: }#1}}
\newcommand\figwidth{0.12}
\newcommand\figwidthFM{0.15}
\newcommand\figwidthnr{0.13}
\newcommand\figwidthtSNE{0.48}

\begin{abstract}
Modeling imaging sensor noise is a fundamental problem for image processing and computer vision applications.
While most previous works adopt statistical noise models, real-world noise is far more complicated and beyond what these models can describe.
To tackle this issue, we propose a data-driven approach, where a generative noise model is learned from real-world noise.
The proposed noise model is camera-aware, that is, different noise characteristics of different camera sensors can be learned simultaneously,
and a single learned noise model can generate different noise for different camera sensors.
Experimental results show that our method quantitatively and qualitatively outperforms existing statistical noise models and learning-based methods.
The source code and more results are available at \url{https://arcchang1236.github.io/CA-NoiseGAN/}.

\end{abstract}

\section{Introduction}

Modeling imaging sensor noise is an important task for many image processing and computer vision applications. 
Besides low-level applications such as image denoising~\cite{Unprocessing,CBDNet,DnCNN,zhang2018ffdnet}, many high-level applications, such as detection or recognition~\cite{he2017mask,liu2016ssd,redmon2016you,ren2015faster}, can benefit from a better noise model.

Many existing works assume statistical noise models in their applications.
The most common and simplest one is signal-independent additive white Gaussian noise (AWGN)~\cite{DnCNN}. 
A combination of Poisson and Gaussian noise, containing both signal-dependent and signal-independent noise, is shown to be a better fit for most camera sensors~\cite{foi2008pg,CBDNet}.

However, the behavior of real-world noise is very complicated. 
Different noise can be induced at different stages of an imaging pipeline. 
Real-world noise includes but is not limited to photon noise, read noise, fixed-pattern noise, dark current noise, row/column noise, and quantization noise. 
Thus simple statistical noise models can not well describe the behavior of real-world noise. 

Recently, several learning-based noise models are proposed to better represent the complexity of real-world noise in a data-driven manner~\cite{NoiseFlow,GCBD,GRDN}.
In this paper, we propose a learning-based generative model for signal-dependent synthetic noise.
The synthetic noise generated by our model is perceptually more realistic than existing statistical models and other learning-based methods.
When used to train a denoising network, better denoising quality can also be achieved.

Moreover, the proposed method is camera-aware. 
Different noise characteristics of different camera sensors can be learned simultaneously by a single generative noise model. 
Then this learned noise model can generate different synthetic noise for different camera sensors respectively. 

Our main contributions are summarized as follows:
\begin{itemize}
    \item propose a learning-based generative model for camera sensor noise
    \item achieve camera awareness by leveraging camera-specific Poisson-Gaussian noise and a camera characteristics encoding network
    \item design a novel feature matching loss for signal-dependent patterns, which leads to significant improvement of visual quality
    \item outperform state-of-the-art noise modeling methods and improve image denoising performance
\end{itemize}

\section{Related Work} \label{sec:related}

Image denoising is one of the most important applications and benchmarks in noise modeling. 
Similar to the recent success of deep learning in many vision tasks, deep neural networks also dominate recent advances of image denoising. 

DnCNN~\cite{DnCNN} shows that a residual neural network can perform blind denoising well and obtains better results than previous methods on additive white Gaussian noise (AWGN). 
However, a recent denoising benchmark DND~\cite{DND}, consisting of real photographs, found that the classic BM3D method~\cite{BM3D} outperforms DnCNN on real-world noise instead. 
The main reason is that real-world noise is more complicated than AWGN, and DnCNN failed to generalize to real-world noise because it was trained only with AWGN.

Instead of AWGN, CBDNet~\cite{CBDNet} and Brooks \etal~\cite{Unprocessing} adopt Poisson-Gaussian noise and demonstrate significant improvement on the DND benchmark. 
Actually, they adopt an approximated version of Poisson-Gaussian noise by a heteroscedastic Gaussian distribution:
\begin{align} \label{eq:pg}
    n \sim \mathcal{N}(0, \delta_{\mathrm{shot}} I + \delta_{\mathrm{read}}) \,,
\end{align}
where $n$ is the noise sampling, $I$ is the intensity of a noise-free image, and $\delta_{\mathrm{shot}}$ and $\delta_{\mathrm{read}}$ denote the Poisson and Gaussian components, respectively. 
Moreover, $\delta_{\mathrm{shot}}$ and $\delta_{\mathrm{read}}$ for a specific camera sensor can be obtained via a calibration process~\cite{NLF}.
The physical meaning of these two components corresponds to the signal-dependent and signal-independent noise of a specific camera sensor.

Recently, several learning-based noise modeling approaches have been proposed~\cite{NoiseFlow,GCBD,GRDN}. 
GCBD~\cite{GCBD} is the first GAN-based noise modeling method. 
Its generative noise model, however, takes only a random vector as input but does not take the intensity of the clean image into account. 
That means the generated noise is not signal-dependent.
Different characteristics between different camera sensors are not considered either.
The synthetic noise is learned and imposed on sRGB images, rather than the raw images.
These are the reasons why GCBD didn't deliver promising denoising performance on the DND benchmark~\cite{DND}.

GRDN~\cite{GRDN} is another GAN-based noise modeling method. 
Their model was trained with paired data of clean images and real noisy images of smartphone cameras, provided by NTIRE 2019 Real Image Denoising Challenge~\cite{NTIRE2019}, which is a subset of the SIDD benchmark~\cite{SIDD}.
In addition to a random seed, the input of the generative noise model also contained many conditioning signals: the noise-free image, an identifier indicating the camera sensor, ISO level, and shutter speed. 
Although GRDN can generate signal-dependent and camera-aware noise, the denoising network trained with this generative noise model only improved slightly.
The potential reasons are two-fold: synthetic noise was learned and imposed on sRGB images, not raw images; a plain camera identifier is too simple to represent noise characteristics of different camera sensors.

Noise Flow~\cite{NoiseFlow} applied a flow-based generative model that maximizes the likelihood of real noise on raw images, and then exactly evaluated the noise modeling performance qualitatively and quantitatively. 
To do so, the authors proposed using Kullback-Leibler divergence and negative log-likelihood as the evaluation metrics. 
Both training and evaluation were conducted on SIDD~\cite{SIDD}. 
To our knowledge, Noise Flow is the first deep learning-based method that demonstrates significant improvement in both noise modeling and image denoising capabilities. 
However, they also fed only a camera identifier into a gain layer to represent complex noise characteristics of different camera sensors.

\begin{figure}[ht!]
    \footnotesize
    \centering
    \includegraphics[width=\linewidth]{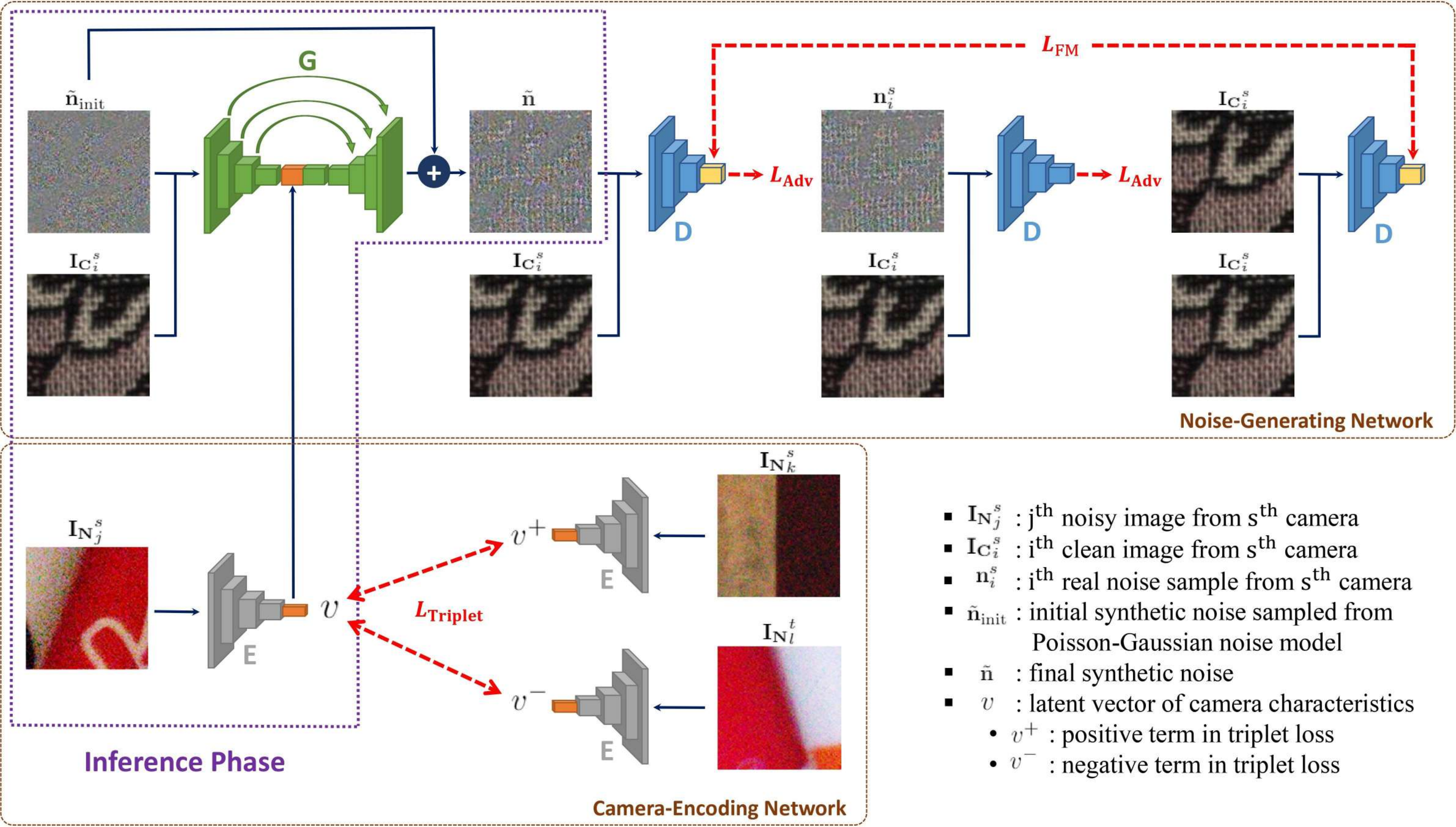}
    \caption{\textbf{An overview of our noise-modeling framework.} The proposed architecture comprises two sub-networks: the Noise-Generating Network and the Camera-Encoding Network. First, a clean image \cleanImg{s}{i} and the initial synthetic noise $\tilde{\mathbf{n}}_\mathrm{init}$ sampled from Poisson-Gaussian noise model are fed into the generator $G$. In addition, a latent vector $\emph{\textbf{v}}$ provided by the camera encoder $E$, which represents the camera characteristics, is concatenated with the features of the middle layers of $G$. Eventually, the final synthetic noise ${\tilde{\mathbf{n}}}$ is generated by $G$. To jointly train $G$ and $E$, a discriminator $D$ is introduced for the adversarial loss $L_\mathrm{Adv}$ and the feature matching loss $L_\mathrm{FM}$. Moreover, a triplet loss $L_\mathrm{Triplet}$ is proposed to let the latent space of $\emph{\textbf{v}}$ be more reliable}
    \label{fig:net}
\end{figure}

\section{Proposed Method} \label{s:method}

Different from most existing works, the proposed learning-based approach aims to model noise characteristics for each camera sensor. 
Fig.~\ref{fig:net} shows an overview of our framework, which comprises two parts: the Noise-Generating Network and the Camera-Encoding Network. 
%First, we illustrate the Noise-Generating Network in section~\ref{subs:noise-generating-network}.
The Noise-Generating Network, introduced in Sec.~\ref{subs:noise-generating-network}, learns to generate synthetic noise according to the content of a clean input image and the characteristics of a target camera. 
The target camera characteristics are extracted via the Camera-Encoding Network from noisy images captured by that target camera, which is illustrated in Sec.~\ref{subs:camera-encoding-network}.
Finally, Sec.~\ref{subs:loss} shows how to train these two networks in an end-to-end scheme.

\subsection{Noise-Generating Network} \label{subs:noise-generating-network}

As depicted in the upper part of Fig.~\ref{fig:net}, a clean image \cleanImg{s}{i} from the $s^\mathrm{th}$ camera and the initial synthetic noise $\tilde{\mathbf{n}}_\mathrm{init}$ are fed into a noise generator $G$ and then transformed into various feature representations through convolutional layers. At the last layer, the network produces a residual image $R(\tilde{\mathbf{n}}_\mathrm{init} | \cleanImg{s}{i})$ that approximates the difference between real noise $\mathbf{n} \sim \mathbb{P}_{r}$ and $\tilde{\mathbf{n}}_\mathrm{init}$, where $\mathbb{P}_{r}$ indicates the real noise distribution. Ideally, we can generate realistic synthetic noise $\tilde{\mathbf{n}} \approx \mathbf{n}$ from the estimated residual image as
\begin{align} \label{eq:g_output}
    \tilde{\mathbf{n}} = G(\tilde{\mathbf{n}}_\mathrm{init} | \cleanImg{s}{i}) = \tilde{\mathbf{n}}_\mathrm{init} + R(\tilde{\mathbf{n}}_\mathrm{init} | \cleanImg{s}{i})\,.
\end{align}

To achieve this objective, we adopt adversarial learning for making the generated noise distribution $\mathbb{P}_{g}$ fit $\mathbb{P}_{r}$ as closely as possible. A discriminator $D$ is used to measure the distance between distributions by distinguishing real samples from fake ones, such that $G$ can minimize the distance through an adversarial loss $L_\mathrm{Adv}$. Therefore, we need to collect pairs of clean images and real noise $(\cleanImg{s}{i}, \realNoise{s}{i})$ as the real samples. 

A real noise sample $\realNoise{s}{i}$ can be acquired by subtracting $\cleanImg{s}{i}$ from the corresponding noisy image $\noisyImg{s}{i}$, \ie, $\realNoise{s}{i} = \noisyImg{s}{i} - \cleanImg{s}{i}$. Note that a clean image could have many corresponding noisy images because noisy images can be captured at different ISOs to cover a wide range of noise levels. For simplicity, we let $i$ denote not only the scene but also the shooting settings of a noisy image.

In addition to measuring the distance in adversarial learning, the discriminator $D$ also plays another role in our framework. It is observed that some signal-dependent patterns like spots or stripes are common in real noise; hence we propose a feature matching loss $L_\mathrm{FM}$ and treat $D$ as a feature extractor. The feature matching loss forces the generated noise $\tilde{\mathbf{n}}$ and the clean image $\cleanImg{s}{i}$ to share similar high-level features because we assume these signal-dependent patterns should be the most salient traits in clean images.

It is worthwhile to mention that a noise model should be capable of generating a variety of reasonable noise samples for the same input image and noise level. GANs usually take a random vector sampled from Gaussian distribution as the input of the generator to ensure this stochastic property. In most cases, this random vector is not directly relevant to the main task. However, our goal is exactly to generate random noise, which implies that this random vector could be treated as the initial synthetic noise. Moreover, Gaussian distribution can be replaced with a more representative statistical noise model. For this reason, we apply Poisson-Gaussian noise model to the initial synthetic noise $\tilde{\mathbf{n}}_\mathrm{init}$ as in (\ref{eq:pg}):
\begin{align} \label{eq:npg}
    \tilde{\mathbf{n}}_{\text{init}} \sim \mathcal{N}(0, {\delta_\mathrm{shot}}_i^s \cleanImg{s}{i} + {{\delta_\mathrm{read}}_i^s}) \,,
\end{align}
where ${\delta_\mathrm{shot}}_i^s$ and ${\delta_\mathrm{read}}_i^s$ are the Poisson and the Gaussian component for $\noisyImg{s}{i}$, respectively. Note that these two parameters not only describe the preliminary noise model for the $s^\mathrm{th}$ camera but also control the noise level of $\tilde{\mathbf{n}}_\mathrm{init}$ and $\tilde{\mathbf{n}}$.

\subsection{Camera-Encoding Network} \label {subs:camera-encoding-network}

FUNIT~\cite{funit} has shown that encoding the class information is helpful to specify the class domain for an input image. Inspired by their work, we would like to encode the camera characteristics in an effective representation. Since ${\delta_\mathrm{shot}}_i^s$ and ${\delta_\mathrm{read}}_i^s$ are related to the $s^\mathrm{th}$ camera in (\ref{eq:npg}), the generator $G$ is actually aware of the camera characteristics from $\tilde{\mathbf{n}}_\mathrm{init}$. However, this awareness is limited to the assumption of the Poisson-Gaussian noise model. We, therefore, propose a novel Camera-Encoding Network to overcome this problem.

As depicted in the lower part of Fig.~\ref{fig:net}, a noisy image $\noisyImg{s}{j}$ is fed into a camera encoder $E$ and then transformed into a latent vector $\emph{\textbf{v}} = E(\noisyImg{s}{j})$. After that, the latent vector $\emph{\textbf{v}}$ is concatenated with the middle layers of $G$. Thus, the final synthetic noise is rewritten as
\begin{align} \label{eq:g_output2}
    \tilde{\mathbf{n}} = G(\tilde{\mathbf{n}}_\mathrm{init} | \cleanImg{s}{i}, \emph{\textbf{v}}) = \tilde{\mathbf{n}}_\mathrm{init} + R(\tilde{\mathbf{n}}_\mathrm{init} | \cleanImg{s}{i}, \emph{\textbf{v}}) \,.
\end{align}
We consider $\emph{\textbf{v}}$ as a representation for the characteristics of the $s^\mathrm{th}$ camera and expect $G$ can generate more realistic noise with this latent vector.

Aiming at this goal, the camera encoder $E$ must have the ability to extract the core information for each camera, regardless of the content of input images. Therefore, a subtle but important detail here is that we feed the $j^\mathrm{th}$ noisy image rather than the $i^\mathrm{th}$ noisy image into $E$, whereas $G$ takes the $i^\mathrm{th}$ clean image as its input. Specifically, the $j^\mathrm{th}$ noisy image is randomly selected from the data of the $s^\mathrm{th}$ camera. Consequently, $E$ has to provide latent vectors beneficial to the generated noise but ignoring the content of input images.

Additionally, some regularization should be imposed on $\emph{\textbf{v}}$ to make the latent space more reliable. FUNIT calculates the mean over a set of class images to provide a representative class code. Nevertheless, this approach assumes that the latent space consists of hypersphere manifolds. Apart from FUNIT, we use a triplet loss $L_\mathrm{Triplet}$ as the regularization. The triplet loss is used to minimize the intra-camera distances while maximizing the inter-camera distances, which allows the latent space to be more robust to image content. The detailed formulation will be shown in the next section.

One more thing worth clarifying is why the latent vector $\emph{\textbf{v}}$ is extracted from the noisy image $\noisyImg{s}{j}$ rather than the real noise sample $\realNoise{s}{j}$. The reason is out of consideration for making data preparation easier in the inference phase, which is shown as the violet block in Fig.~\ref{fig:net}. Collecting paired data $(\cleanImg{s}{j}, \noisyImg{s}{j})$ to acquire $\realNoise{s}{j}$ is cumbersome and time-consuming in real world. With directly using noisy images to extract latent vectors, there is no need to prepare a large number of paired data during the inference phase.

\subsection{Learning} \label{subs:loss}

To jointly train the aforementioned networks, we have briefly introduced three loss functions: 1) the adversarial loss $L_\mathrm{Adv}$, 2) the feature matching loss $L_\mathrm{FM}$, and 3) the triplet loss $L_\mathrm{Triplet}$. In this section, we describe the formulations for these loss functions in detail.

\subsubsection{Adversarial Loss.}
GANs are well-known for reducing the divergence between the generated data distribution and real data distribution in the high-dimensional image space. However, there are several GAN frameworks for achieving this goal. Among these frameworks, we choose WGAN-GP~\cite{WGAN-GP} to calculate the adversarial loss $L_\mathrm{Adv}$, which minimizes Wasserstein distance for stabilizing the training. The $L_\mathrm{Adv}$ is thus defined as
\begin{align} \label{eq:advG}
    L_\mathrm{Adv} = - \underset{\tilde{\mathbf{n}} \sim \mathbb{P}_{g}} {\mathbb{E}}[D(\tilde{\mathbf{n}} | \mathbf{I}_C)]\,,
\end{align}
where $D$ scores the realness of the generated noise. In more depth, scores are given at the scale of patches rather than whole images because we apply a PatchGAN~\cite{PatchGAN} architecture to $D$. The advantage of using this architecture is that it prefers to capture high-frequency information, which is associated with the characteristics of noise.

On the other hand, the discriminator $D$ is trained by
\begin{align} \label{eq:advD}
    L_{D} = \underset{\tilde{\mathbf{n}} \sim \mathbb{P}_{g}} {\mathbb{E}}[D(\tilde{\mathbf{n}} | \mathbf{I}_C)] - \underset{\mathbf{n} \sim \mathbb{P}_{r}}{\mathbb{E}}[D(\mathbf{n} | \mathbf{I}_C)] + \lambda_\mathrm{gp} \underset{\hat{\mathbf{n}} \sim \mathbb{P}_{\hat{\mathbf{n}}}} {\mathbb{E}}[(\|\nabla_{\hat{\mathbf{n}}}D(\hat{\mathbf{n}} | \mathbf{I}_C) \|_{2}-1)^{2}]\,,
\end{align}
where $\lambda_\mathrm{gp}$ is the weight of gradient penalty, and $\mathbb{P}_{\hat{\mathbf{n}}}$ is the distribution sampling uniformly along straight lines between paired points sampled from $\mathbb{P}_{g}$ and $\mathbb{P}_{r}$.

\subsubsection{Feature Matching Loss.}

In order to regularize the training for GANs, some works \cite{funit,FMLoss} apply the feature matching loss and extract features through the discriminator networks. Following these works, we propose a feature matching loss $L_\mathrm{FM}$ to encourage $G$ to generate signal-dependent patterns in synthetic noise. The $L_\mathrm{FM}$ is then calculated as
\begin{align} \label{eq:fm}
    L_\mathrm{FM} = \underset{\tilde{\mathbf{n}} \sim \mathbb{P}_{g}} {\mathbb{E}}[\| D_{f}(\tilde{\mathbf{n}} | \mathbf{I}_C) - D_{f}(\mathbf{I}_C | \mathbf{I}_C) \|_{1} ]\,,
\end{align}
where $D_{f}$ denotes the feature extractor constructed by removing the last layer from $D$. Note that $D_f$ is not optimized by $L_\mathrm{FM}$.
\subsubsection{Triplet Loss.}

The triplet loss was first proposed to illustrate the triplet relation in embedding space by~\cite{TripletLoss}. We use the triplet loss to let the latent vector $\emph{\textbf{v}} = E(\noisyImg{s}{j})$ be more robust to the content of noisy image. Here we define the positive term $\emph{\textbf{v}}^{+}$ as the latent vector also extracted from the $s^\mathrm{th}$ camera, and the negative term $\emph{\textbf{v}}^{-}$ is from a different camera on the contrary. In particular, $\emph{\textbf{v}}^{+}$ and $\emph{\textbf{v}}^{-}$ are obtained by encoding the randomly selected noisy images $\noisyImg{s}{k}$ and $\noisyImg{t}{l}$, respectively. Note that $\noisyImg{s}{k}$ is not restricted to any shooting setting, which means the images captured with different shooting settings of the same camera are treated as positive samples. The objective is to minimize the intra-camera distances while maximizing the inter-camera distances. The triplet loss $L_{\text{Triplet}}$ is thus given by
\begin{align} \label{eq:camera}
    L_\mathrm{Triplet} = \underset{\emph{\textbf{v}}, \emph{\textbf{v}}^{+}, \emph{\textbf{v}}^{-} \sim \mathbb{P}_{e}}
    {\mathbb{E}} \left[\max(0, \left\lVert \emph{\textbf{v}} - \emph{\textbf{v}}^{+} \right\rVert_2 - \left\lVert \emph{\textbf{v}} - \emph{\textbf{v}}^{-} \right\rVert_2 + \alpha) \right] \,,
\end{align}
where $\mathbb{P}_e$ is the latent space distribution and $\alpha$ is the margin between positive and negative pairs.

\subsubsection{Full Loss.}
The full objective of the generator $G$ is combined as
\begin{align} \label{eq:fullG}
    L_\mathrm{G} = L_\mathrm{Adv} + \lambda_\mathrm{FM} L_\mathrm{FM} + \lambda_\mathrm{Triplet} L_\mathrm{Triplet} \,,
\end{align}
where $\lambda_\mathrm{FM}$ and $\lambda_\mathrm{Triplet}$ control the relative importance for each loss term.

\section{Experimental Results} \label{s:results}
In this section, we first describe our experiment settings and the implementation details.
Then, Sec.~\ref{subs:results:nm} shows the quantitative and qualitative results.
Sec.~\ref{subs:results:ablation} presents extensive ablation studies to justify our design choices. The effectiveness and robustness of the Camera-Encoding Network are evaluated in Sec.~\ref{subs:results:camera:encoding}.

\subsubsection{Dataset.}
We train and evaluate our method on Smartphone Image Denoising Dataset (SIDD)~\cite{SIDD},
which consists of approximately 24{,}000 pairs of real noisy-clean images.
The images are captured by five different smartphone cameras: Google Pixel, iPhone 7, Samsung Galaxy S6 Edge, Motorola Nexus 6, and LG G4. 
These images are taken in ten different scenes and under a variety of lighting conditions and ISOs.
SIDD is currently the most abundant dataset available for real noisy and clean image pairs.
\subsubsection{Implementation Details.}

We apply Bayer preserving augmentation~\cite{BayerAug} to all SIDD images, including random cropping and horizontal flipping.
At both training and testing phases, the images are cropped into $64\times64$ patches.
Totally 650{,}000 pairs of noisy-clean patches are generated. 
Then we randomly select 500{,}000 pairs as the training set and 150{,}000 pairs as the test set.
The scenes in the training set and test set are mutually exclusive to prevent overfitting. Specifically, the scene indices of the test set are 001, 002 and 008, and the remaining indices are used for the training set.

To synthesize the initial synthetic noise $\tilde{\mathbf{n}}_\mathrm{init}$, we set the Poisson component ${\delta_{\mathrm{shot}}}_i^s$ and Gaussian component ${\delta_{\mathrm{read}}}_i^s$ in (\ref{eq:npg}) to the values provided by SIDD, which are estimated using the method proposed by~\cite{NLF}.
The weight of gradient penalty of $L_D$ in (\ref{eq:advD}) is set to $\lambda_\mathrm{gp} = 10$, and the margin of $L_{\mathrm{Triplet}}$ in (\ref{eq:camera}) is set to $\alpha = 0.2$.
The loss weights of $L_G$ in (\ref{eq:fullG}) are set to $\lambda_\mathrm{FM} = 1$ and $\lambda_\mathrm{Triplet} = 0.5$.  

We use the Adam optimizer~\cite{kingma2014adam} in all of our experiments, with an initial learning rate of 0.0002, $\beta_{1} = 0.5$, and $\beta_{2} = 0.999$.
Each training batch contains 64 pairs of noisy-clean patches. The generator $G$, discriminator $D$, and camera encoder $E$ are jointly trained to convergence with 300 epochs. It takes about 3 days on a single GeForce GTX 1080 Ti GPU.

All of our experiments are conducted on linear raw images. 
Previous works have shown that many image processing methods perform better in Bayer RAW domain than in sRGB domain~\cite{DND}. 
For noise modeling or image denoising, avoiding non-linear transforms (such as gamma correction) or spatial operations (such as demosaicking) is beneficial because we can prevent noise characteristics from being dramatically changed by these operations.

\subsubsection{Methods in Comparison.}
We compare our method with two mostly-used statistical models: Gaussian noise model and Poisson-Gaussian noise model, and one state-of-the-art learning-based method: Noise Flow~\cite{NoiseFlow}.
\subsection{Quantitative and Qualitative Results} \label{subs:results:nm}
To perform the quantitative comparison, we adopt the Kullback-Leibler divergence ($D_\mathrm{KL}$) as suggested in~\cite{NoiseFlow}.  
Table~\ref{table:kld} shows the average $D_\mathrm{KL}$ between real noise and synthetic noise generated by different noise models.
Our method achieves the smallest average Kullback-Leibler divergence, which means that our method can synthesize more realistic noise than existing methods.

\setlength{\tabcolsep}{4pt}
\begin{table}[tb]
    \footnotesize
    \begin{center}
    \caption{\textbf{Quantitative evaluation of different noise models.} Our proposed noise model yields the best Kullback-Leibler divergence ($D_\mathrm{KL}$). Relative improvements of our method over other baselines are shown in parentheses}
    \label{table:kld}
    \begin{tabular}{c|cccc}
        \toprule
        & Gaussian & Poisson-Gaussian & Noise Flow & Ours \\
        \midrule
        $D_\mathrm{KL}$ & 0.54707 (99.5\%) & 0.01006 (74.7\%) & 0.00912 (72.0\%) & \pmb{0.00159} \\
        \bottomrule
    \end{tabular}
    \end{center}
\end{table}
\setlength{\tabcolsep}{1.4pt}

\begin{figure}[ht!]
    \tiny
    \centering
    \begin{tabular}{ccccccc}
       Clean &  & Gaussian & \makebox[2em][c]{Poisson-Gauss.} &  Noise Flow & Ours & Real Noise \\           
        % \multirow{2}{*}{\raisebox{0.2\normalbaselineskip}[0pt][0pt]{\rotatebox[origin=c]{90}{100-L}}} &
        \includegraphics[width=\figwidth\columnwidth]{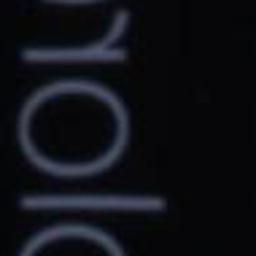} &
        \raisebox{3.0\normalbaselineskip}[0pt][0pt]{\rotatebox[origin=c]{90}{Noisy}} &
        \includegraphics[width=\figwidth\columnwidth]{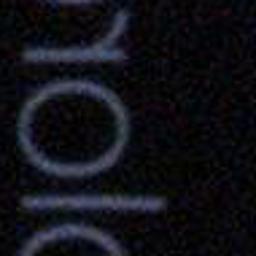} &
        \includegraphics[width=\figwidth\columnwidth]{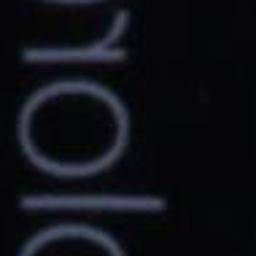} &
        \includegraphics[width=\figwidth\columnwidth]{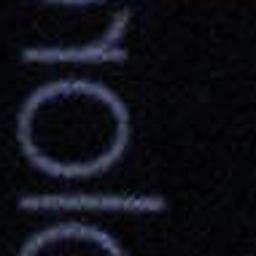} &
        \includegraphics[width=\figwidth\columnwidth]{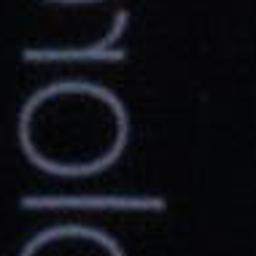} &
        \includegraphics[width=\figwidth\columnwidth]{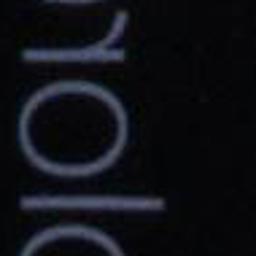}
        \\
        & 
        \raisebox{3.0\normalbaselineskip}[0pt][0pt]{\rotatebox[origin=c]{90}{Noise}} &
        \includegraphics[width=\figwidth\columnwidth]{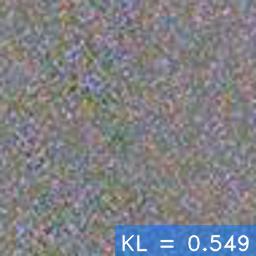} &
        \includegraphics[width=\figwidth\columnwidth]{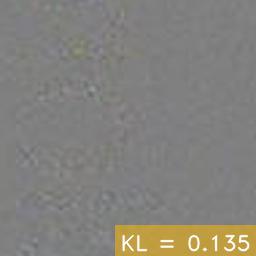} &
        \includegraphics[width=\figwidth\columnwidth]{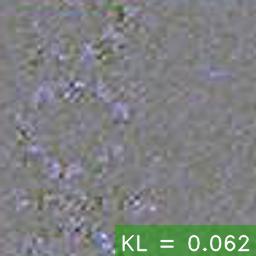} &
        \includegraphics[width=\figwidth\columnwidth]{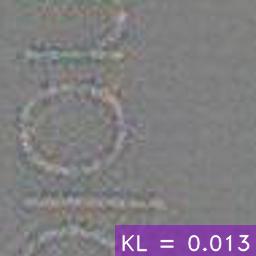} &
        \includegraphics[width=\figwidth\columnwidth]{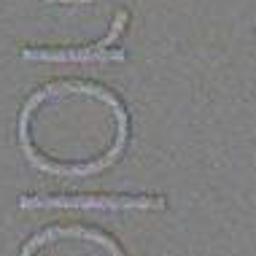}
        \\
        \multicolumn{7}{c}{(a) 100-L} \\
        
        %
        % \multirow{2}{*}{\raisebox{0.2\normalbaselineskip}[0pt][0pt]{\rotatebox[origin=c]{90}{100-N}}} &
        \includegraphics[width=\figwidth\columnwidth]{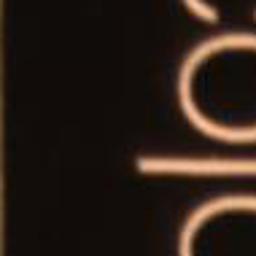} &
        \raisebox{3.0\normalbaselineskip}[0pt][0pt]{\rotatebox[origin=c]{90}{Noisy}} &
        \includegraphics[width=\figwidth\columnwidth]{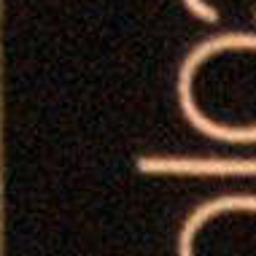} &
        \includegraphics[width=\figwidth\columnwidth]{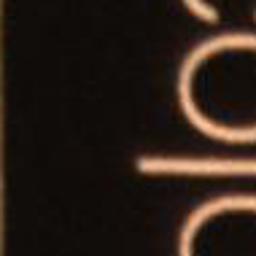} &
        \includegraphics[width=\figwidth\columnwidth]{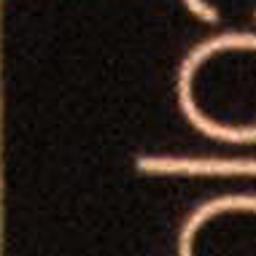} &
        \includegraphics[width=\figwidth\columnwidth]{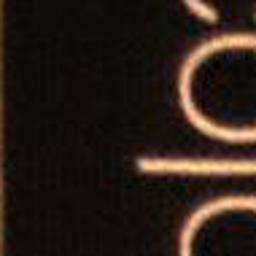} &
        \includegraphics[width=\figwidth\columnwidth]{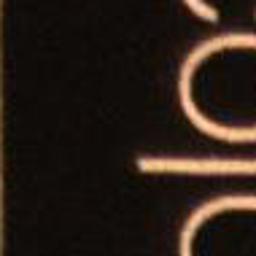} 
        \\
        & 
        \raisebox{3.0\normalbaselineskip}[0pt][0pt]{\rotatebox[origin=c]{90}{Noise}} &
        \includegraphics[width=\figwidth\columnwidth]{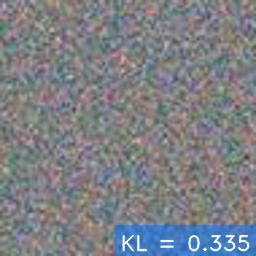} &
        \includegraphics[width=\figwidth\columnwidth]{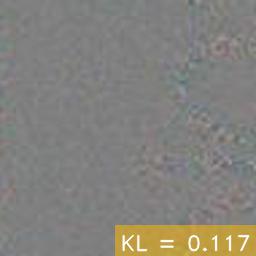} &
        \includegraphics[width=\figwidth\columnwidth]{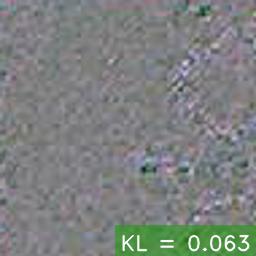} &
        \includegraphics[width=\figwidth\columnwidth]{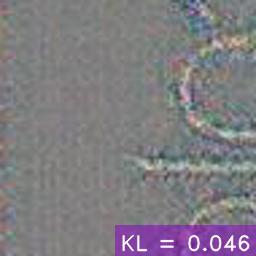} &
        \includegraphics[width=\figwidth\columnwidth]{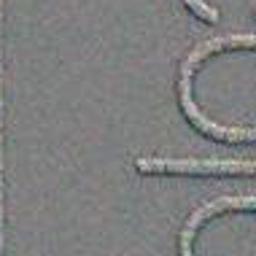}
        \\
        \multicolumn{7}{c}{(b) 100-N}
        \\
        
        %
        % \multirow{2}{*}{\raisebox{0.2\normalbaselineskip}[0pt][0pt]{\rotatebox[origin=c]{90}{1600-L}}} &
        \includegraphics[width=\figwidth\columnwidth]{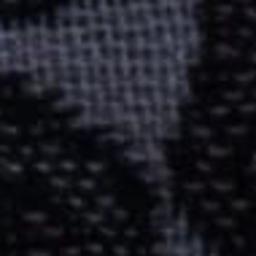} &
        \raisebox{3.0\normalbaselineskip}[0pt][0pt]{\rotatebox[origin=c]{90}{Noisy}} &
        \includegraphics[width=\figwidth\columnwidth]{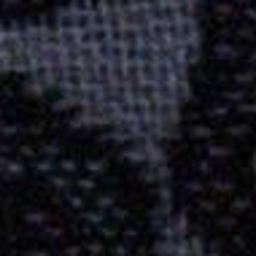} &
        \includegraphics[width=\figwidth\columnwidth]{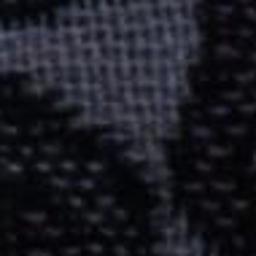} &
        \includegraphics[width=\figwidth\columnwidth]{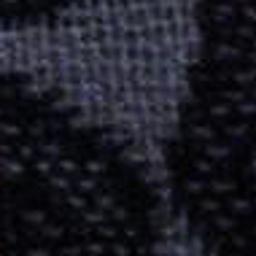} &
        \includegraphics[width=\figwidth\columnwidth]{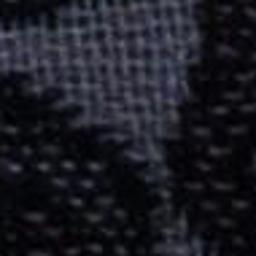} &
        \includegraphics[width=\figwidth\columnwidth]{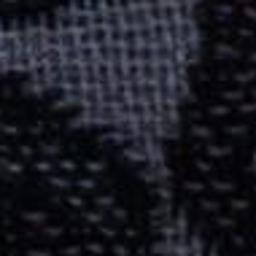} 
        \\
        & 
        \raisebox{3.0\normalbaselineskip}[0pt][0pt]{\rotatebox[origin=c]{90}{Noise}} &
        \includegraphics[width=\figwidth\columnwidth]{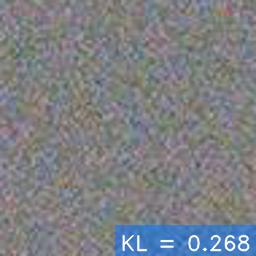} &
        \includegraphics[width=\figwidth\columnwidth]{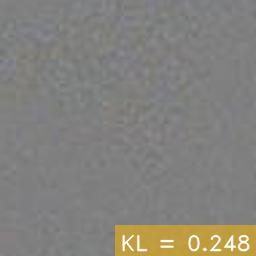} &
        \includegraphics[width=\figwidth\columnwidth]{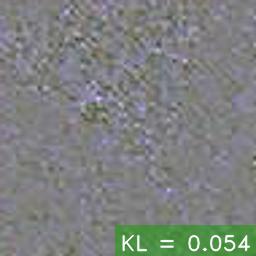} &
        \includegraphics[width=\figwidth\columnwidth]{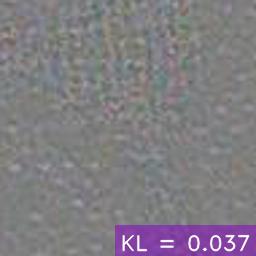} &
        \includegraphics[width=\figwidth\columnwidth]{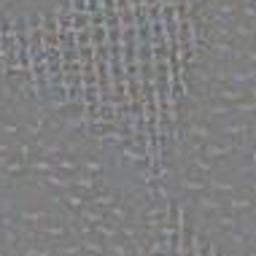}
        \\
        \multicolumn{7}{c}{(c) 100-N}
        \\
        
        %
        % \multirow{2}{*}{\raisebox{0.2\normalbaselineskip}[0pt][0pt]{\rotatebox[origin=c]{90}{3200-N}}} &
        \includegraphics[width=\figwidth\columnwidth]{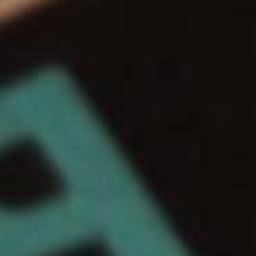} &
        \raisebox{3.0\normalbaselineskip}[0pt][0pt]{\rotatebox[origin=c]{90}{Noisy}} &
        \includegraphics[width=\figwidth\columnwidth]{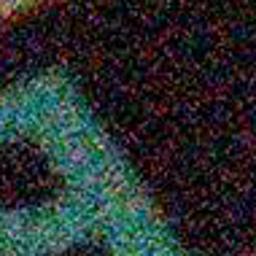} &
        \includegraphics[width=\figwidth\columnwidth]{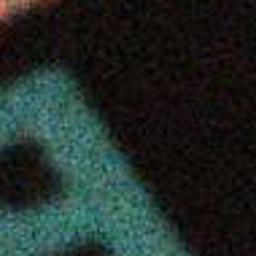} &
        \includegraphics[width=\figwidth\columnwidth]{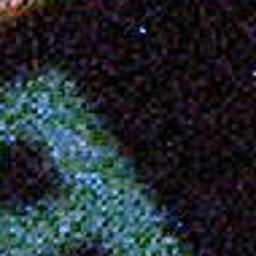} &
        \includegraphics[width=\figwidth\columnwidth]{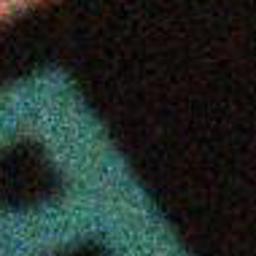} &
        \includegraphics[width=\figwidth\columnwidth]{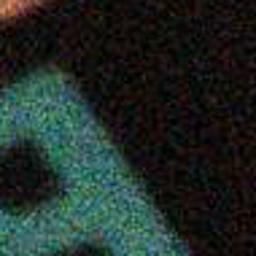} 
        \\
        & 
        \raisebox{3.0\normalbaselineskip}[0pt][0pt]{\rotatebox[origin=c]{90}{Noise}} &
        \includegraphics[width=\figwidth\columnwidth]{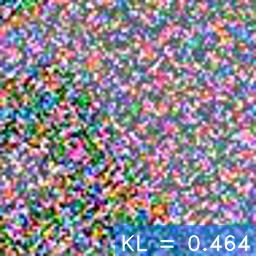} &
        \includegraphics[width=\figwidth\columnwidth]{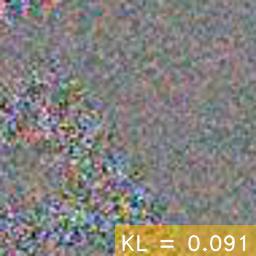} &
        \includegraphics[width=\figwidth\columnwidth]{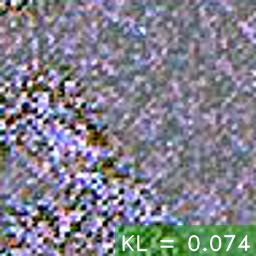} &
        \includegraphics[width=\figwidth\columnwidth]{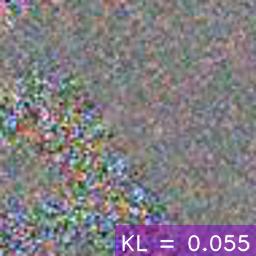} &
        \includegraphics[width=\figwidth\columnwidth]{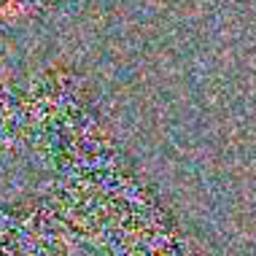}
        \\
        \multicolumn{7}{c}{(d) 1600-L}
        \\
        
        %
        % \multirow{2}{*}{\raisebox{0.2\normalbaselineskip}[0pt][0pt]{\rotatebox[origin=c]{90}{100-N}}} &
        \includegraphics[width=\figwidth\columnwidth]{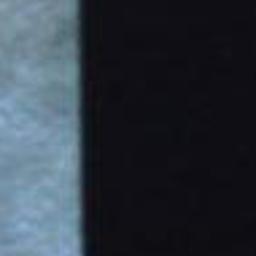} &
        \raisebox{3.0\normalbaselineskip}[0pt][0pt]{\rotatebox[origin=c]{90}{Noisy}} &
        \includegraphics[width=\figwidth\columnwidth]{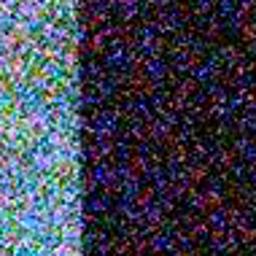} &
        \includegraphics[width=\figwidth\columnwidth]{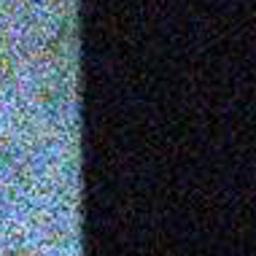} &
        \includegraphics[width=\figwidth\columnwidth]{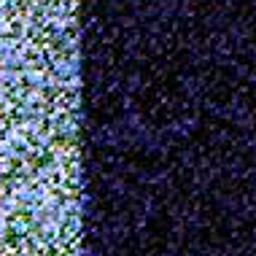} &
        \includegraphics[width=\figwidth\columnwidth]{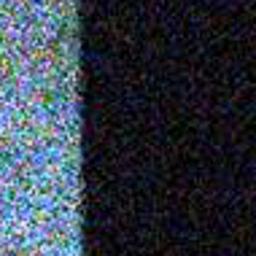} &
        \includegraphics[width=\figwidth\columnwidth]{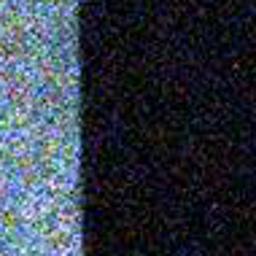} 
        \\
        & 
        \raisebox{3.0\normalbaselineskip}[0pt][0pt]{\rotatebox[origin=c]{90}{Noise}} &
        \includegraphics[width=\figwidth\columnwidth]{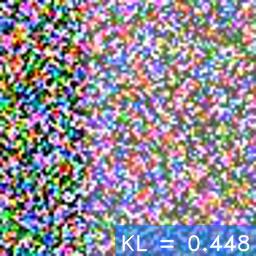} &
        \includegraphics[width=\figwidth\columnwidth]{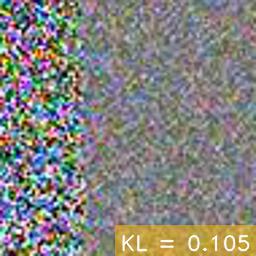} &
        \includegraphics[width=\figwidth\columnwidth]{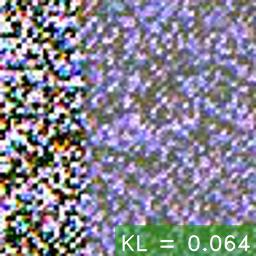} &
        \includegraphics[width=\figwidth\columnwidth]{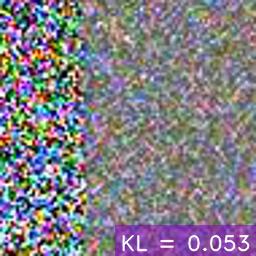} &
        \includegraphics[width=\figwidth\columnwidth]{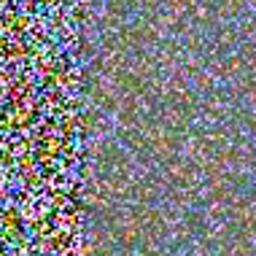}
        \\
        \multicolumn{7}{c}{(e) 3200-N}
        \\

    \end{tabular}
    \caption{\textbf{Visualization of noise models.}
    The synthetic noise samples of several noise modeling methods on different clean images with different ISO/lighting conditions are illustrated
    Quantitatively, the proposed method outperforms others in terms of $D_\mathrm{KL}$ measurement with real noise distribution.
    Furthermore, ours have many clear structures that fit the texture of clean images and real noise.
    Note that the noise value is scaled up for better visualization purpose
    % \yulunliu{Need to mention how we visualize the noise. (((Noisy - Clean)*scale)+1.0)/2.0?}
    }
    \label{fig:solvepg}
\end{figure}

Fig.~\ref{fig:solvepg} shows the synthetic noise generated in linear RAW domain by all noise models and then processed by the camera pipeline toolbox provided by SIDD~\cite{SIDD}.
Each two consecutive rows represent an image sample for different ISOs (indicated by a number) and different lighting conditions (L and N denote low and normal lighting conditions respectively).
Our method can indeed generate synthetic noise that is more realistic and perceptually closer to the real noise.

\def\fo{02_0100_693}
\def\foo{08_0100_106}
\def\fooo{180_0100_394}
\def\foooo{52_1600_701}
\def\fooooo{14_3200_631}
\setlength{\tabcolsep}{1.4pt}

\subsection{Ablation Studies} \label{subs:results:ablation}
In this section, we perform ablation studies to investigate how each component contributes to our method, including the feature matching loss $L_\mathrm{FM}$, the Camera-Encoding Network $E$, the triplet loss $L_\mathrm{Triplet}$, and the initial synthetic noise $\tilde{\mathbf{n}}_\mathrm{init}$. 
The results are shown in Table~\ref{table:ablation} and~\ref{table:ablation_n_init}. 

\setlength{\tabcolsep}{4pt}
\begin{table}[tb]
    \footnotesize
    \begin{center}
    \caption{\textbf{Ablation study of our model.} $L_\mathrm{adv}$: the adversarial loss, $L_\mathrm{FM}$: the feature matching loss, $E$: the Camera-Encoding Network, $L_\mathrm{Triplet}$: the triplet loss.
    The Kullback-Leibler divergence $D_\mathrm{KL}$ is measured in different settings}
    \label{table:ablation}
    \begin{tabular}{l|cccc}
        \toprule
        $L_\mathrm{Adv}$ & $\surd$ & $\surd$ & $\surd$ & $\surd$ \\
        $L_\mathrm{FM}$ & & $\surd$ & $\surd$ & $\surd$ \\
        $E$ & & & $\surd$ & $\surd$ \\
        $L_\mathrm{Triplet}$ & & & & $\surd$ \\
        \midrule
        $D_\mathrm{KL}$ & 0.01445 & 0.01374 & 0.01412 & \pmb{0.00159} \\
        \bottomrule
    \end{tabular}
    \end{center}
\end{table}

\subsubsection{Feature Matching Loss $L_\mathrm{FM}$.}
\begin{figure}[ht!]
    \footnotesize
    \centering
    \footnotesize
    \renewcommand{\tabcolsep}{1pt} % adjust horizontal space
    \renewcommand{\arraystretch}{1} % adjust vertical space
    \begin{tabular}{cccc}
        w/o $L_\mathrm{FM}$ & with $L_\mathrm{FM}$ & Real Noise & Clean \\
        \includegraphics[width=\figwidthFM\columnwidth]{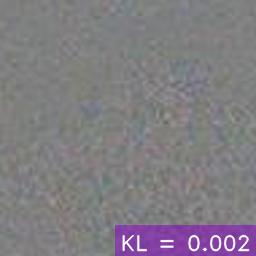} & 
        \includegraphics[width=\figwidthFM\columnwidth]{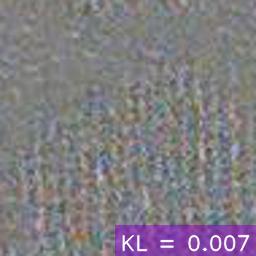} &
        \includegraphics[width=\figwidthFM\columnwidth]{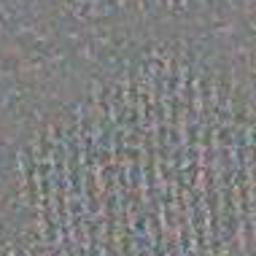} &
        \includegraphics[width=\figwidthFM\columnwidth]{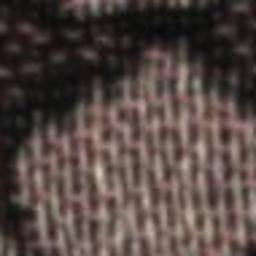} \\
    \end{tabular}
    \caption{\textbf{Visualization of synthetic noise with and without feature matching loss.}
    With the feature matching loss $L_\mathrm{FM}$, the generated noise is highly correlated to the image content.
    Hence, they have more distinct structures resembling the texture of clean images.
    Besides, the $D_\mathrm{KL}$ measurements are slightly improved
    }
    \label{fig:FM}
\end{figure}

Fig.~\ref{fig:FM} shows that the feature matching loss is effective in synthesizing signal-dependent noise patterns and achieving better visual quality.
With the feature matching loss $L_\mathrm{FM}$, the network is more capable of capturing low-frequency signal-dependent patterns.
As shown in Table~\ref{table:ablation}, the Kullback-Leibler divergence can also be improved from 0.01445 to 0.01374.

\subsubsection{Camera-Encoding Network and Triplet Loss.}
The Camera-Encoding Network is designed to represent the noise characteristics of different camera sensors.
However, simply adding a Camera-Encoding Network alone provides no advantage (0.01374 $\rightarrow$ 0.01412), as shown in Table~\ref{table:ablation}.
The triplet loss is essential to learn effective camera-specific latent vectors, and the KL divergence can be significantly reduced from 0.01412 to 0.00159.

\begin{figure}[ht!]
    \footnotesize
    \centering
    \footnotesize
    \renewcommand{\tabcolsep}{1pt} % adjust horizontal space
    \renewcommand{\arraystretch}{1} % adjust vertical space
    \begin{tabular}{cc}
        w/o $L_\mathrm{Triplet}$ & with $L_\mathrm{Triplet}$\\
        \includegraphics[width=\figwidthtSNE\columnwidth]{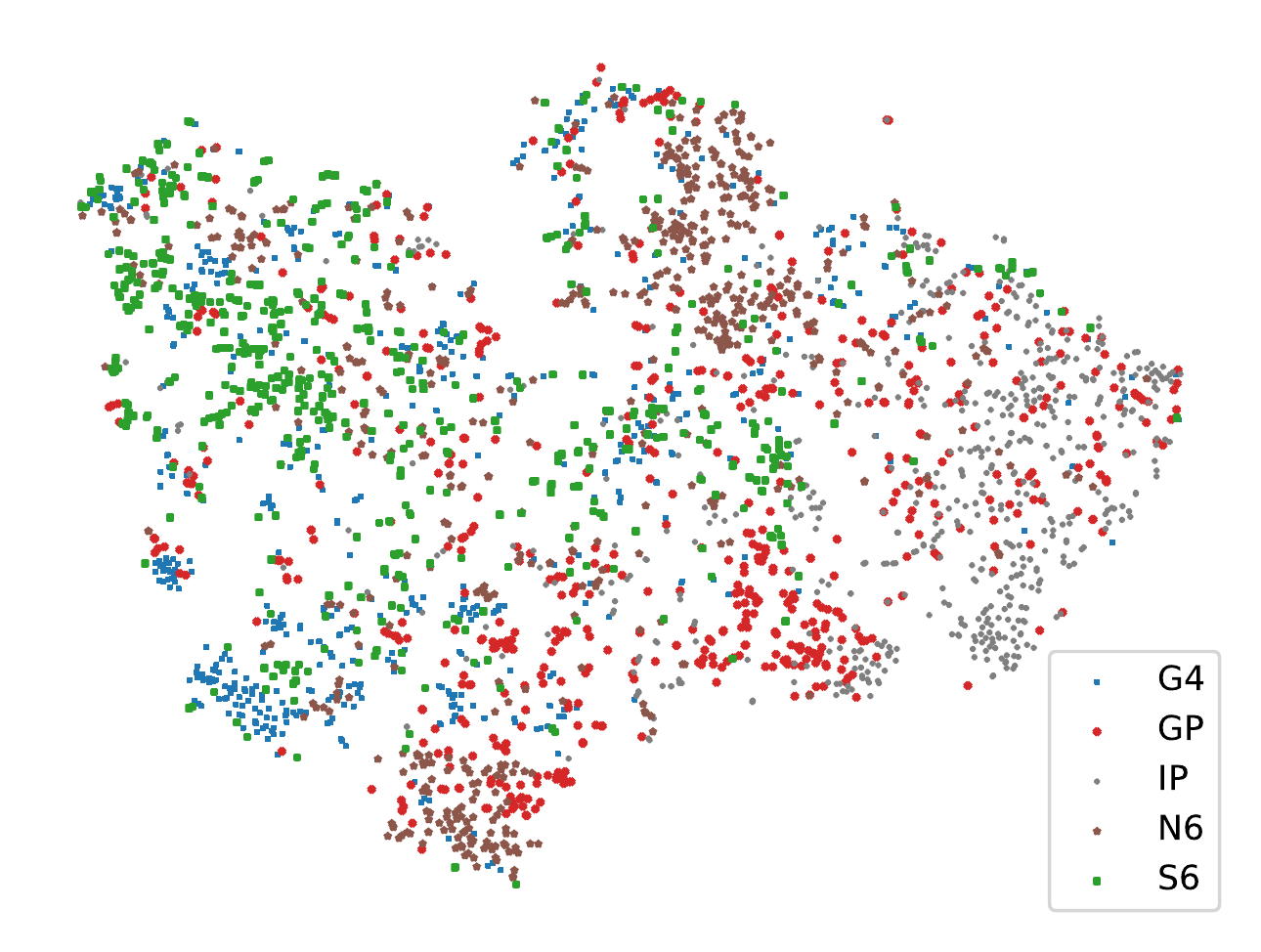} &
        \includegraphics[width=\figwidthtSNE\columnwidth]{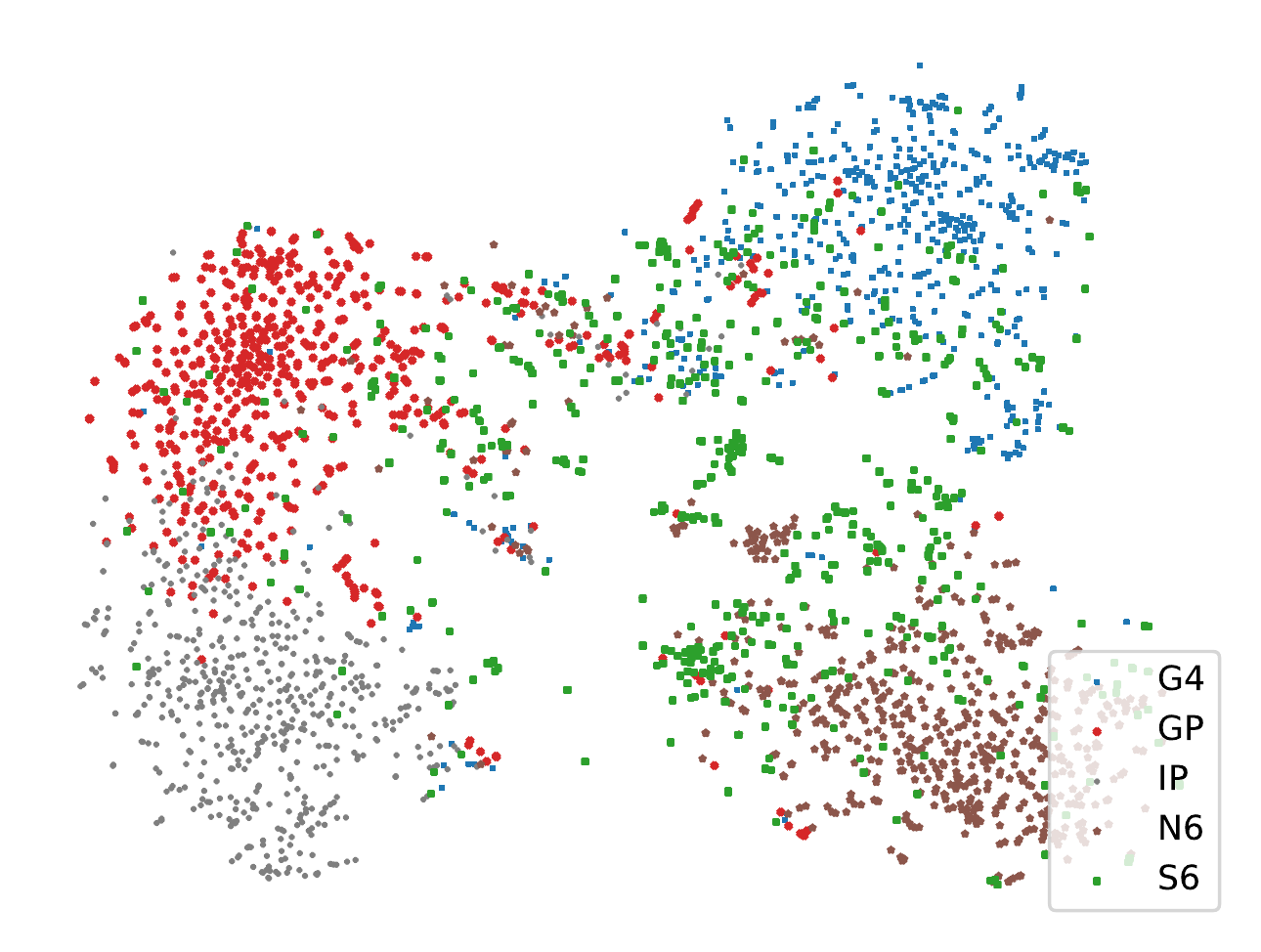} \\
    \end{tabular}
    \caption{\textbf{Ablation study on the distributions of latent vectors from a camera encoder trained with or without $L_\mathrm{Triplet}$.} We project the encoded latent vectors $\emph{\textbf{v}}$ of noisy images from five different cameras with $t$-SNE.
    The Camera-Encoding Network trained with $L_\mathrm{Triplet}$ can effectively group the characteristics of different cameras
    }
    \label{fig:tSNE}
\end{figure}

The camera-specific latent vectors can also be visualized in the $t$-SNE space~\cite{maaten2008visualizing}. 
As shown in Fig.~\ref{fig:tSNE}, the Camera-Encoding Network can effectively extract camera-specific latent vectors from a single noisy image with the triplet loss.

\subsubsection{Initial Synthetic Noise $\tilde{\mathbf{n}}_{\text{init}}$.}
Table~\ref{table:ablation_n_init} shows the average KL divergence when using Gaussian or Poisson-Gaussian noise as the initial noise $\tilde{\mathbf{n}}_{\text{init}}$.

The KL divergence severely degrades from 0.00159 to 0.06265 if we use Gaussian noise instead of Poisson-Gaussian noise. 
This result shows that using a better synthetic noise as initial and predicting a residual to refine it can yield better-synthesized noise.
\setlength{\tabcolsep}{4pt}
\begin{table}[t]
    \footnotesize
    \begin{center}
    \caption{\textbf{Ablation study of the initial synthetic noise $\tilde{\mathbf{n}}_{\text{init}}$} Using Poisson-Gaussian as initial synthetic noise model performs better than using Gaussian}
    \label{table:ablation_n_init}
    \begin{tabular}{l|cc}
        \toprule
        $\tilde{\mathbf{n}}_{\text{init}}$ & Gaussian & Poisson-Gaussian \\
        \midrule
        $D_\mathrm{KL}$ & 0.06265 & \pmb{0.00159} \\
        \bottomrule
    \end{tabular}
    \end{center}
\end{table}

\subsection{Robustness Analysis of the Camera-Encoding Network}\label{subs:results:camera:encoding}
To further verify the behavior and justify the robustness of the Camera-Encoding Network, we design several experiments with different input conditions.

\subsubsection{Comparing Noise for Different Imaging Conditions or Different Cameras.}
Given a clean image $\cleanImg{s}{i}$ and a noisy image $\noisyImg{s}{j}$ also from the $s^{\text{th}}$ camera,
our noise model should generate noise $\tilde{\mathbf{n}}_{A} = G(\tilde{\mathbf{n}}_{\text{init}} | \cleanImg{s}{i}, E(\noisyImg{s}{j}))$. The Kullback-Leibler divergence $D_{\text{KL}}(\tilde{\mathbf{n}}_{A}\|\mathbf{n}_i^s)$ between the generated noise and the corresponding real noise should be very small (0.00159 in Table~\ref{table:ablation_ce}). On the other hand, $D_{\text{KL}}(\tilde{\mathbf{n}}_{A}\|\mathbf{n}_j^s)$ between the generated noise and a non-corresponding real noise should be quite large (0.17921 in Table~\ref{table:ablation_ce}), owing to the different imaging conditions, even though the real noise $\mathbf{n}_j^s$ is from the same $s^{\text{th}}$ camera. 

If the latent vector is extracted by a noisy image of the $t^{\text{th}}$ camera instead of the $s^{\text{th}}$ camera, the generated noise becomes $\tilde{\mathbf{n}}_{B} = G(\tilde{\mathbf{n}}_{\text{init}} | \cleanImg{s}{i}, E(\noisyImg{t}{k}))$. Because the latent vector is from a different camera, we expect that $D_{\text{KL}}(\tilde{\mathbf{n}}_{A}\|\mathbf{n}_i^s) < D_{\text{KL}}(\tilde{\mathbf{n}}_{B}\|\mathbf{n}_i^s)$.  Table~\ref{table:ablation_ce} also verifies these results.

\setlength{\tabcolsep}{4pt}
\begin{table}[t]
    \footnotesize
    \begin{center}
    \caption{\textbf{Analysis of noisy images from different cameras.} The comparison of Kullback-Leibler divergence for different cameras of the noisy image, where $\tilde{\mathbf{n}}_{A} = G(\tilde{\mathbf{n}}_{\text{init}} | \cleanImg{s}{i}, E(\noisyImg{s}{j}))$ and $\tilde{\mathbf{n}}_{B} = G(\tilde{\mathbf{n}}_{\text{init}} | \cleanImg{s}{i}, E(\noisyImg{t}{k}))$
    }
    \label{table:ablation_ce}
    \begin{tabular}{c|ccc}
        \toprule
        & $(\tilde{\mathbf{n}}_{A}\|\mathbf{n}_i^s)$ & $(\tilde{\mathbf{n}}_{A}\|\mathbf{n}_j^s)$ & $(\tilde{\mathbf{n}}_{B}\|\mathbf{n}_i^s)$ \\
        \midrule
        $D_{\text{KL}}$ & 0.00159 & 0.17921 & 0.01324 \\
        \bottomrule
    \end{tabular}
    \end{center}
\end{table}

\subsubsection{Analysis of Different Noisy Images from the Same Camera.}
Another important property of the Camera-Encoding Network is that it must capture camera-specific characteristics from a noisy image, and the extracted latent vector should be irrelevant to the image content of the input noisy image.
To verify this, we randomly select five different noisy images from the same camera. 
These different noisy images are fed into the Camera-Encoding Network, while other inputs for the Noise Generating Network are kept fixed.
Because these noisy images are from the same camera, the generated noise should be robust and consistent.
Table~\ref{table:noisydiff} shows that the $D_\mathrm{KL}$ between the generated noise and real noise remains low for different noisy images.
\setlength{\tabcolsep}{4pt}
\begin{table}[t]
    \footnotesize
    \begin{center}
    \caption{\textbf{Analysis of different noisy images from the same camera.} The Kullback-Leibler divergence results from five randomly selected noisy images but fixed inputs for the generator}
    \label{table:noisydiff}
    \begin{tabular}{c|ccccc}
        \toprule
        Noisy image sets & $1^\mathrm{st}$ & $2^\mathrm{nd}$ & $3^\mathrm{rd}$ & $4^\mathrm{th}$ & $5^\mathrm{th}$ \\ 
        \midrule
        $D_{\text{KL}}$ & 0.00159 & 0.00180 & 0.00183 & 0.00163 & 0.00176 \\ 
        \bottomrule
    \end{tabular}
    \end{center}
\end{table}
\setlength{\tabcolsep}{1.4pt}

\section{Application to Real Image Denoising} \label{subs:results:nr}
\subsection{Real-world Image Denoising}
We conduct real-world denoising experiments to further compare different noise models.
For all noise models, we follow Noise Flow~\cite{NoiseFlow} to use the same 9-layer DnCNN network~\cite{DnCNN} as the baseline denoiser. 
Learning-based noise models (Noise Flow and ours) are trained with SIDD dataset. We then train a denoiser network with synthetic training pairs generated by each noise model separately.

Table~\ref{table:nr} shows the average PSNR and SSIM~\cite{wang2004image}  on the test set.
The denoisers trained with statistical noise models (Gaussian and Poisson-Gaussian) are worse than those trained with learning-based noise models (Noise Flow and Ours), which also outperform the denoiser trained with real data only (the last row of Table~\ref{table:nr}).
This is because the amount of synthetic data generated by noise models is unlimited, while the amount of real data is fixed.

Our noise model outperforms Noise Flow in terms of both PSNR and SSIM while using more training data for training noise models leads to better denoising performance.
Table~\ref{table:nr} also shows that using both real data and our noise model results in further improved PSNR and SSIM.

\setlength{\tabcolsep}{4pt}
\begin{table}[t]
    \footnotesize
    \begin{center}
    \caption{\textbf{Real-World image denoising.}
    The denoising networks using our noise model outperform those using existing statistical noise models and learning-based models.
    \textcolor{red}{\pmb{Red}} indicates the best and \textcolor{blue}{\underline{blue}} indicates the second best performance
    (While training using both synthetic and real data, Ours + Real, synthetic and real data are sampled by a ratio of $5:1$ in each mini-batch) }  \label{table:nr}
    % \resizebox{\linewidth}{!}{
    % \renewcommand{\tabcolsep}{1pt} % adjust horizontal space
    % \renewcommand{\arraystretch}{0.5} % adjust vertical space
    \begin{tabular}{l|cc|cc}
        \toprule
        & \# of training data &\# of \emph{real} training  && \\
        Noise model & for noise model & data for denoiser & PSNR & SSIM\\
        \midrule
        Gaussian & -  & - & 43.63 & 0.968 \\
        \midrule
        Poisson-Gaussian & -  & - & 44.99 & 0.982 \\
        \midrule
        \multirow{2}{*}{Noise Flow~\cite{NoiseFlow}} & 100k  & - & 47.49 & 0.991 \\
        & 500k  & - & 48.52 & 0.992 \\
        \midrule 
        \multirow{2}{*}{Ours}   & 100k  & - & 47.97 & 0.992 \\ 
        & 500k  & - & \textcolor{blue}{\underline{48.71}} & \textcolor{blue}{\underline{0.993}} \\
        \midrule 
        \multirow{2}{*}{Ours + Real}  & 100k  & 100k & 47.93 & \textcolor{red}{\pmb{0.994}} \\
        & 500k  & 500k & \textcolor{red}{\pmb{48.72}} & \textcolor{red}{\pmb{0.994}} \\
        \midrule
        \multirow{2}{*}{Real only}   & -    & 100k     & 47.08 & 0.989 \\
        & -    & 500k     & 48.30 & \textcolor{red}{\pmb{0.994}} \\
        \bottomrule
    \end{tabular}
    % }
    \end{center}
\end{table}
\setlength{\tabcolsep}{1.4pt}

\subsection{Camera-Specific Denoising Networks}
To verify the camera-aware ability of our method, we train denoiser networks with our generative noise models, which are trained with and without the Camera-Encoding Network (and with and without the triplet loss) respectively.

For our noise model without the Camera-Encoding Network, we train a single generic denoiser network for all cameras.
For our noise models with the Camera-Encoding Network, we train camera-specific denoiser networks with and without the triplet loss for each camera.
The denoising performance is shown in Table~\ref{table:cameradiff}.

The results show that the Camera-Encoding Network with the triplet loss can successfully capture camera-specific noise characteristics and thus enhance the performance of camera-specific denoiser networks.

\setlength{\tabcolsep}{4pt}
\begin{table}[t]
    \footnotesize
    \begin{center}
    \caption{\textbf{Real-world image denoising using camera-aware noise model.} Grouping the proposed Camera-Encoding Network and triplet loss $L_{\text{Triplet}}$ can extract camera-specific latent vectors and thus improve camera-specific denoiser networks}
    \label{table:cameradiff}
    \begin{tabular}{l|ccccc}
        \toprule
        & \multicolumn{5}{c}{PSNR on test cameras} \\
        Model & IP & GP & S6 & N6 & G4 \\
        \midrule
        w/o $(E+L_{\text{Triplet}})$  & 57.4672 & 44.5180 & 40.0183 & 44.7954 & 51.8048 \\
        with $E$, w/o $L_{\text{Triplet}}$ & 49.8788 & 45.7755 & 40.4976 & 41.8447 & 51.8139 \\
        with $(E+L_{\text{Triplet}})$ & \pmb{58.6073} & \pmb{45.9624} & \pmb{41.8881} & \pmb{46.4726} & \pmb{53.2610} \\
        \bottomrule
    \end{tabular}
    \end{center}
\end{table}
\setlength{\tabcolsep}{1.4pt}
\section{Conclusions} \label{s:conclusion}
We have presented a novel learning-based generative method for real-world noise. 
The proposed noise model outperforms existing statistical models and learning-based methods quantitatively and qualitatively.
Moreover, the proposed method can capture different characteristics of different camera sensors in a single noise model. 
We have also demonstrated that the real-world image denoising task can benefit from our noise model.
As for future work, modeling real-world noise with few-shot or one-shot learning could be a possible direction. This could reduce the burden of collecting real data for training a learning-based noise model.

\def\fo{1014}
\def\foo{1073}
\def\fooo{1240}
\setlength{\tabcolsep}{1.4pt}
\begin{figure}[ht]
    \centering
    \tiny
    \begin{tabular}{ccccccc}
        Real Noisy & Poisson-Gauss. & Real Only & Noise Flow & Ours & Ours + Real & Ground Truth \\
        \includegraphics[width=\figwidthnr\columnwidth]{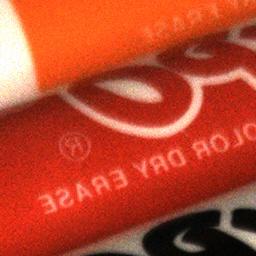} &
        \includegraphics[width=\figwidthnr\columnwidth]{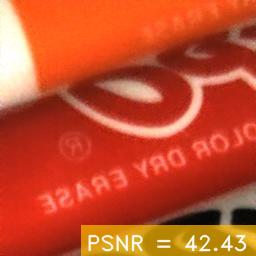} &
        \includegraphics[width=\figwidthnr\columnwidth]{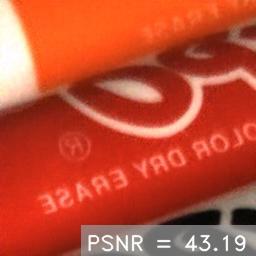} &
        \includegraphics[width=\figwidthnr\columnwidth]{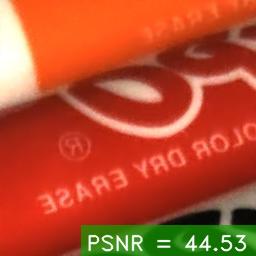} &
        \includegraphics[width=\figwidthnr\columnwidth]{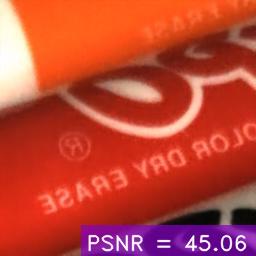} &
        \includegraphics[width=\figwidthnr\columnwidth]{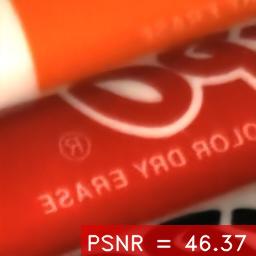} &
        \includegraphics[width=\figwidthnr\columnwidth]{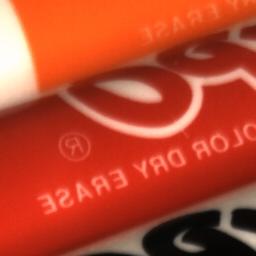} \\
        \includegraphics[width=\figwidthnr\columnwidth]{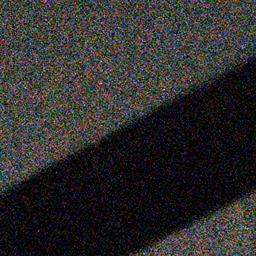} &
        \includegraphics[width=\figwidthnr\columnwidth]{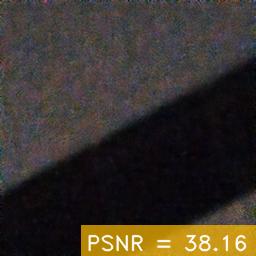} &
        \includegraphics[width=\figwidthnr\columnwidth]{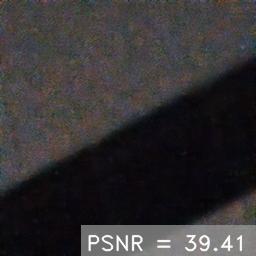} &
        \includegraphics[width=\figwidthnr\columnwidth]{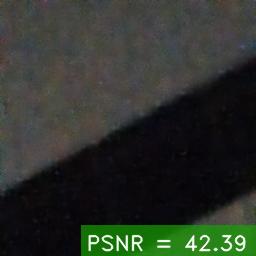} &
        \includegraphics[width=\figwidthnr\columnwidth]{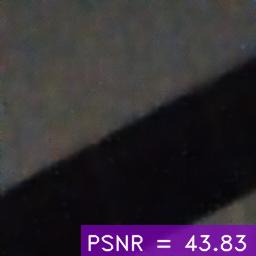} &
        \includegraphics[width=\figwidthnr\columnwidth]{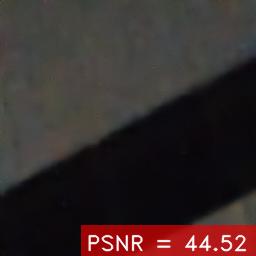} &
        \includegraphics[width=\figwidthnr\columnwidth]{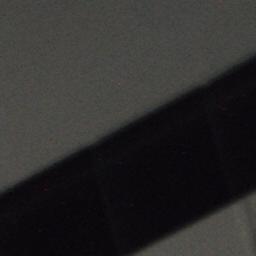} \\
        \includegraphics[width=\figwidthnr\columnwidth]{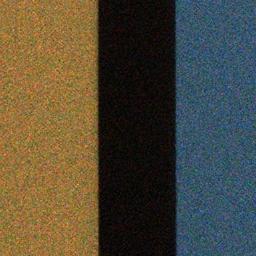} &
        \includegraphics[width=\figwidthnr\columnwidth]{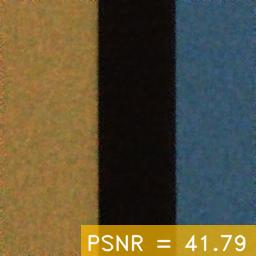} &
        \includegraphics[width=\figwidthnr\columnwidth]{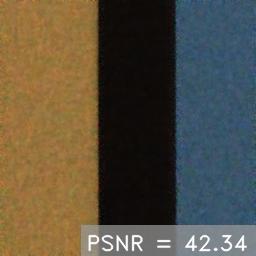} &
        \includegraphics[width=\figwidthnr\columnwidth]{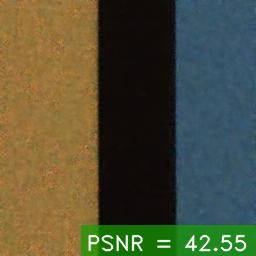} &
        \includegraphics[width=\figwidthnr\columnwidth]{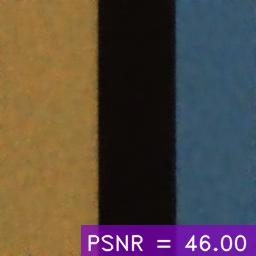} &
        \includegraphics[width=\figwidthnr\columnwidth]{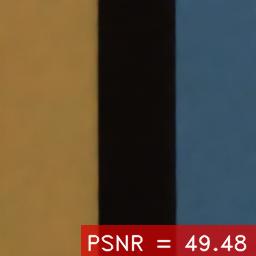} &
        \includegraphics[width=\figwidthnr\columnwidth]{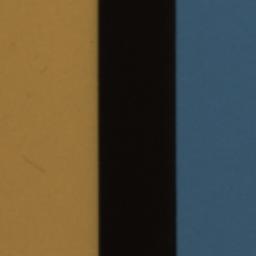} \\
    \end{tabular}
    \caption{\textbf{Results of denoisers trained on different noise models.} We compare the denoised results trained on different settings, 1) only real pairs, 2) synthetic pairs with Noise Flow or the proposed method, and 3) the mixture of synthetic pairs from ours and real pairs
    }
    \label{fig:nr}
\end{figure}

\clearpage
\bibliographystyle{splncs04}
\bibliography{egbib}

\clearpage
% ---- Appendix ----

\newcommand\figwidthiso{0.152}
\newcommand\figwidthsuppnr{0.2}
\newcommand\figwidthsuppnm{0.19}

\appendix

\setcounter{table}{0}
\renewcommand\thetable{\Alph{section}-\arabic{table}}
\setcounter{figure}{0}
\renewcommand\thefigure{\Alph{section}-\arabic{figure}}

\section{Network Architectures}
\label{network_architecture}
We apply a U-Net~\cite{UNet} architecture to the generator $G$, and Table~\ref{table:ArchG} shows the detailed configuration. The first five layers correspond to the encoder followed by four residual blocks, and the last five layers are the decoder. The residual block consists of two consecutive convolutions as well as a skip connection across the block. Besides, the latent vector from the camera encoder $E$ is concatenated with the output of the $2^{\text{nd}}$ residual block.

Table~\ref{table:ArchD} shows the architecture of the discriminator $D$, which is similar to PatchGAN~\cite{PatchGAN}. The $\text{out}_{D_f}$ and $\text{out}_{D}$ are used for the feature matching loss $L_\mathrm{FM}$ and adversarial loss $L_\mathrm{Adv}$, respectively. Note that $D$ determines the score of realness at the scale of $46\times46$ according to the receptive field.

Finally, the architecture of the camera encoder $E$ is shown in Table~\ref{table:ArchE}. To make latent vectors irrelevant to the spatial domain, we perform a global average pooling at the last layer. The latent vector is then concatenated with the middle features of $G$ by expanding the spatial dimension. 

\setlength{\tabcolsep}{4pt}
\begin{table}[b]
    \begin{center}
    \caption{\textbf{Architecture of the generator.} The notation $[\cdot, \cdot]$ represents concatenation and C, RES, T respectively denote convolution, residual block, and transposed convolution. The $\emph{SN-IN}$ indicates a Spectral Normalization~\cite{SN} followed by an Instance Normalization~\cite{IN} and LReLU is the Leaky ReLU~\cite{LReLU}}
    \label{table:ArchG}
    \begin{tabular}{c|ccccccccc}
    \toprule
    
       &       &        & Kernel & \multicolumn{2}{c}{Channels}&        &      &       & Output \\
       & Input & Output & Size   & In & Out & Stride & Norm. & Activ. & Size \\
    \midrule
    C &{$\text{in}_{G}$}&{c1}& $4 \times 4$ & 8   & 64  & 2 & -     &LReLU&$\frac{h}{2}\times\frac{w}{2}$ \\
    C &{c1}&{c2}& $4 \times 4$ & 64  & 128 & 2 &\emph{SN-IN}&LReLU&$\frac{h}{4}\times\frac{w}{4}$ \\
    C &{c2}&{c3}& $4 \times 4$ & 128 & 256 & 2 &\emph{SN-IN}&LReLU&$\frac{h}{8}\times\frac{w}{8}$ \\
    C &{c3}&{c4}& $4 \times 4$ & 256 & 512 & 2 &\emph{SN-IN}&LReLU&$\frac{h}{16}\times\frac{w}{16}$ \\
    C &{c4}&{c5}& $4 \times 4$ & 512 & 512 & 2 &\emph{SN-IN}&LReLU&$\frac{h}{32}\times\frac{w}{32}$ \\
    \midrule
    RES &{c5}&{res1}& $3 \times 3$ & 512 & 512 & 1 & - & - & $\frac{h}{32}\times\frac{w}{32}$ \\
    RES &{res1}&{res2}& $3 \times 3$ & 512 & 512 & 1 & - & - & $\frac{h}{32}\times\frac{w}{32}$ \\
    RES &{[res2, $\text{out}_{E}$]}&{res3}& $3 \times 3$ & 1024 & 1024 & 1 & - & - & $\frac{h}{32}\times\frac{w}{32}$ \\
    RES &{res3}&{res4}& $3 \times 3$ & 1024 & 1024 & 1 & - & - & $\frac{h}{32}\times\frac{w}{32}$ \\
    \midrule
    T &{res4}&{t1}& $4 \times 4$ & 1024& 512 & 1/2 &\emph{SN-IN}& LReLU & $\frac{h}{16}\times\frac{w}{16}$ \\
    T &{[t1, c4]}&{t2}& $4 \times 4$ & 1024& 256 & 1/2 &\emph{SN-IN}& LReLU & $\frac{h}{8}\times\frac{w}{8}$ \\
    T &{[t2, c3]}&{t3}& $4 \times 4$ & 512 & 128 & 1/2 &\emph{SN-IN}& LReLU & $\frac{h}{4}\times\frac{w}{4}$ \\
    T &{[t3, c2]}&{t4}& $4 \times 4$ & 256 & 64  & 1/2 &\emph{SN-IN}& LReLU & $\frac{h}{2}\times\frac{w}{2}$ \\
    T &{[t4, c1]}&{$\text{out}_{G}$}& $4 \times 4$ & 128 & 4   & 1/2 &\emph{SN-IN}& Tanh  & $h\times w$ \\
    
    \bottomrule
    \end{tabular}
    \end{center}
\end{table}
\setlength{\tabcolsep}{1.4pt}

\setlength{\tabcolsep}{4pt}
\begin{table}[ht]
    \begin{center}
    \caption{\textbf{Architecture of the discriminator}}
    \label{table:ArchD}
    \begin{tabular}{c|ccccccccc}
    \toprule
    
       &       &        & Kernel & \multicolumn{2}{c}{Channels}&        &      &       & Output \\
       & Input & Output & Size   & In & Out & Stride & Norm. & Activ. & Size \\
    \midrule
    C &{$\text{in}_{D}$}&{d1}& $4 \times 4$ & 8   & 64  & 2 & -    &LReLU& $\frac{h}{2}\times\frac{w}{2}$ \\
    C &{d1}&{d2}& $4 \times 4$ & 64  & 128 & 2 &\emph{SN-IN}&LReLU& $\frac{h}{4}\times\frac{w}{4}$ \\
    C &{d2}&{$\text{out}_{D_{f}}$}& $4 \times 4$ & 128 & 256 & 2 &\emph{SN-IN}&LReLU& $\frac{h}{8}\times\frac{w}{8}$ \\
    C &{$\text{out}_{D_{f}}$}&{$\text{out}_{D}$}& $4 \times 4$ & 256 & 1 & 1 &\emph{SN-IN}& -   & $\frac{h}{16}\times\frac{w}{16}$ \\
    
    \bottomrule
    \end{tabular}
    \end{center}
\end{table}
\setlength{\tabcolsep}{1.4pt}

\setlength{\tabcolsep}{4pt}
\begin{table}[ht]
    \begin{center}
    \caption{\textbf{Architecture of the camera encoder.} Note that POOL represents global average pooling}
    \label{table:ArchE}
    \begin{tabular}{c|ccccccccc}
    \toprule
    
       &       &        & Kernel & \multicolumn{2}{c}{Channels}&        &      &       & Output \\
       & Input & Output & Size   & In & Out & Stride & Norm. & Activ. & Size \\
    \midrule
    C &{$\text{in}_{E}$}&{e1}& $7 \times 7$ & 4   & 64  & 1 & -     &LReLU&$h \times w$ \\
    C &{e1}&{e2}  & $4 \times 4$ & 64  & 128 & 2 &\emph{SN-IN}&LReLU&$\frac{h}{2}\times\frac{w}{2}$ \\
    C &{e2}&{e3}  & $4 \times 4$ & 128 & 256 & 2 &\emph{SN-IN}&LReLU&$\frac{h}{4}\times\frac{w}{4}$ \\
    C &{e3}&{e4}  & $4 \times 4$ & 256 & 512 & 2 &\emph{SN-IN}&LReLU&$\frac{h}{8}\times\frac{w}{8}$ \\
    POOL &{e4}&{$\text{out}_{E}$} & - & 512 & 512& - & - & - & $1 \times 1$ \\
    
    \bottomrule
    \end{tabular}
    \end{center}
\end{table}
\setlength{\tabcolsep}{1.4pt}

\section{Control of Noise Levels}
\label{iso}
\setcounter{table}{0}
\renewcommand\thetable{\Alph{section}-\arabic{table}}
\setcounter{figure}{0}
\renewcommand\thefigure{\Alph{section}-\arabic{figure}}

Recall that the noise level of the final synthetic noise $\tilde{\mathbf{n}}$ can be controlled by adjusting the parameters of Poisson-Gaussian noise model for the initial synthetic noise $\tilde{\mathbf{n}}_\text{init}$. For the same camera, these parameters are proportional to the digital gain, which is highly correlated to the ISO. Therefore, different noise levels should be observed in different ISOs. Fig.~\ref{fig:isonm} shows the examples of noise and noisy image pairs from various noise models in a wide range of ISOs. We can find that as the ISO ascends, our noise samples become much noisier obviously. Moreover, our noise model always outperforms the compared methods in terms of Kullaback-Leibler divergence measurement.

\def\foo{3}
\setlength{\tabcolsep}{1.4pt}
\begin{figure}[t]
    \centering
    \scriptsize
    \renewcommand{\tabcolsep}{1pt} % adjust horizontal space
    \renewcommand{\arraystretch}{1} % adjust vertical space
    \begin{tabular}{ccccccc}
        & Gaussian & Poisson-Gauss. & Noise Flow & Ours & Real Noise & Clean \\
        \multirow{2}{*}{\raisebox{0.2\normalbaselineskip}[0pt][0pt]{\rotatebox[origin=c]{90}{100-N}}} &
        \includegraphics[width=\figwidthiso\columnwidth]{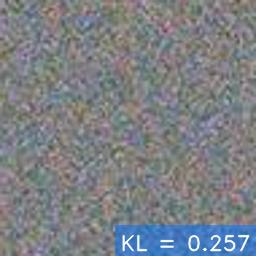} &
        \includegraphics[width=\figwidthiso\columnwidth]{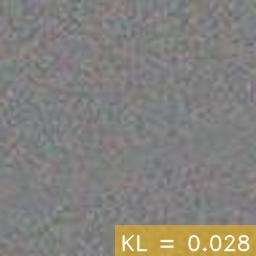} &
        \includegraphics[width=\figwidthiso\columnwidth]{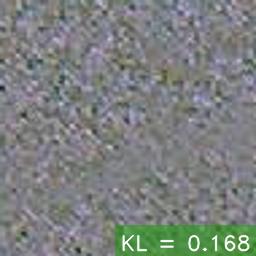} &
        \includegraphics[width=\figwidthiso\columnwidth]{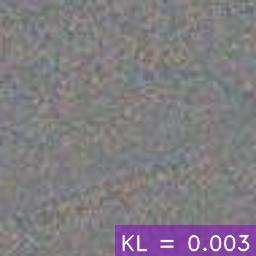} &
        \includegraphics[width=\figwidthiso\columnwidth]{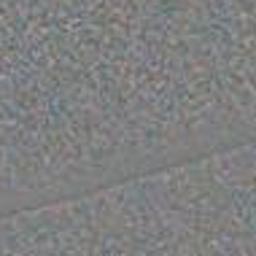} &
        \includegraphics[width=\figwidthiso\columnwidth]{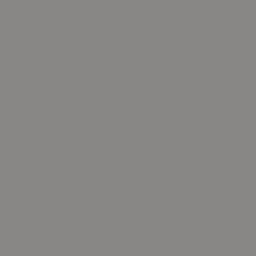} \\
        &
        \includegraphics[width=\figwidthiso\columnwidth]{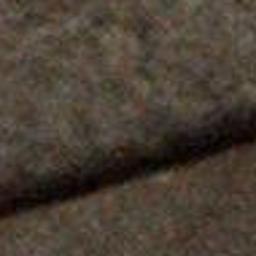} &
        \includegraphics[width=\figwidthiso\columnwidth]{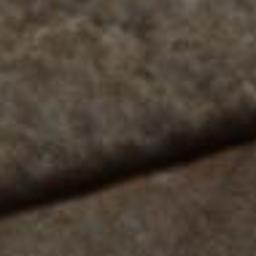} &
        \includegraphics[width=\figwidthiso\columnwidth]{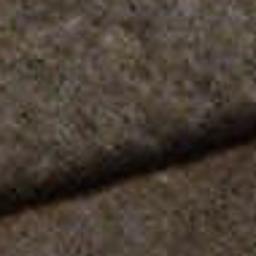} &
        \includegraphics[width=\figwidthiso\columnwidth]{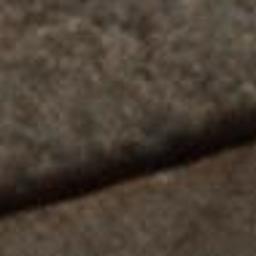} &
        \includegraphics[width=\figwidthiso\columnwidth]{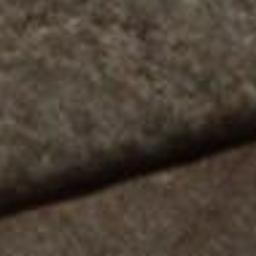} &
        \includegraphics[width=\figwidthiso\columnwidth]{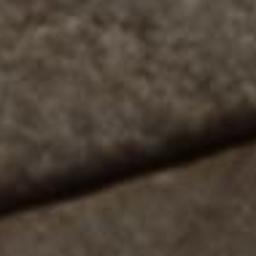} \\ \\
        \multirow{2}{*}{\raisebox{0.2\normalbaselineskip}[0pt][0pt]{\rotatebox[origin=c]{90}{400-N}}} &
        \includegraphics[width=\figwidthiso\columnwidth]{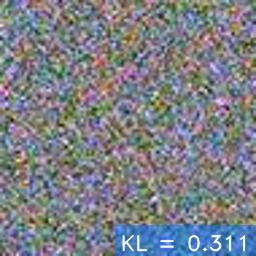} &
        \includegraphics[width=\figwidthiso\columnwidth]{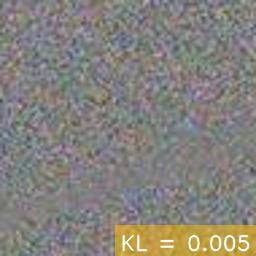} &
        \includegraphics[width=\figwidthiso\columnwidth]{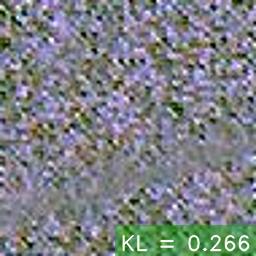} &
        \includegraphics[width=\figwidthiso\columnwidth]{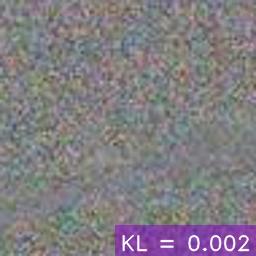} &
        \includegraphics[width=\figwidthiso\columnwidth]{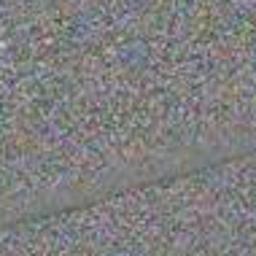} &
        \includegraphics[width=\figwidthiso\columnwidth]{fig/gray.jpg} \\
        &
        \includegraphics[width=\figwidthiso\columnwidth]{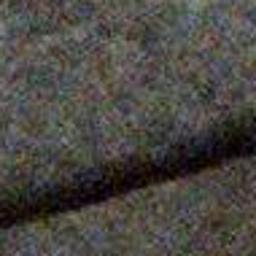} &
        \includegraphics[width=\figwidthiso\columnwidth]{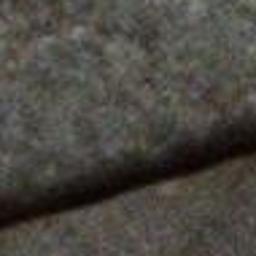} &
        \includegraphics[width=\figwidthiso\columnwidth]{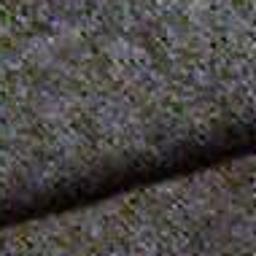} &
        \includegraphics[width=\figwidthiso\columnwidth]{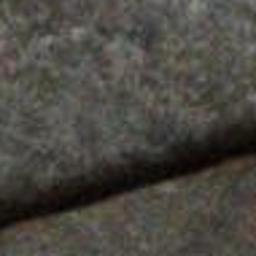} &
        \includegraphics[width=\figwidthiso\columnwidth]{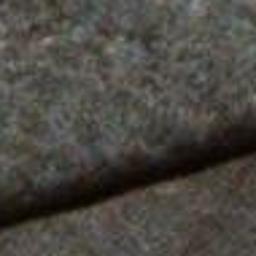} &
        \includegraphics[width=\figwidthiso\columnwidth]{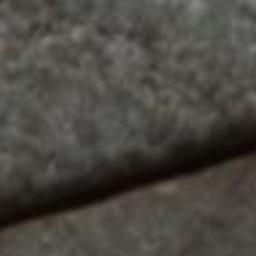}\\ \\
        \multirow{2}{*}{\raisebox{0.2\normalbaselineskip}[0pt][0pt]{\rotatebox[origin=c]{90}{800-N}}} &
        \includegraphics[width=\figwidthiso\columnwidth]{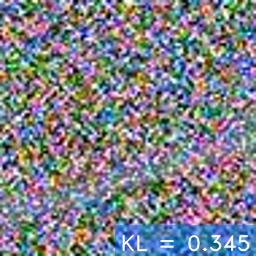} &
        \includegraphics[width=\figwidthiso\columnwidth]{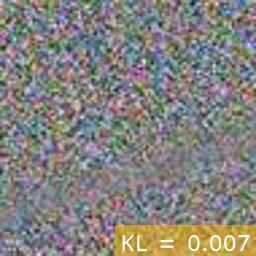} &
        \includegraphics[width=\figwidthiso\columnwidth]{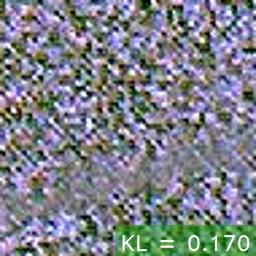} &
        \includegraphics[width=\figwidthiso\columnwidth]{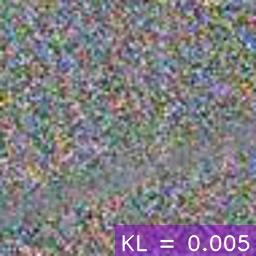} &
        \includegraphics[width=\figwidthiso\columnwidth]{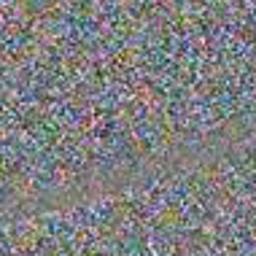} &
        \includegraphics[width=\figwidthiso\columnwidth]{fig/gray.jpg} \\
        &
        \includegraphics[width=\figwidthiso\columnwidth]{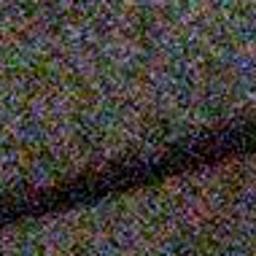} &
        \includegraphics[width=\figwidthiso\columnwidth]{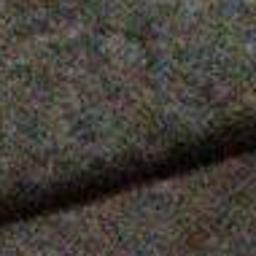} &
        \includegraphics[width=\figwidthiso\columnwidth]{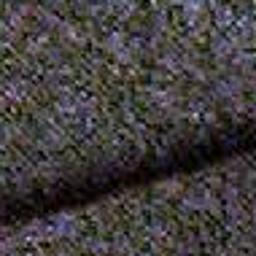} &
        \includegraphics[width=\figwidthiso\columnwidth]{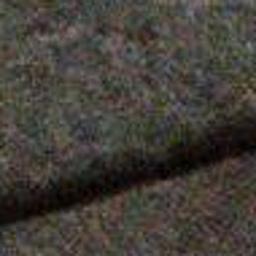} &
        \includegraphics[width=\figwidthiso\columnwidth]{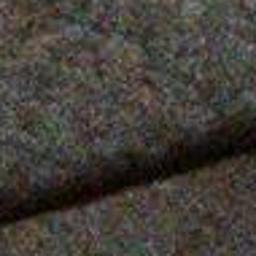} &
        \includegraphics[width=\figwidthiso\columnwidth]{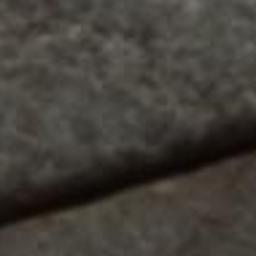}\\ \\
        \multirow{2}{*}{\raisebox{0.2\normalbaselineskip}[0pt][0pt]{\rotatebox[origin=c]{90}{1600-N}}} &
        \includegraphics[width=\figwidthiso\columnwidth]{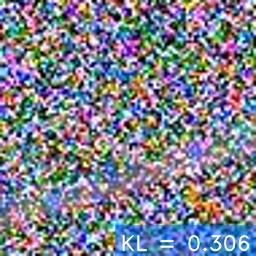} &
        \includegraphics[width=\figwidthiso\columnwidth]{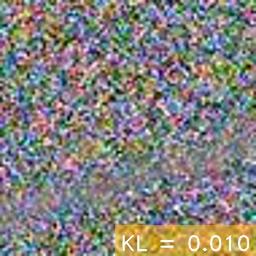} &
        \includegraphics[width=\figwidthiso\columnwidth]{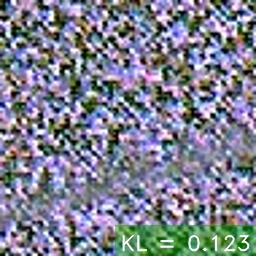} &
        \includegraphics[width=\figwidthiso\columnwidth]{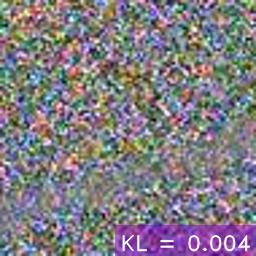} &
        \includegraphics[width=\figwidthiso\columnwidth]{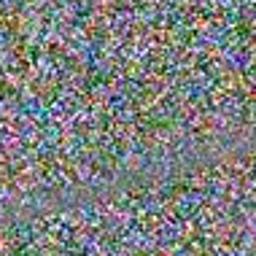} &
        \includegraphics[width=\figwidthiso\columnwidth]{fig/gray.jpg} \\
        
        &
        \includegraphics[width=\figwidthiso\columnwidth]{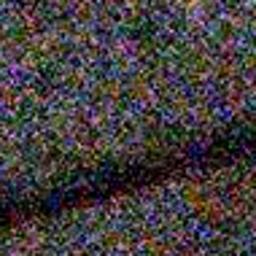} &
        \includegraphics[width=\figwidthiso\columnwidth]{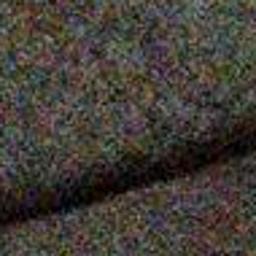} &
        \includegraphics[width=\figwidthiso\columnwidth]{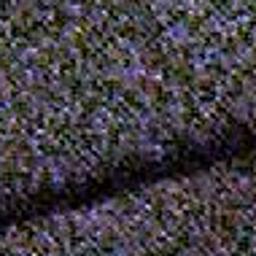} &
        \includegraphics[width=\figwidthiso\columnwidth]{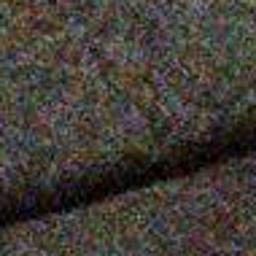} &
        \includegraphics[width=\figwidthiso\columnwidth]{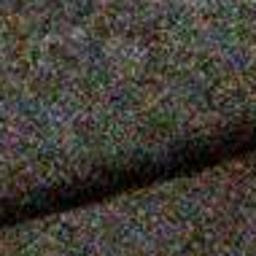} &
        \includegraphics[width=\figwidthiso\columnwidth]{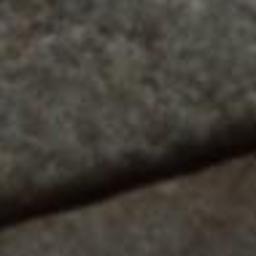}\\
        
    \end{tabular}
    \caption{\textbf{Different noise levels in different ISOs.} Each column represents a noise modeling method, and each two consecutive rows correspond to a pair of noise and noisy image in a specific ISO (from 100-N to 1600-N)}
    \label{fig:isonm}
\end{figure}

\section{More Qualitative Results}
\label{more_results_nm}
\setcounter{table}{0}
\renewcommand\thetable{\Alph{section}-\arabic{table}}
\setcounter{figure}{0}
\renewcommand\thefigure{\Alph{section}-\arabic{figure}}

\begin{figure}[ht]
    \centering
    \tiny
    \begin{tabular}{ccccc}
    \multicolumn{5}{c}{\includegraphics[width=0.98\columnwidth]{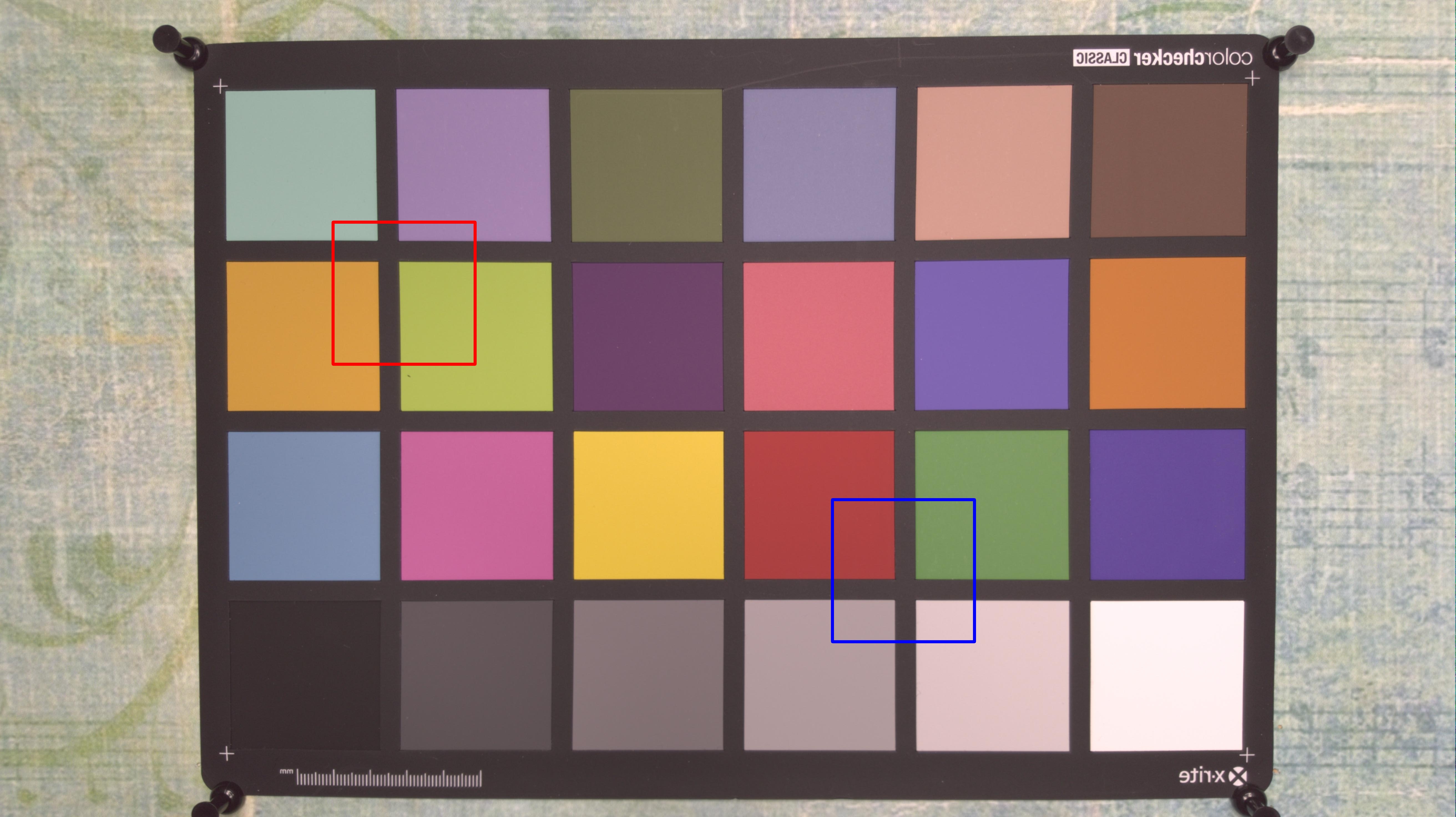}}\\
    %
    %&&&&\\
    %
    \multicolumn{5}{c}{Reference Clean Image}\\
    \includegraphics[width=\figwidthsuppnm\columnwidth]{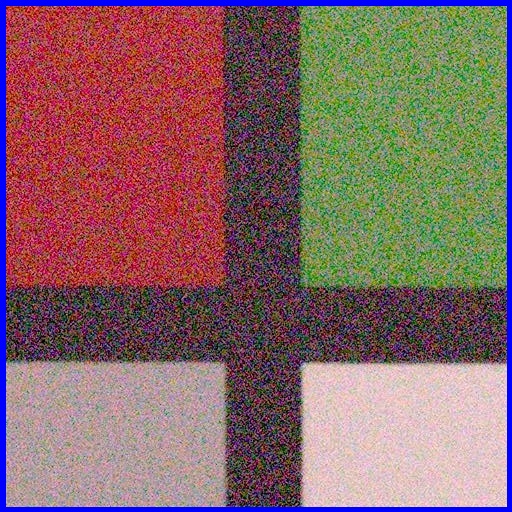} &
    \includegraphics[width=\figwidthsuppnm\columnwidth]{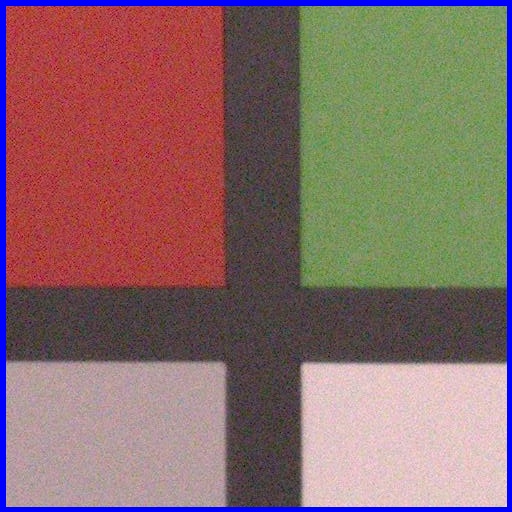} &
    \includegraphics[width=\figwidthsuppnm\columnwidth]{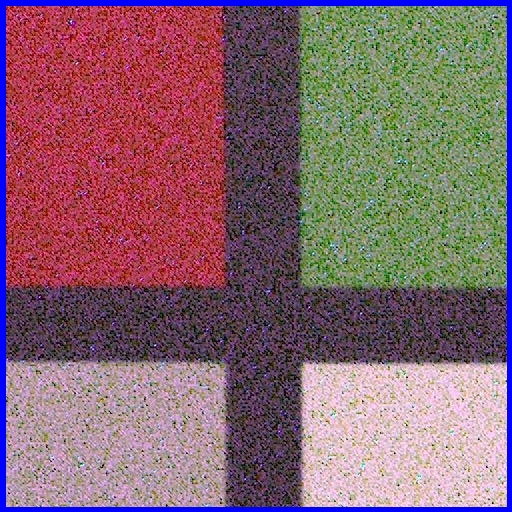} &
    \includegraphics[width=\figwidthsuppnm\columnwidth]{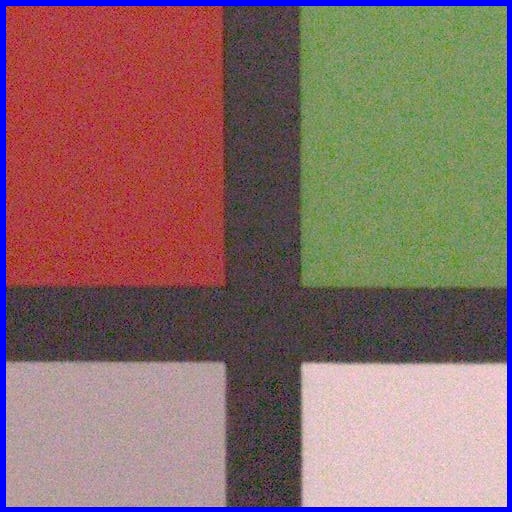} &
    \includegraphics[width=\figwidthsuppnm\columnwidth]{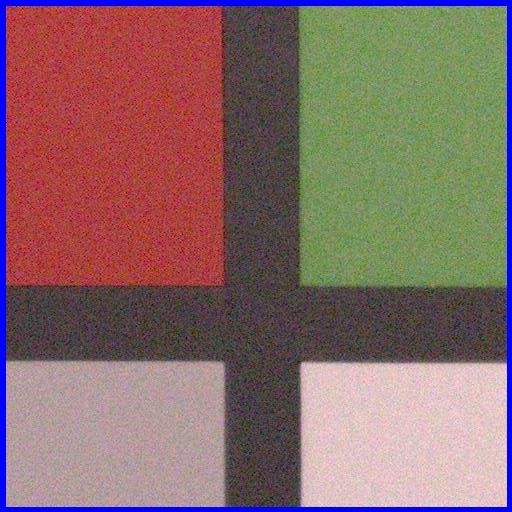} \\
    \includegraphics[width=\figwidthsuppnm\columnwidth]{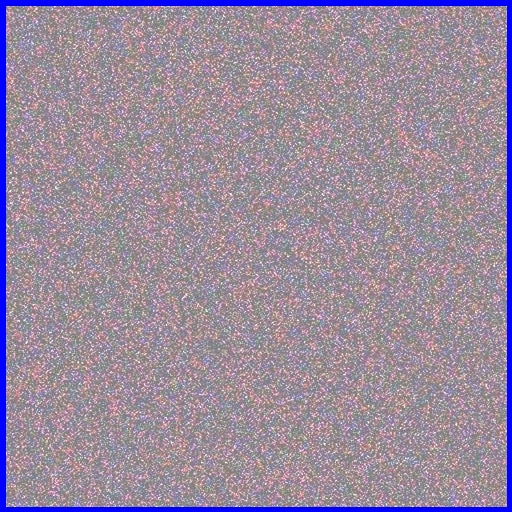} &
    \includegraphics[width=\figwidthsuppnm\columnwidth]{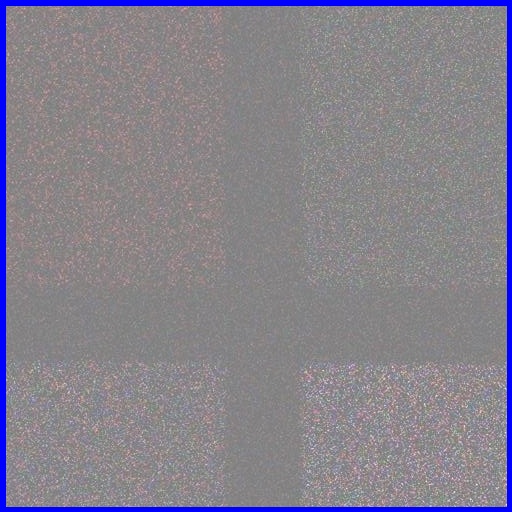} &
    \includegraphics[width=\figwidthsuppnm\columnwidth]{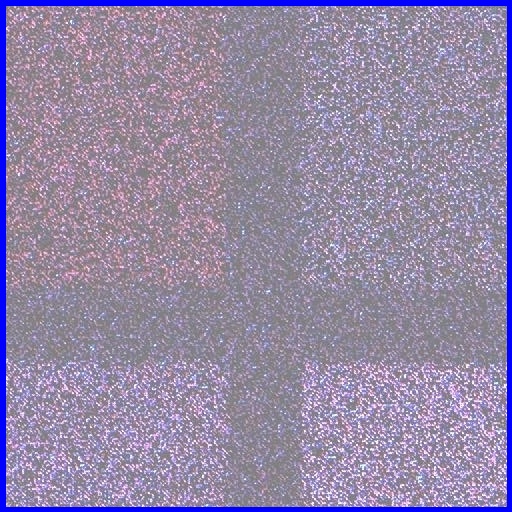} &
    \includegraphics[width=\figwidthsuppnm\columnwidth]{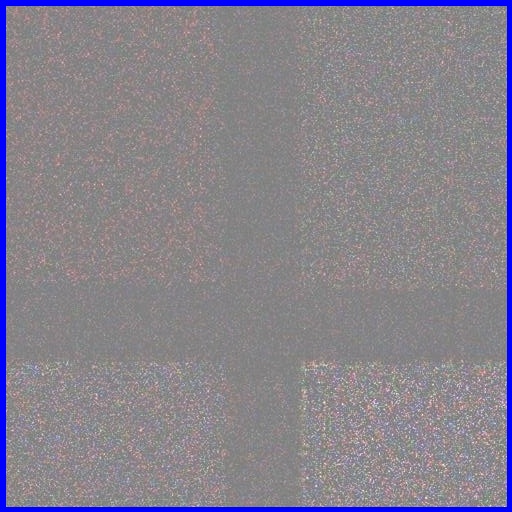} &
    \includegraphics[width=\figwidthsuppnm\columnwidth]{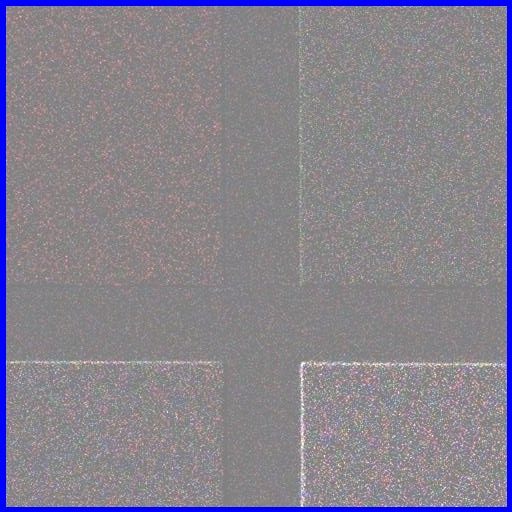} \\
    \includegraphics[width=\figwidthsuppnm\columnwidth]{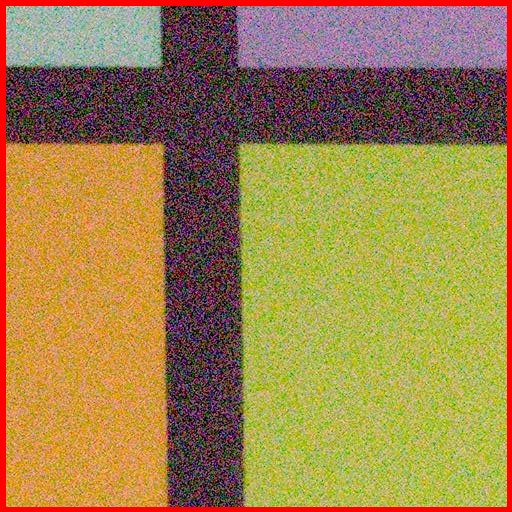} &
    \includegraphics[width=\figwidthsuppnm\columnwidth]{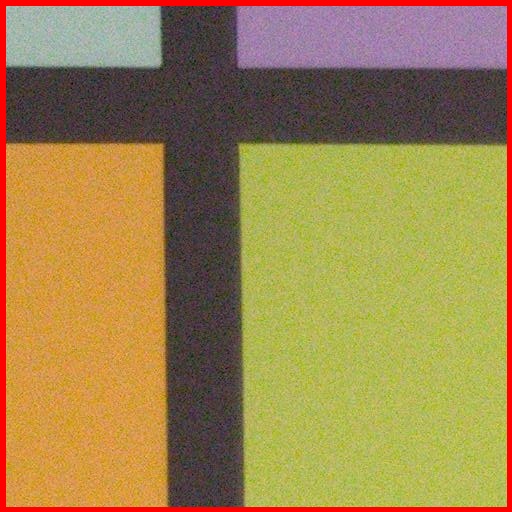} &
    \includegraphics[width=\figwidthsuppnm\columnwidth]{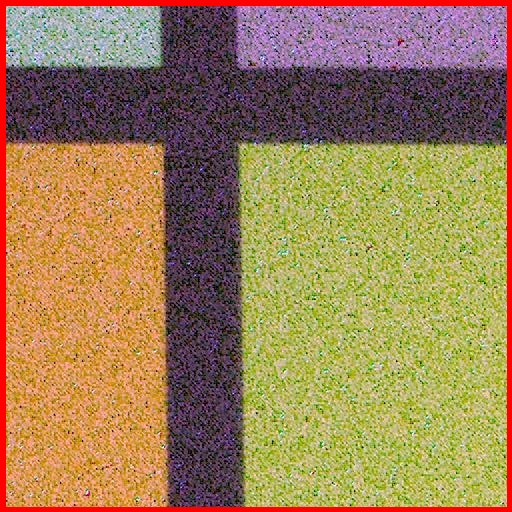} &
    \includegraphics[width=\figwidthsuppnm\columnwidth]{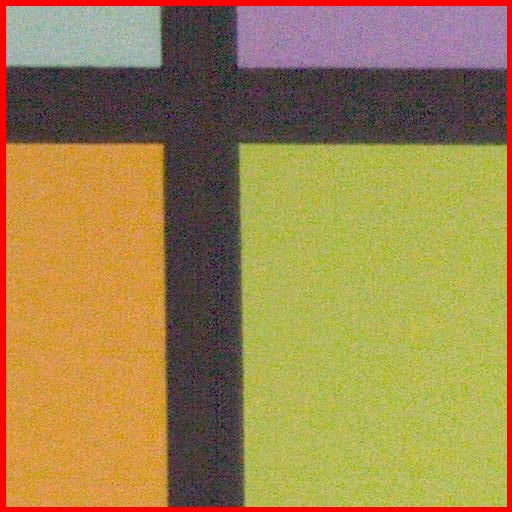} &
    \includegraphics[width=\figwidthsuppnm\columnwidth]{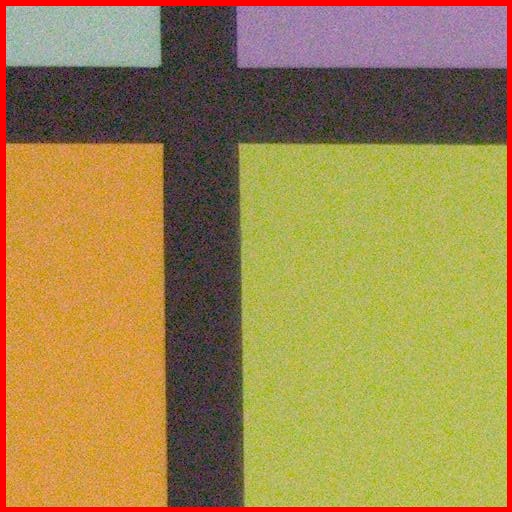} \\
    \includegraphics[width=\figwidthsuppnm\columnwidth]{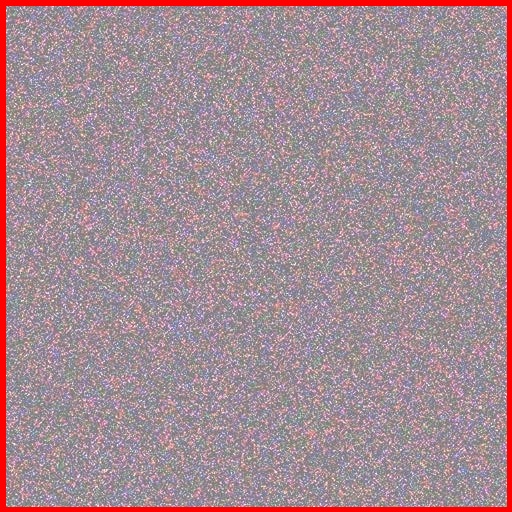} &
    \includegraphics[width=\figwidthsuppnm\columnwidth]{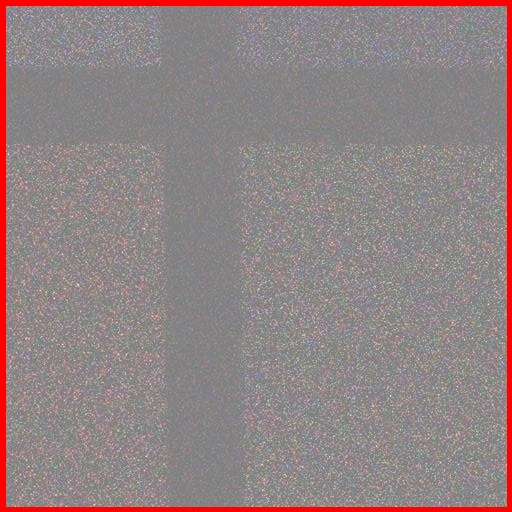} &
    \includegraphics[width=\figwidthsuppnm\columnwidth]{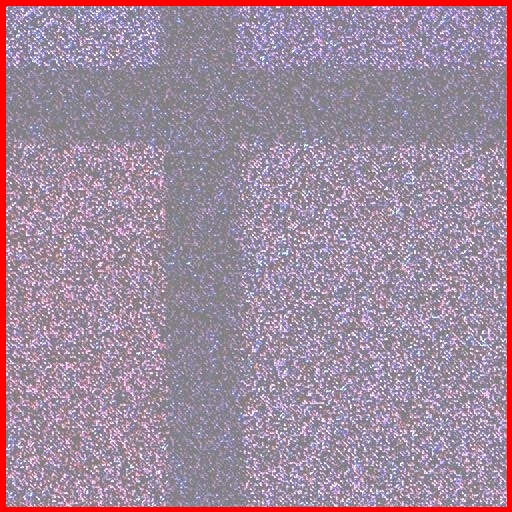} &
    \includegraphics[width=\figwidthsuppnm\columnwidth]{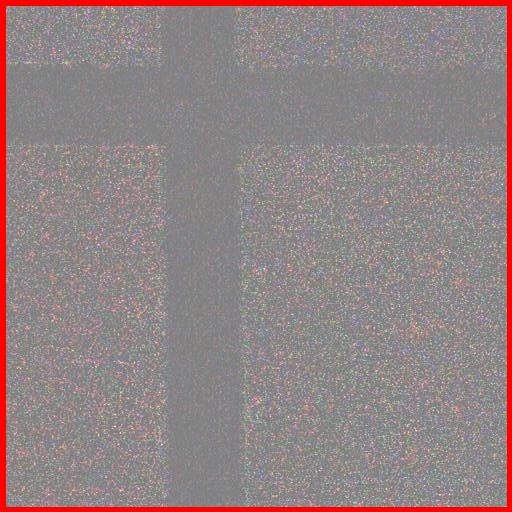} &
    \includegraphics[width=\figwidthsuppnm\columnwidth]{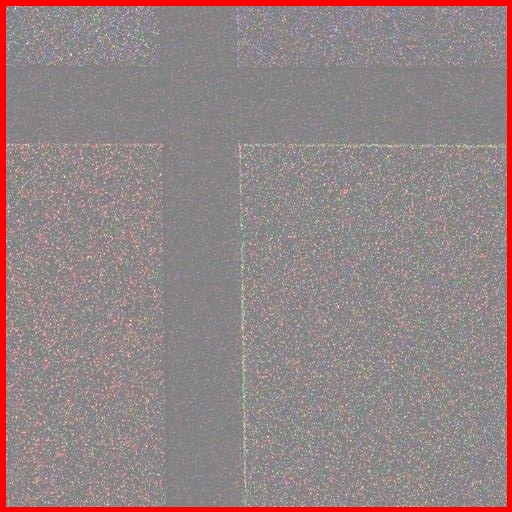} \\
    Gaussian & Poisson-Gauss. & Noise Flow & Ours & Real Noise \\
    \end{tabular}
    \caption{\textbf{More visual results of different noise models (scene 001)}}
    \label{fig:m1}
\end{figure}

\begin{figure}[ht]
    \centering
    \tiny
    \begin{tabular}{ccccc}
    \multicolumn{5}{c}{\includegraphics[width=0.98\columnwidth]{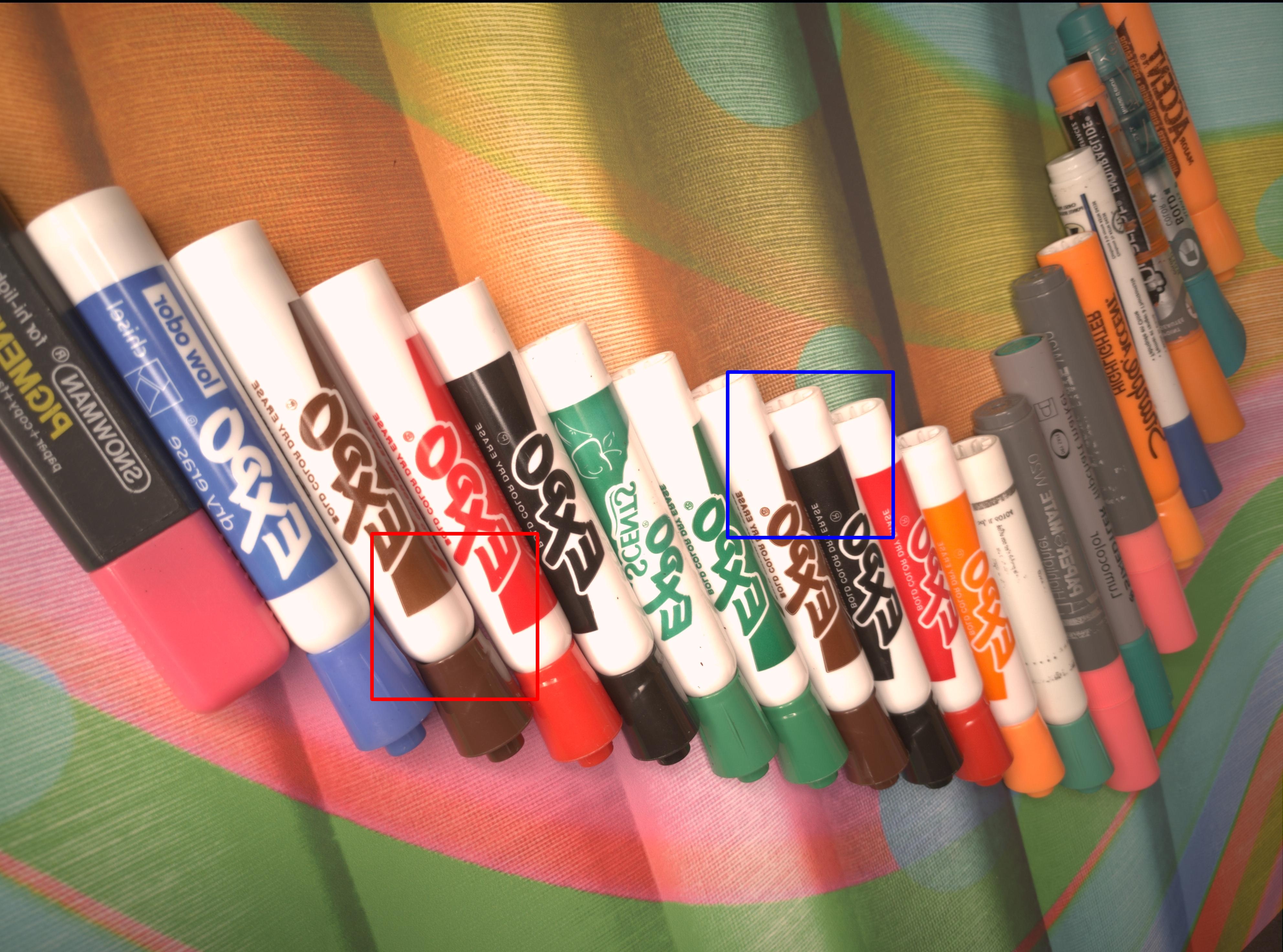}}\\
    %
    %&&&&\\
    %
    \multicolumn{5}{c}{Reference Clean Image}\\
    \includegraphics[width=\figwidthsuppnm\columnwidth]{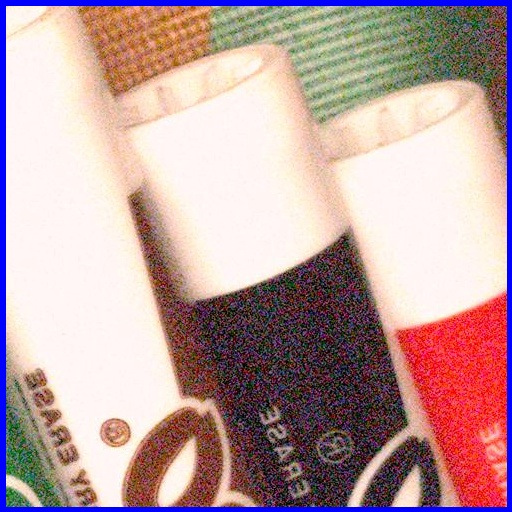} &
    \includegraphics[width=\figwidthsuppnm\columnwidth]{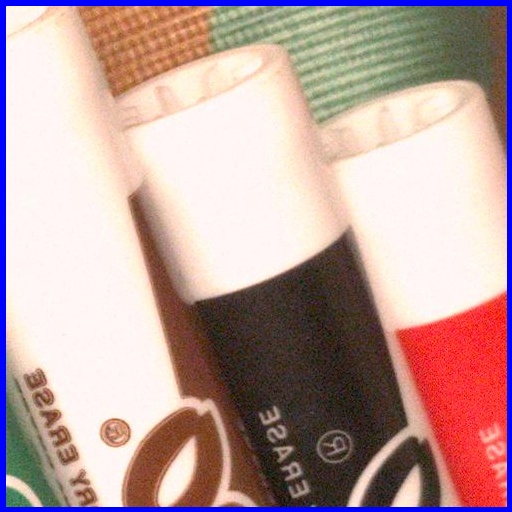} &
    \includegraphics[width=\figwidthsuppnm\columnwidth]{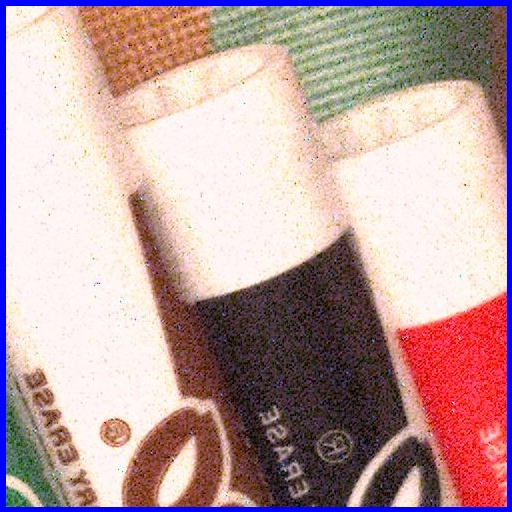} &
    \includegraphics[width=\figwidthsuppnm\columnwidth]{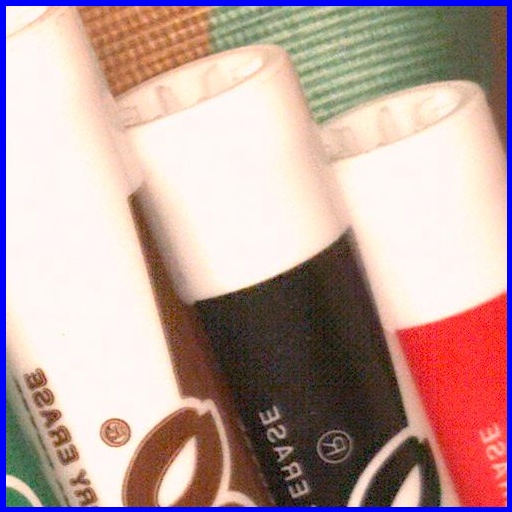} &
    \includegraphics[width=\figwidthsuppnm\columnwidth]{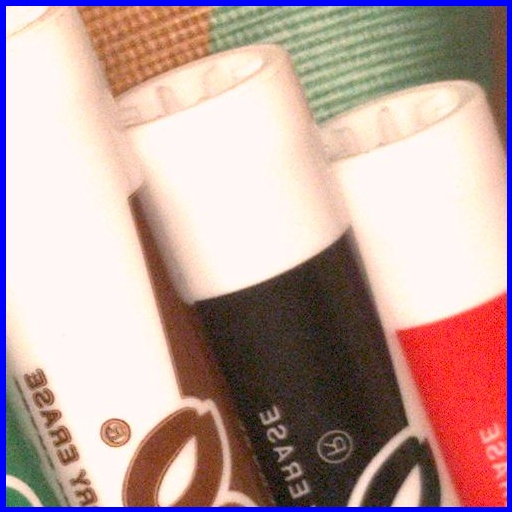} \\
    \includegraphics[width=\figwidthsuppnm\columnwidth]{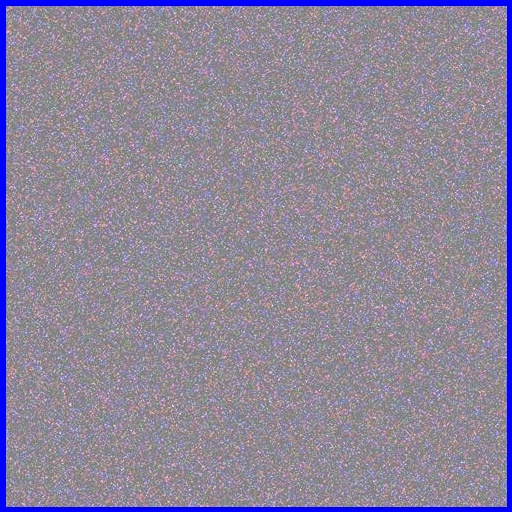} &
    \includegraphics[width=\figwidthsuppnm\columnwidth]{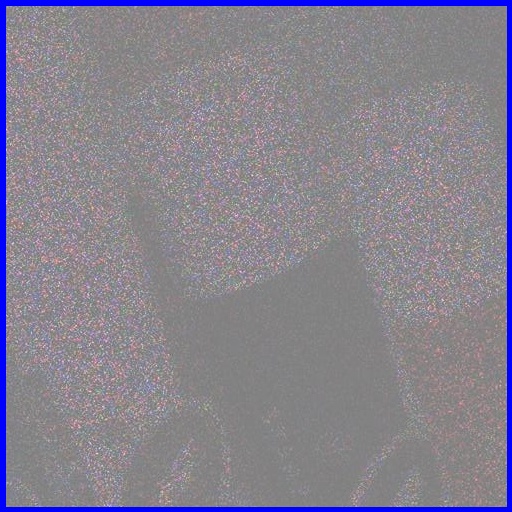} &
    \includegraphics[width=\figwidthsuppnm\columnwidth]{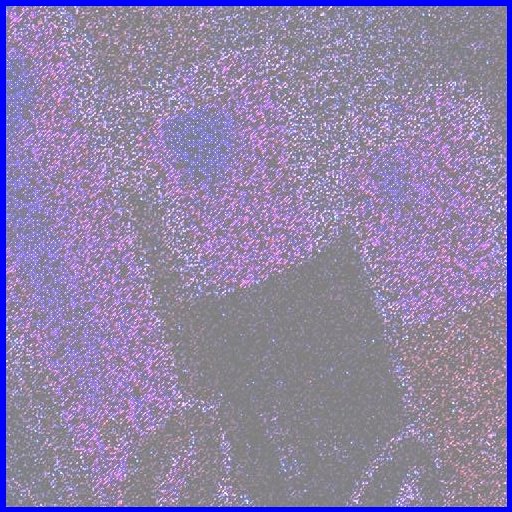} &
    \includegraphics[width=\figwidthsuppnm\columnwidth]{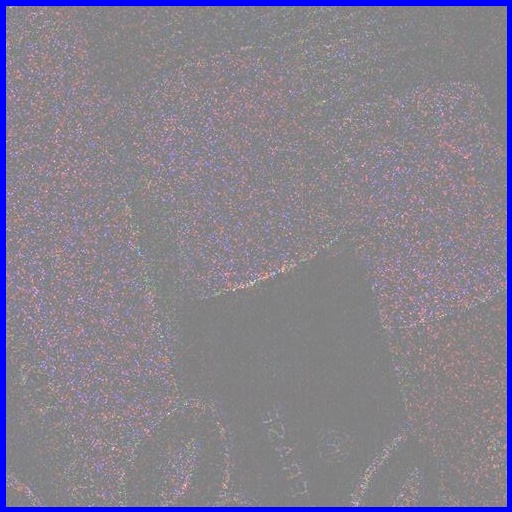} &
    \includegraphics[width=\figwidthsuppnm\columnwidth]{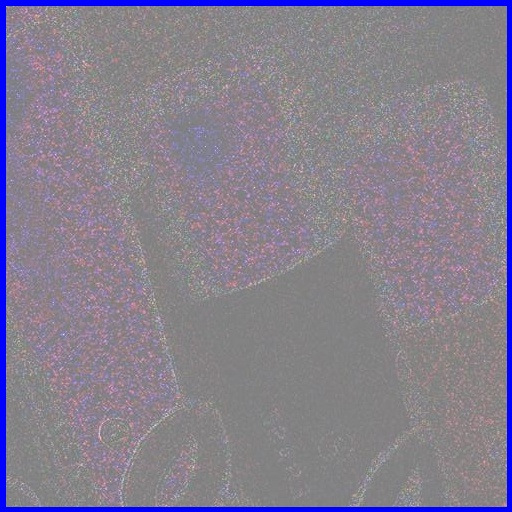} \\
    \includegraphics[width=\figwidthsuppnm\columnwidth]{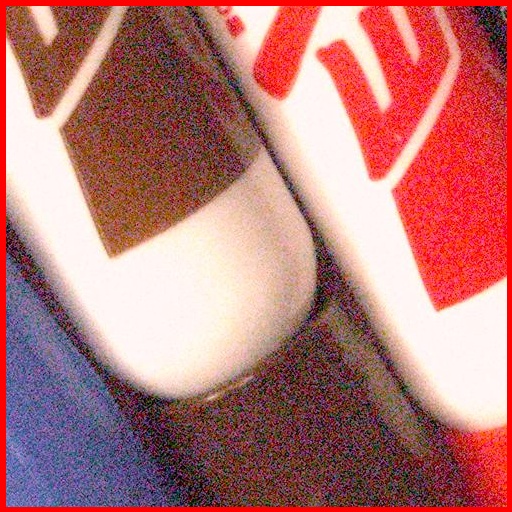} &
    \includegraphics[width=\figwidthsuppnm\columnwidth]{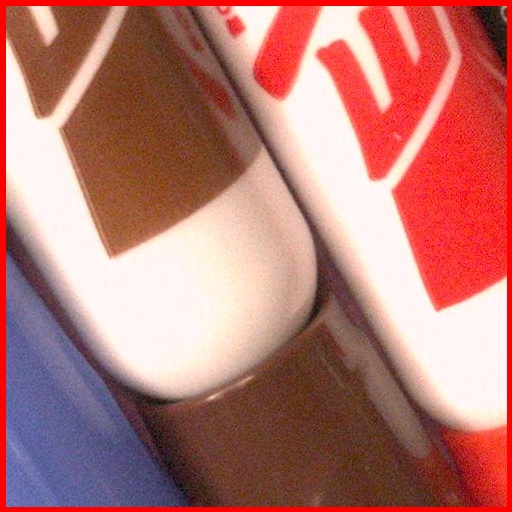} &
    \includegraphics[width=\figwidthsuppnm\columnwidth]{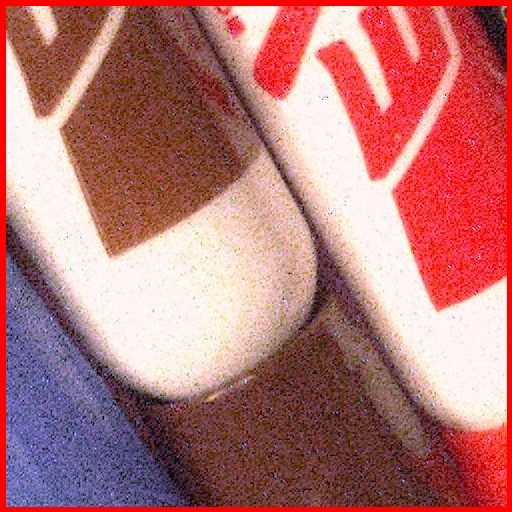} &
    \includegraphics[width=\figwidthsuppnm\columnwidth]{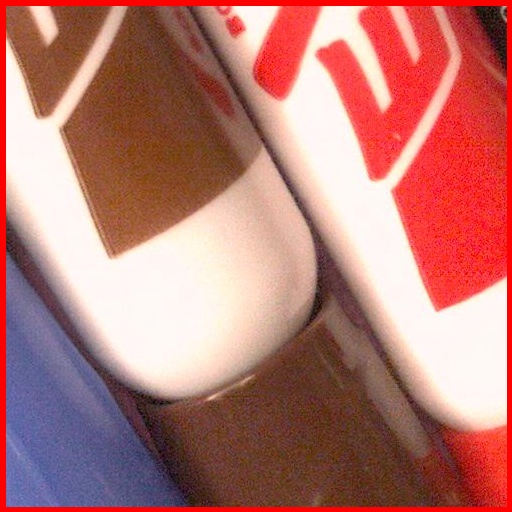} &
    \includegraphics[width=\figwidthsuppnm\columnwidth]{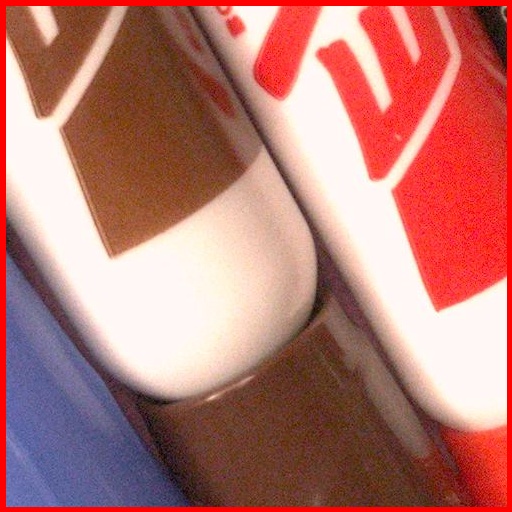} \\
    \includegraphics[width=\figwidthsuppnm\columnwidth]{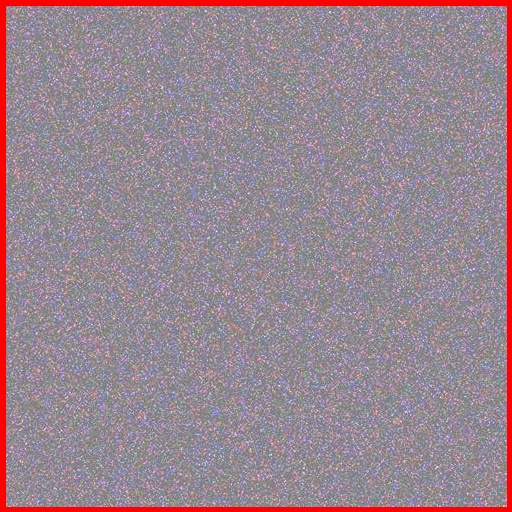} &
    \includegraphics[width=\figwidthsuppnm\columnwidth]{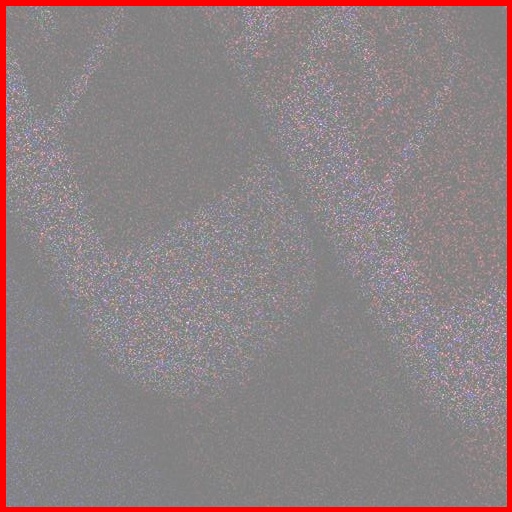} &
    \includegraphics[width=\figwidthsuppnm\columnwidth]{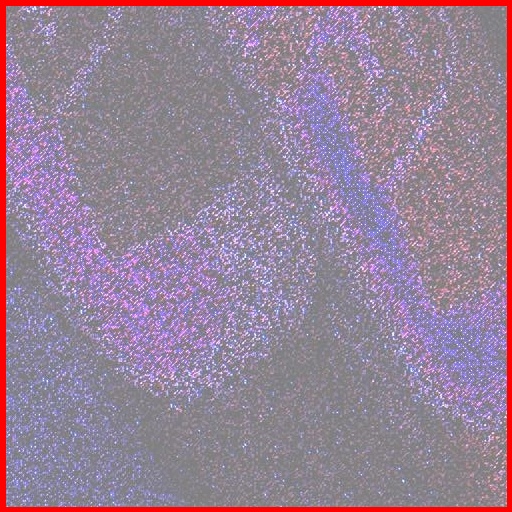} &
    \includegraphics[width=\figwidthsuppnm\columnwidth]{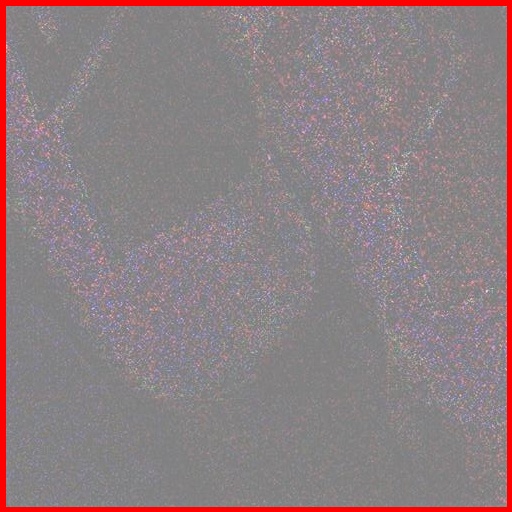} &
    \includegraphics[width=\figwidthsuppnm\columnwidth]{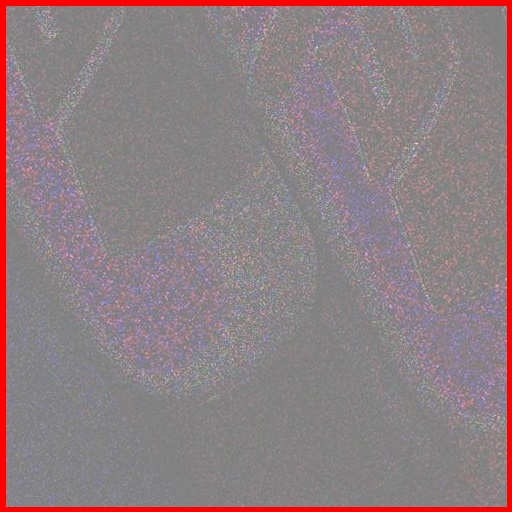} \\
    Gaussian & Poisson-Gauss. & Noise Flow & Ours & Real Noise \\
    \end{tabular}
    \caption{\textbf{More visual results of different noise models (scene 002)}}
    \label{fig:m2}
\end{figure}

\begin{figure}[ht]
    \centering
    \tiny
    \begin{tabular}{ccccc}
    \multicolumn{5}{c}{\includegraphics[width=0.98\columnwidth]{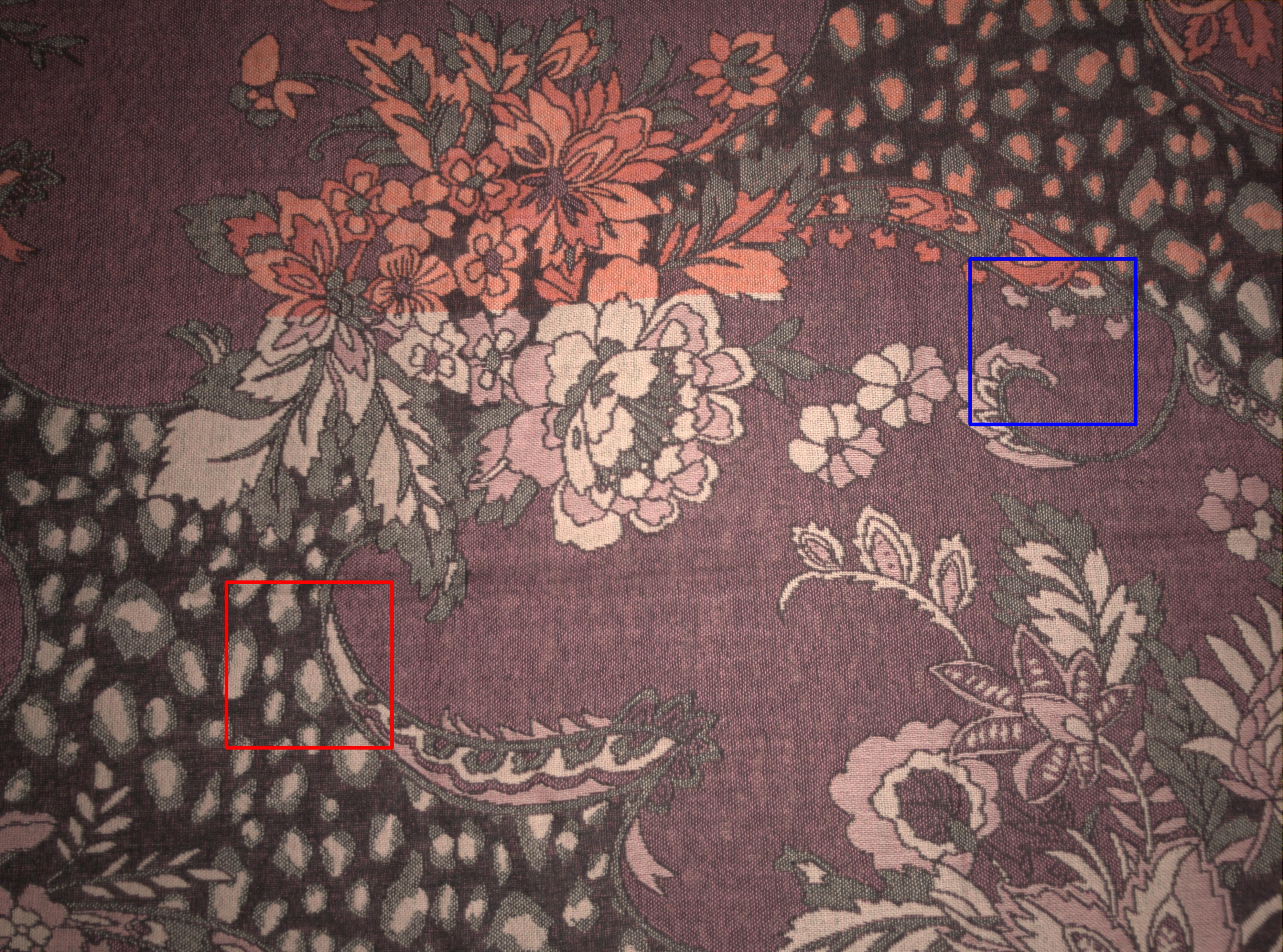}}\\
    %
    %&&&&\\
    %
    \multicolumn{5}{c}{Reference Clean Image}\\
    \includegraphics[width=\figwidthsuppnm\columnwidth]{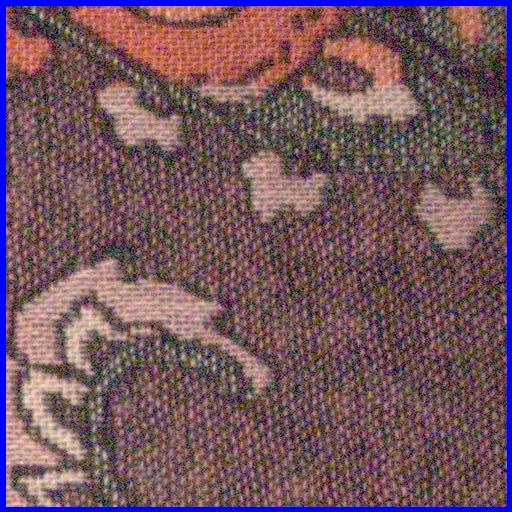} &
    \includegraphics[width=\figwidthsuppnm\columnwidth]{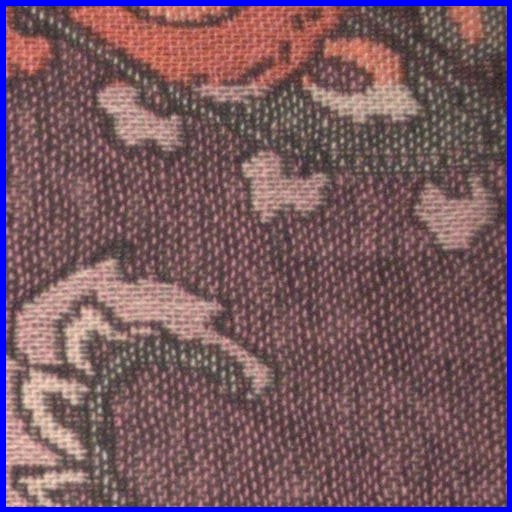} &
    \includegraphics[width=\figwidthsuppnm\columnwidth]{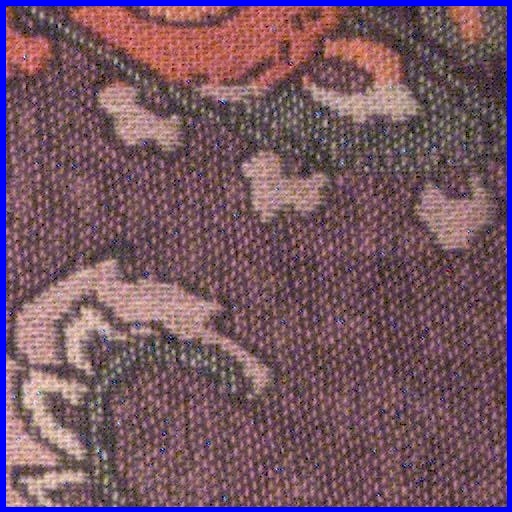} &
    \includegraphics[width=\figwidthsuppnm\columnwidth]{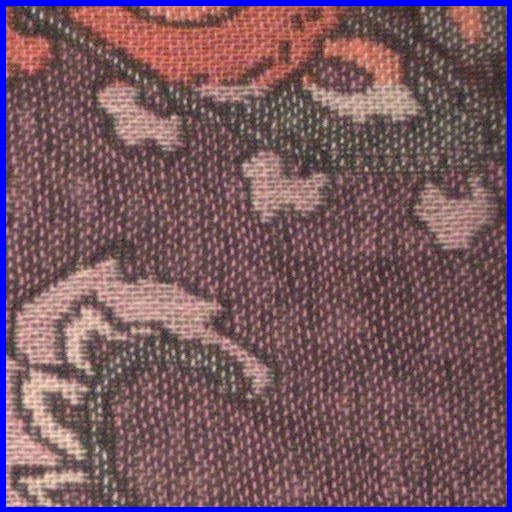} &
    \includegraphics[width=\figwidthsuppnm\columnwidth]{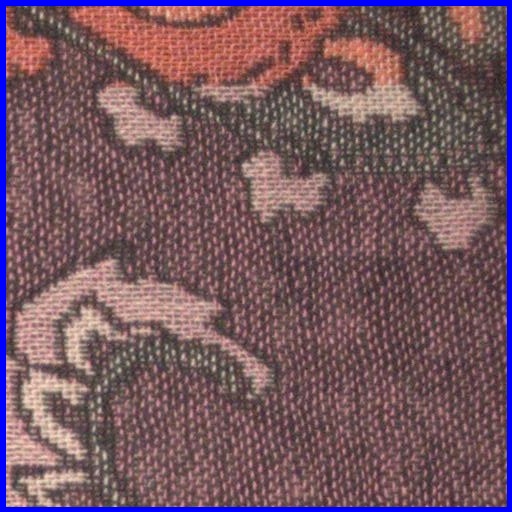} \\
    \includegraphics[width=\figwidthsuppnm\columnwidth]{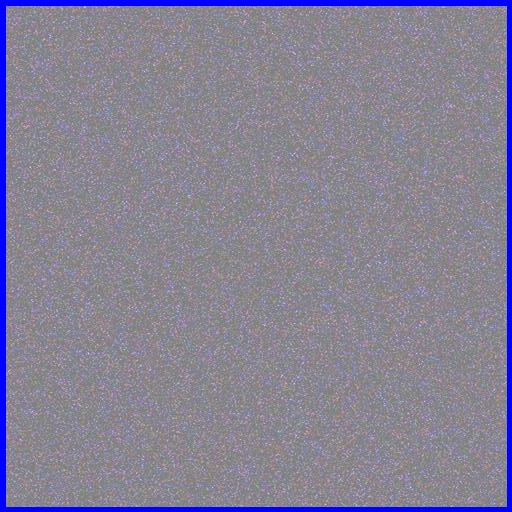} &
    \includegraphics[width=\figwidthsuppnm\columnwidth]{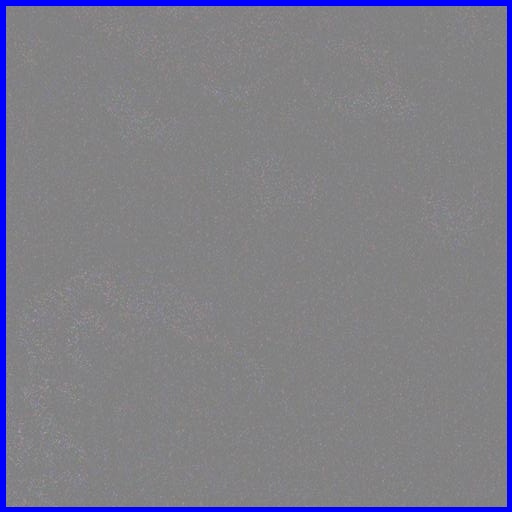} &
    \includegraphics[width=\figwidthsuppnm\columnwidth]{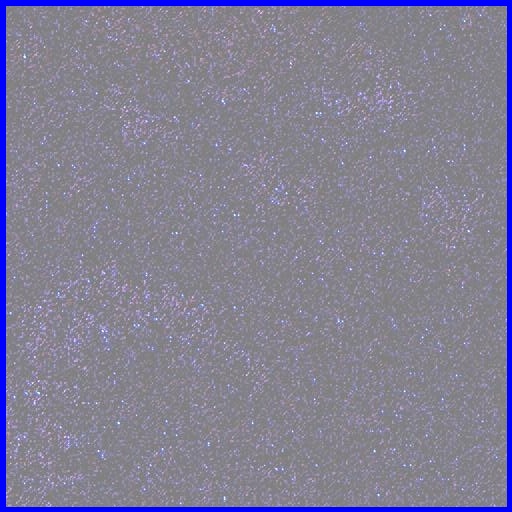} &
    \includegraphics[width=\figwidthsuppnm\columnwidth]{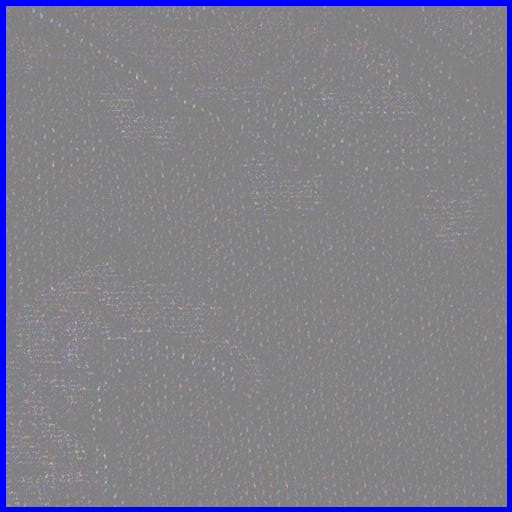} &
    \includegraphics[width=\figwidthsuppnm\columnwidth]{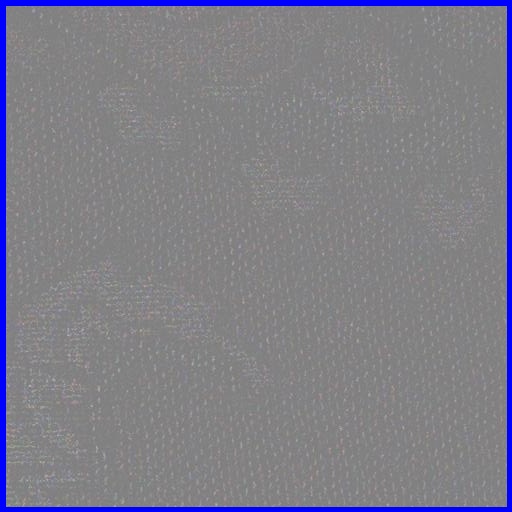} \\
    \includegraphics[width=\figwidthsuppnm\columnwidth]{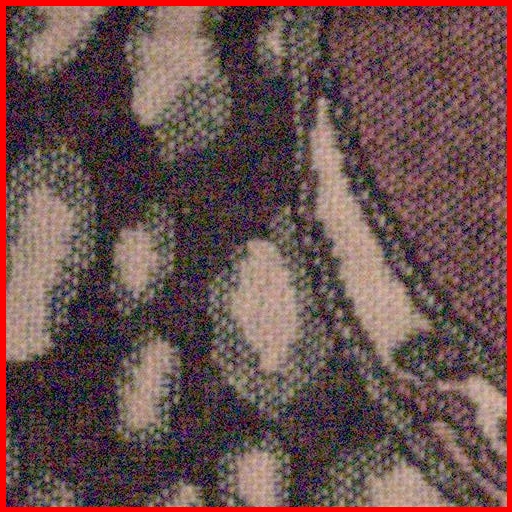} &
    \includegraphics[width=\figwidthsuppnm\columnwidth]{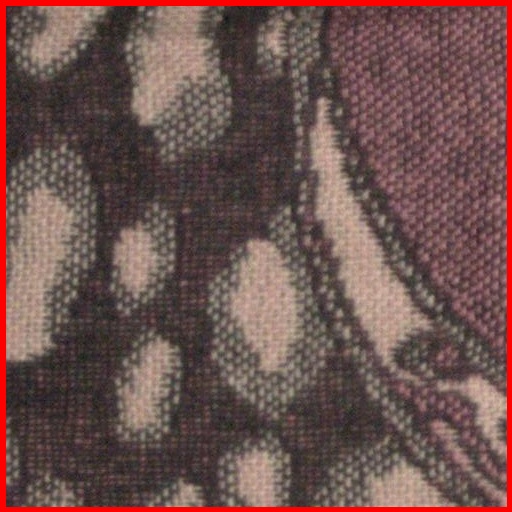} &
    \includegraphics[width=\figwidthsuppnm\columnwidth]{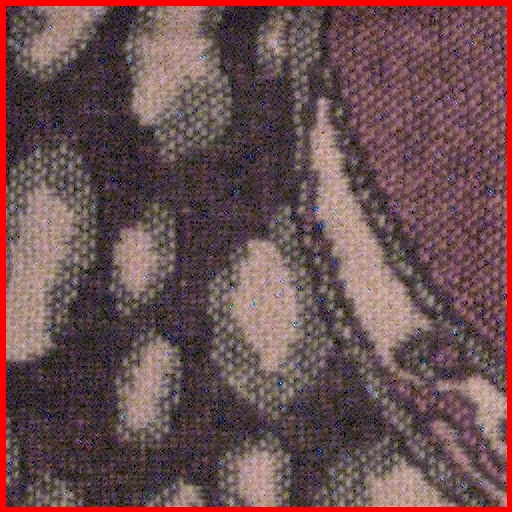} &
    \includegraphics[width=\figwidthsuppnm\columnwidth]{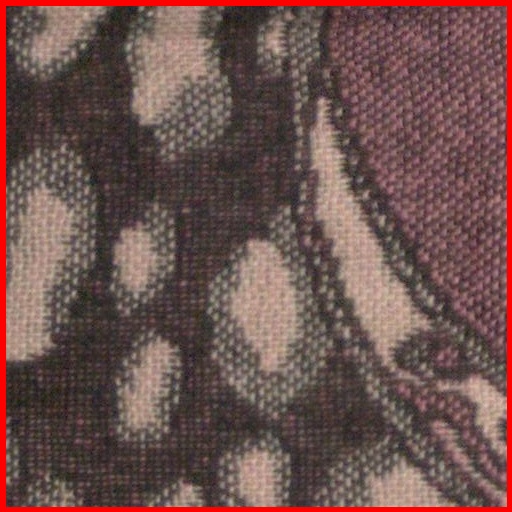} &
    \includegraphics[width=\figwidthsuppnm\columnwidth]{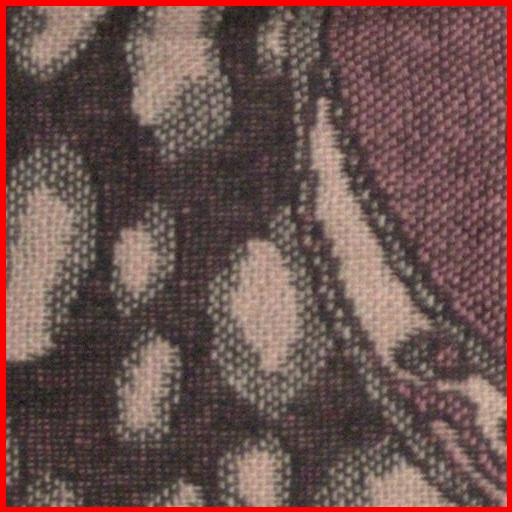} \\
    \includegraphics[width=\figwidthsuppnm\columnwidth]{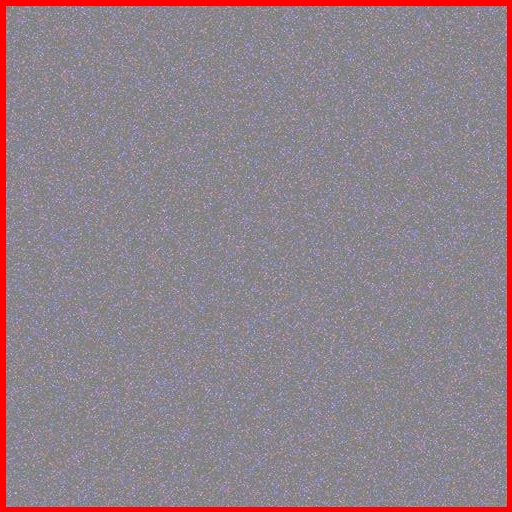} &
    \includegraphics[width=\figwidthsuppnm\columnwidth]{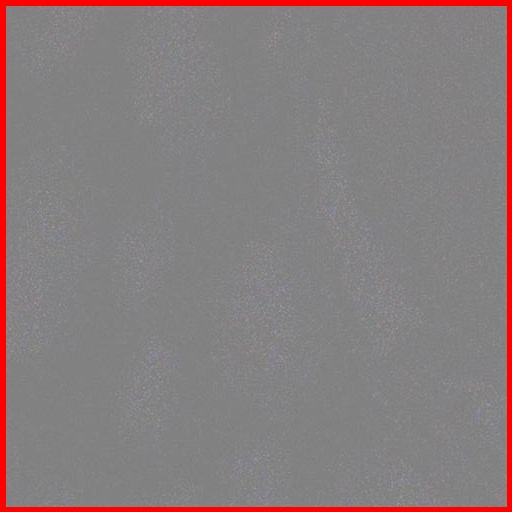} &
    \includegraphics[width=\figwidthsuppnm\columnwidth]{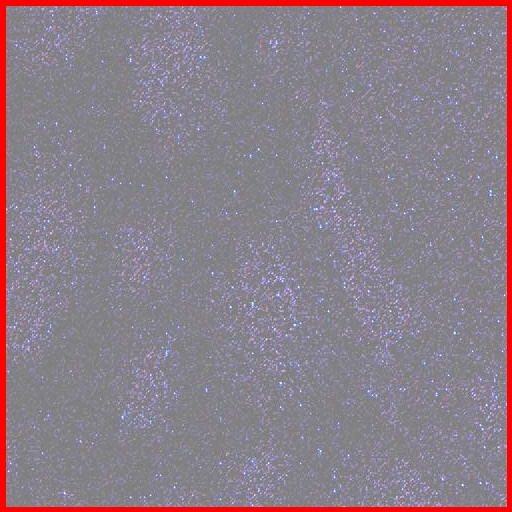} &
    \includegraphics[width=\figwidthsuppnm\columnwidth]{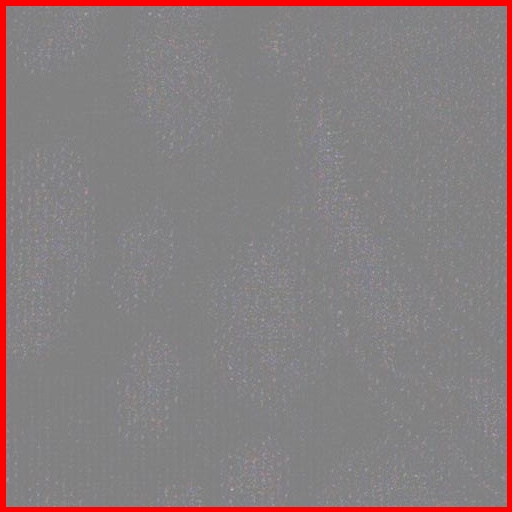} &
    \includegraphics[width=\figwidthsuppnm\columnwidth]{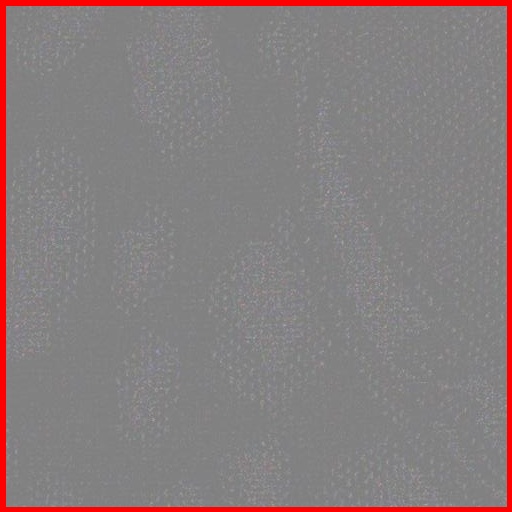} \\
    Gaussian & Poisson-Gauss. & Noise Flow & Ours & Real Noise \\
    \end{tabular}
    \caption{\textbf{More visual results of different noise models (scene 008)}}
    \label{fig:m3}
\end{figure}

\begin{figure}[ht]
    \centering
    \tiny
    \begin{tabular}{cccc}

    \multicolumn{4}{c}{\includegraphics[width=0.82\columnwidth]{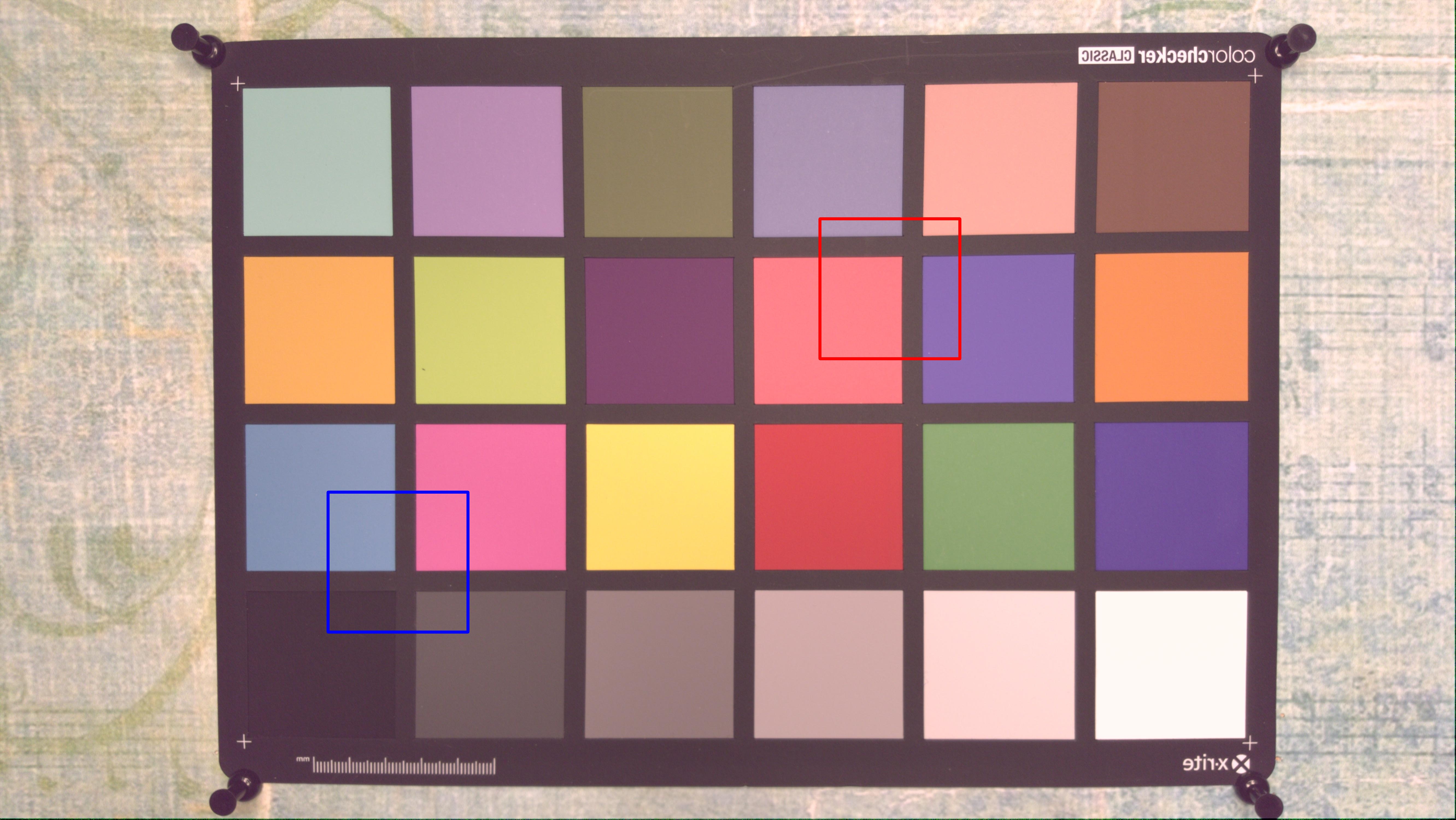}} \\
    \multicolumn{4}{c}{Reference Clean Image} \\
    
    \includegraphics[width=\figwidthsuppnr\columnwidth]{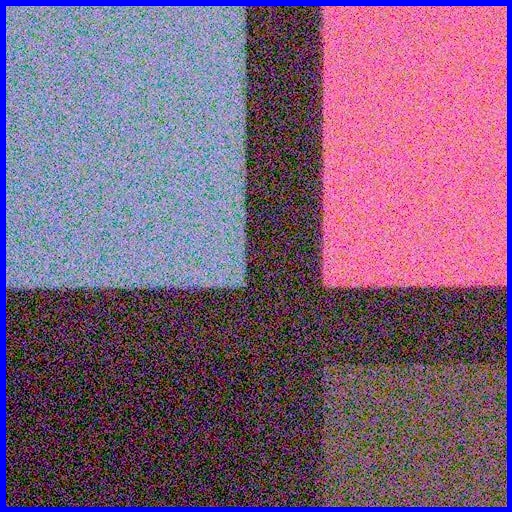} &
    \includegraphics[width=\figwidthsuppnr\columnwidth]{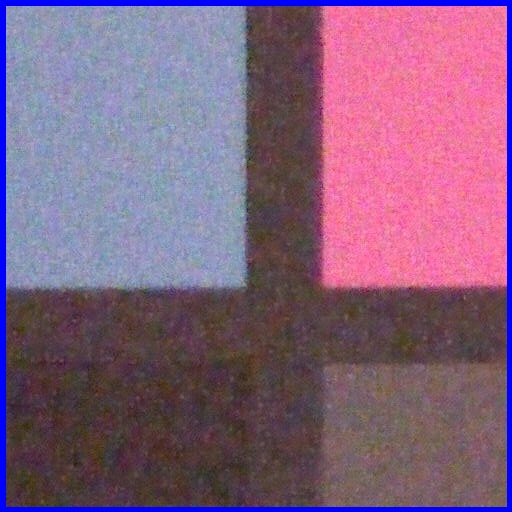} &
    \includegraphics[width=\figwidthsuppnr\columnwidth]{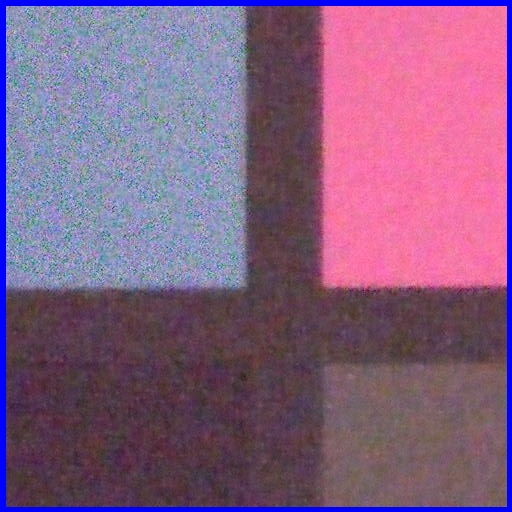} &
    \includegraphics[width=\figwidthsuppnr\columnwidth]{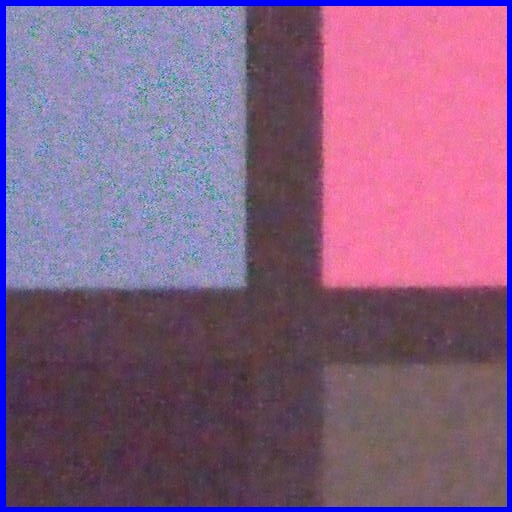} \\
    \includegraphics[width=\figwidthsuppnr\columnwidth]{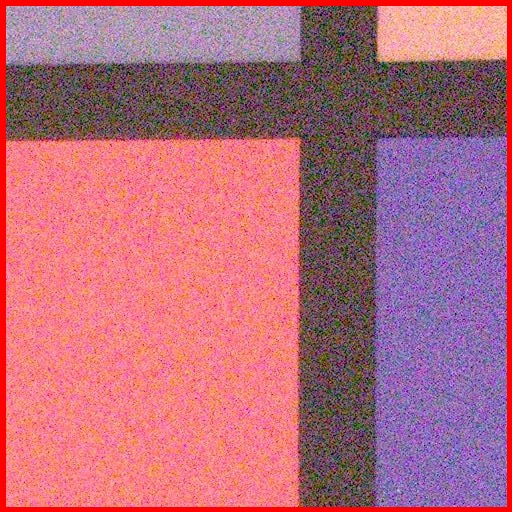} &
    \includegraphics[width=\figwidthsuppnr\columnwidth]{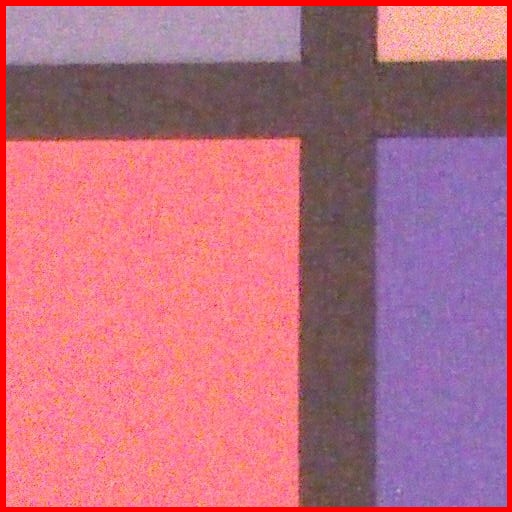} &
    \includegraphics[width=\figwidthsuppnr\columnwidth]{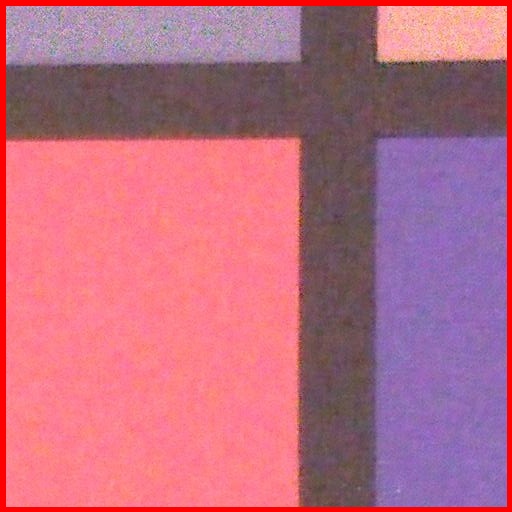} &
    \includegraphics[width=\figwidthsuppnr\columnwidth]{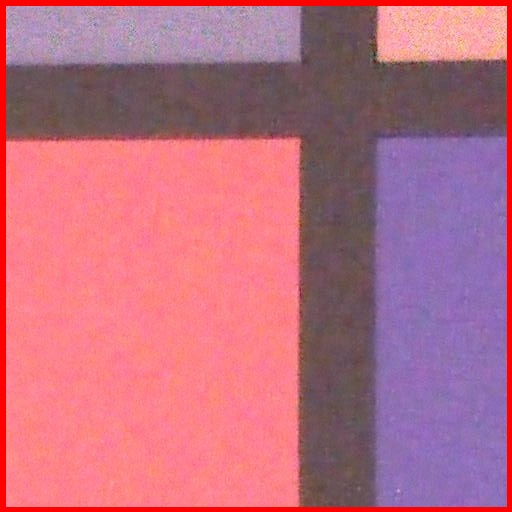} \\
    Noisy & Gaussian & Poisson-Gauss. & Real Only \\
    \includegraphics[width=\figwidthsuppnr\columnwidth]{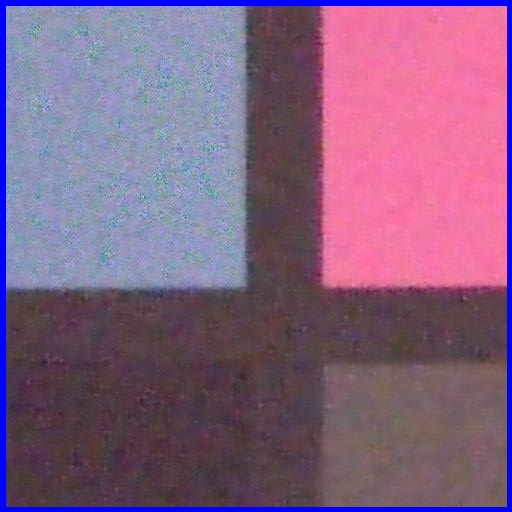} &
    \includegraphics[width=\figwidthsuppnr\columnwidth]{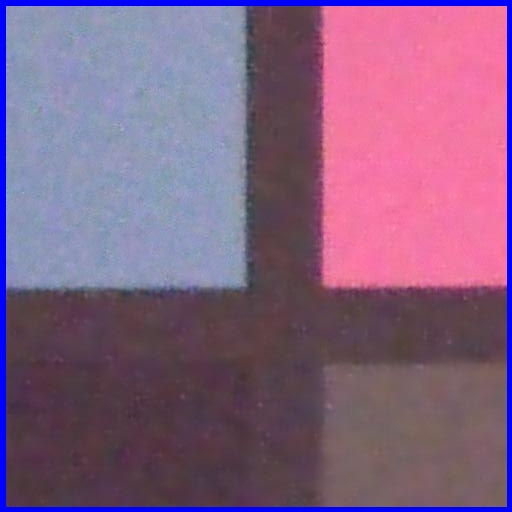} &
    \includegraphics[width=\figwidthsuppnr\columnwidth]{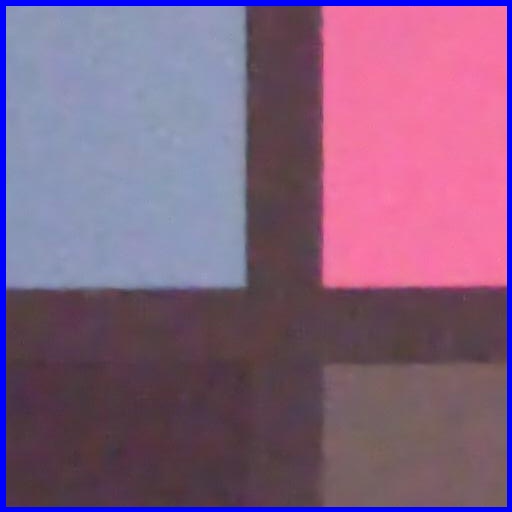} &
    \includegraphics[width=\figwidthsuppnr\columnwidth]{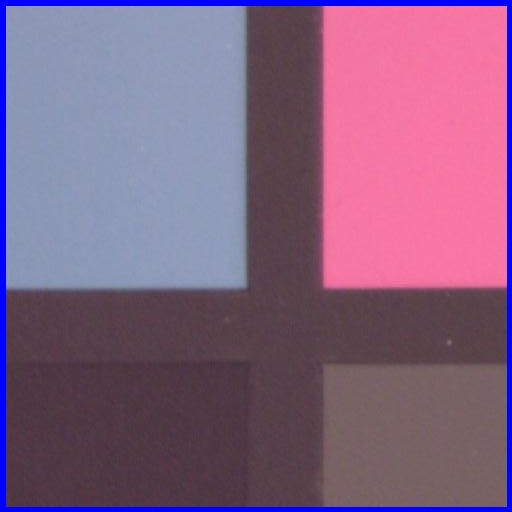} \\
    \includegraphics[width=\figwidthsuppnr\columnwidth]{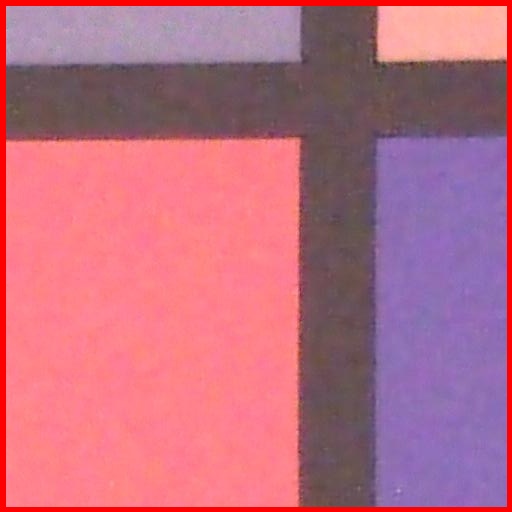} &
    \includegraphics[width=\figwidthsuppnr\columnwidth]{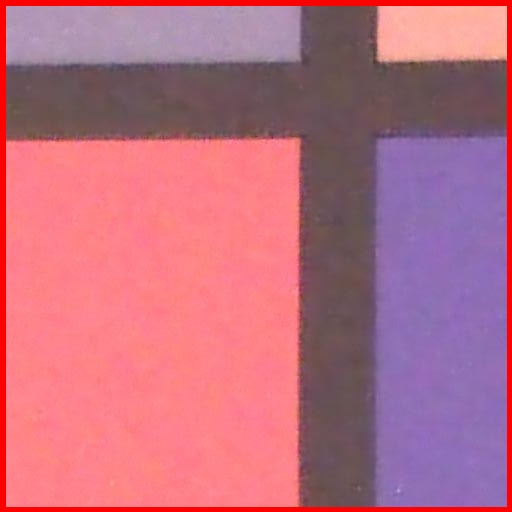} &
    \includegraphics[width=\figwidthsuppnr\columnwidth]{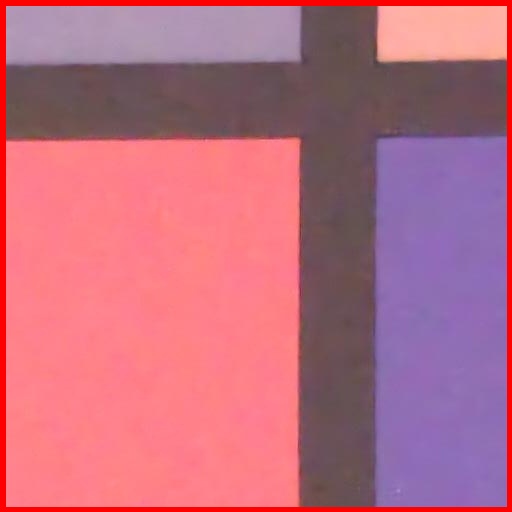} & 
    \includegraphics[width=\figwidthsuppnr\columnwidth]{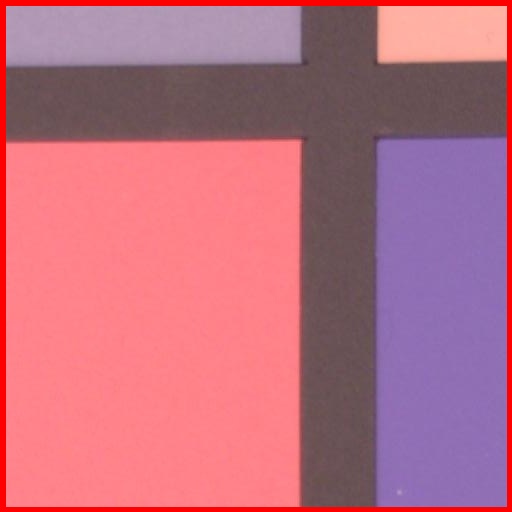} \\
    Noise Flow & Ours & Ours + Real & Clean \\
    \end{tabular}
    \caption{\textbf{More qualitative results of real image denoising (scene 1)}}
    \label{fig:r1}
\end{figure}

\begin{figure}[ht]
    \centering
    \tiny
    \begin{tabular}{cccc}

    \multicolumn{4}{c}{\includegraphics[width=0.82\columnwidth]{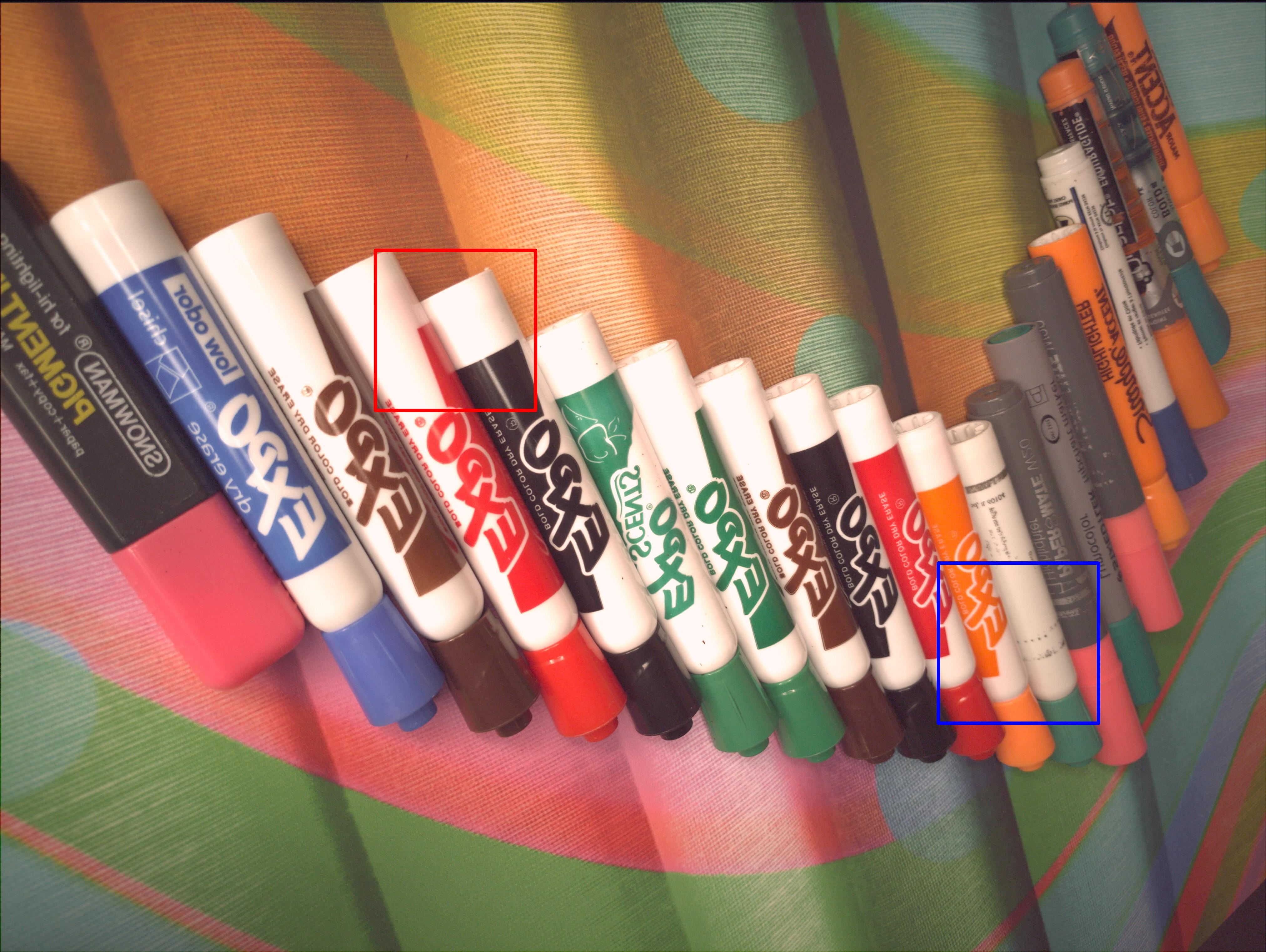}} \\
    \multicolumn{4}{c}{Reference Clean Image} \\
    
    \includegraphics[width=\figwidthsuppnr\columnwidth]{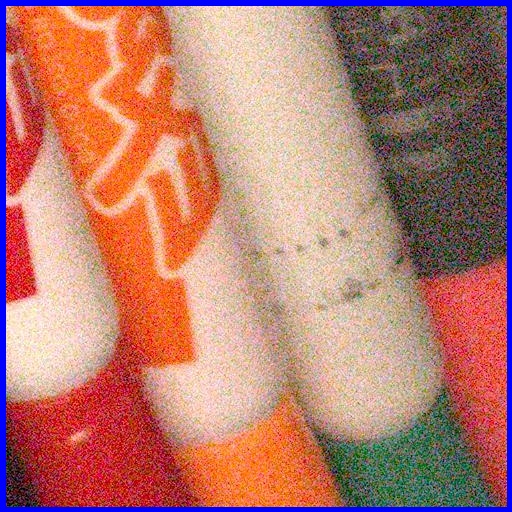} &
    \includegraphics[width=\figwidthsuppnr\columnwidth]{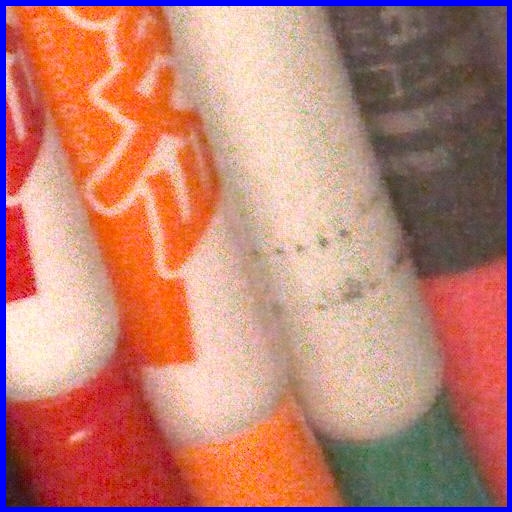} &
    \includegraphics[width=\figwidthsuppnr\columnwidth]{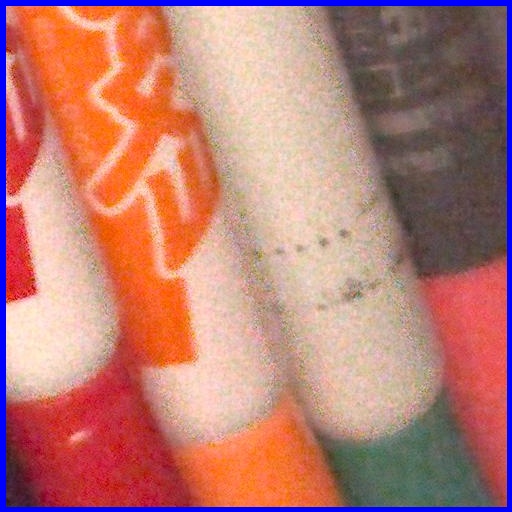} &
    \includegraphics[width=\figwidthsuppnr\columnwidth]{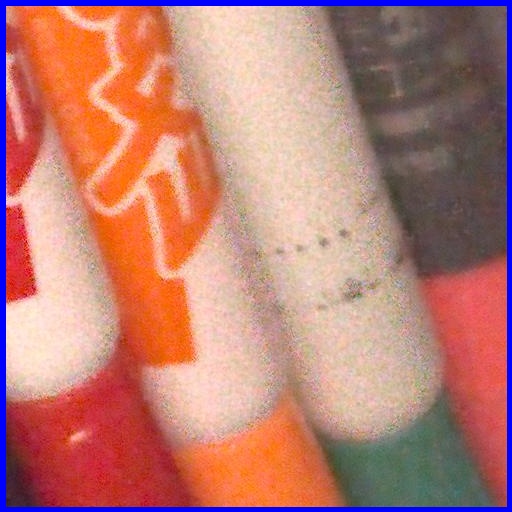} \\
    \includegraphics[width=\figwidthsuppnr\columnwidth]{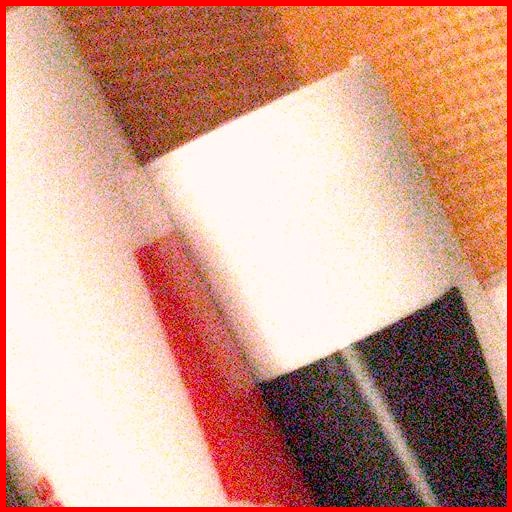} &
    \includegraphics[width=\figwidthsuppnr\columnwidth]{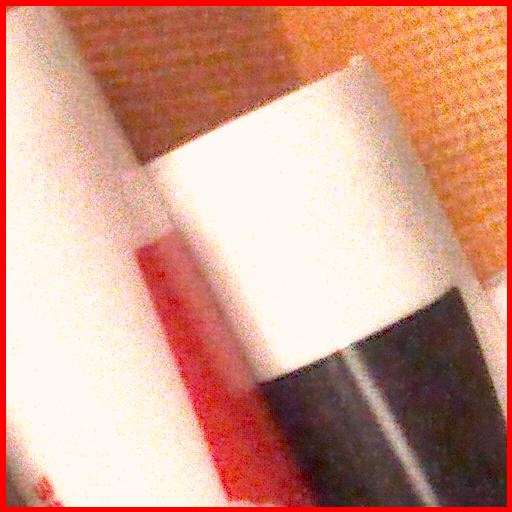} &
    \includegraphics[width=\figwidthsuppnr\columnwidth]{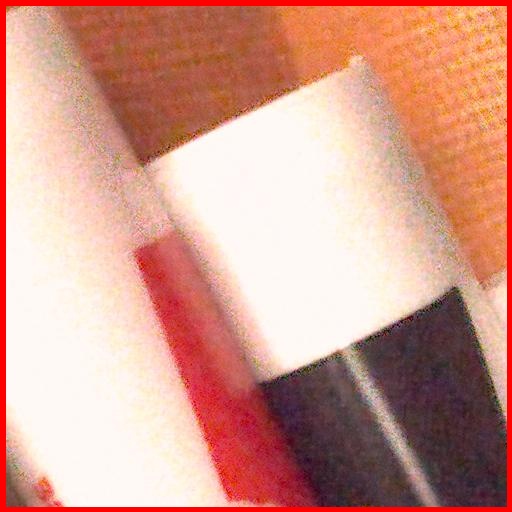} &
    \includegraphics[width=\figwidthsuppnr\columnwidth]{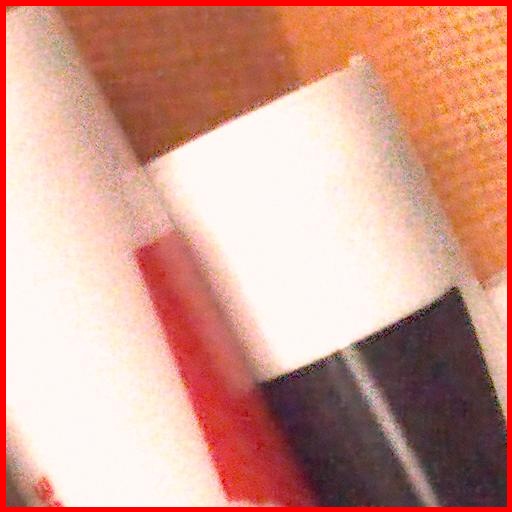} \\
    Noisy & Gaussian & Poisson-Gauss. & Real Only \\
    \includegraphics[width=\figwidthsuppnr\columnwidth]{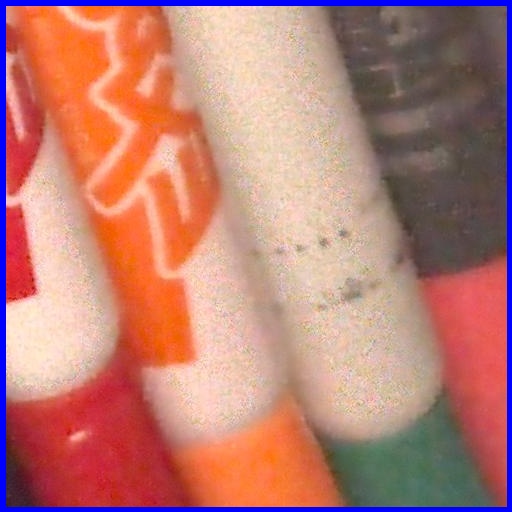} &
    \includegraphics[width=\figwidthsuppnr\columnwidth]{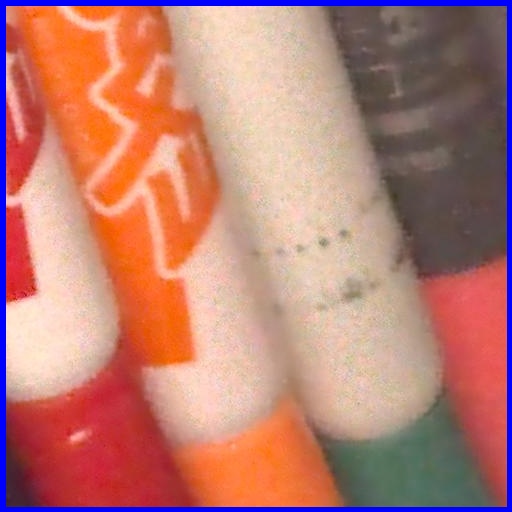} &
    \includegraphics[width=\figwidthsuppnr\columnwidth]{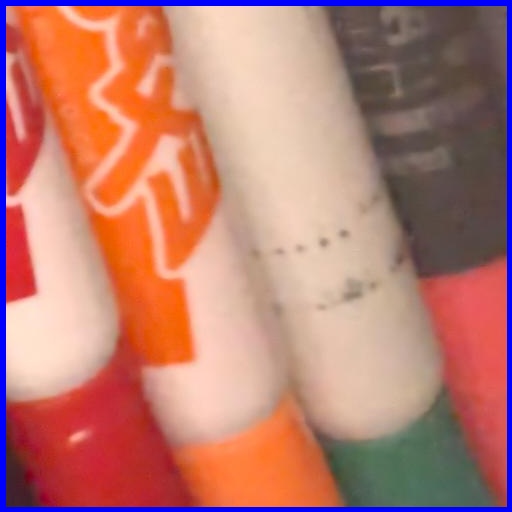} &
    \includegraphics[width=\figwidthsuppnr\columnwidth]{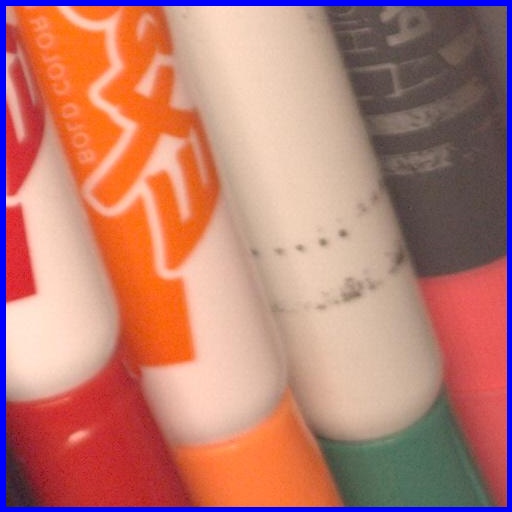} \\
    \includegraphics[width=\figwidthsuppnr\columnwidth]{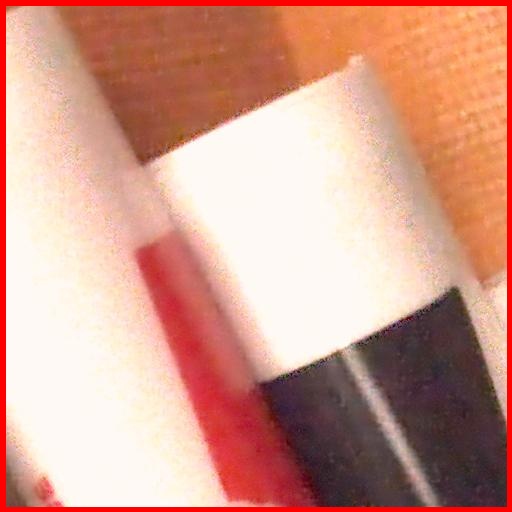} &
    \includegraphics[width=\figwidthsuppnr\columnwidth]{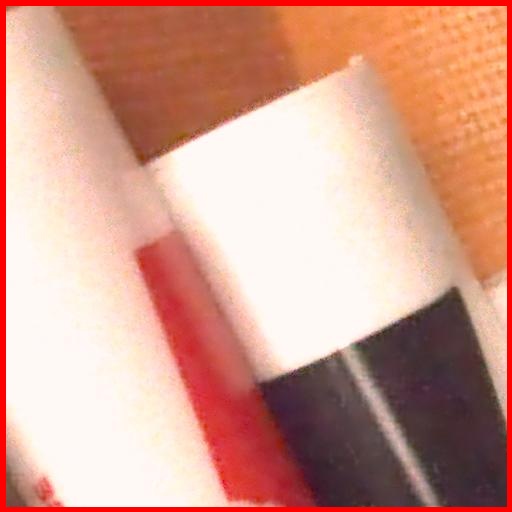} &
    \includegraphics[width=\figwidthsuppnr\columnwidth]{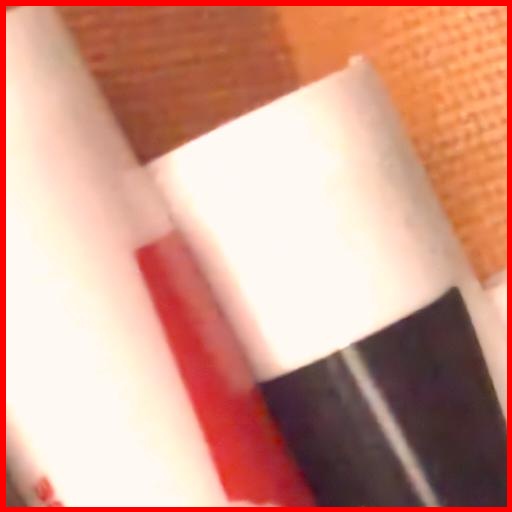} & 
    \includegraphics[width=\figwidthsuppnr\columnwidth]{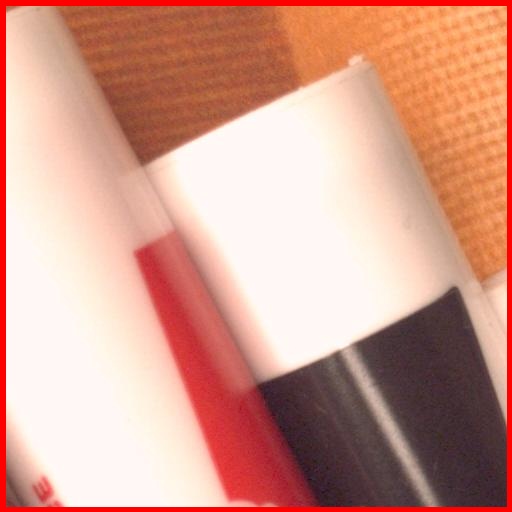} \\
    Noise Flow & Ours & Ours + Real & Clean \\
    \end{tabular}
    \caption{\textbf{More qualitative results of real image denoising (scene 2)}}
    \label{fig:r2}
\end{figure}

\begin{figure}[ht]
    \centering
    \tiny
    \begin{tabular}{cccc}

    \multicolumn{4}{c}{\includegraphics[width=0.82\columnwidth]{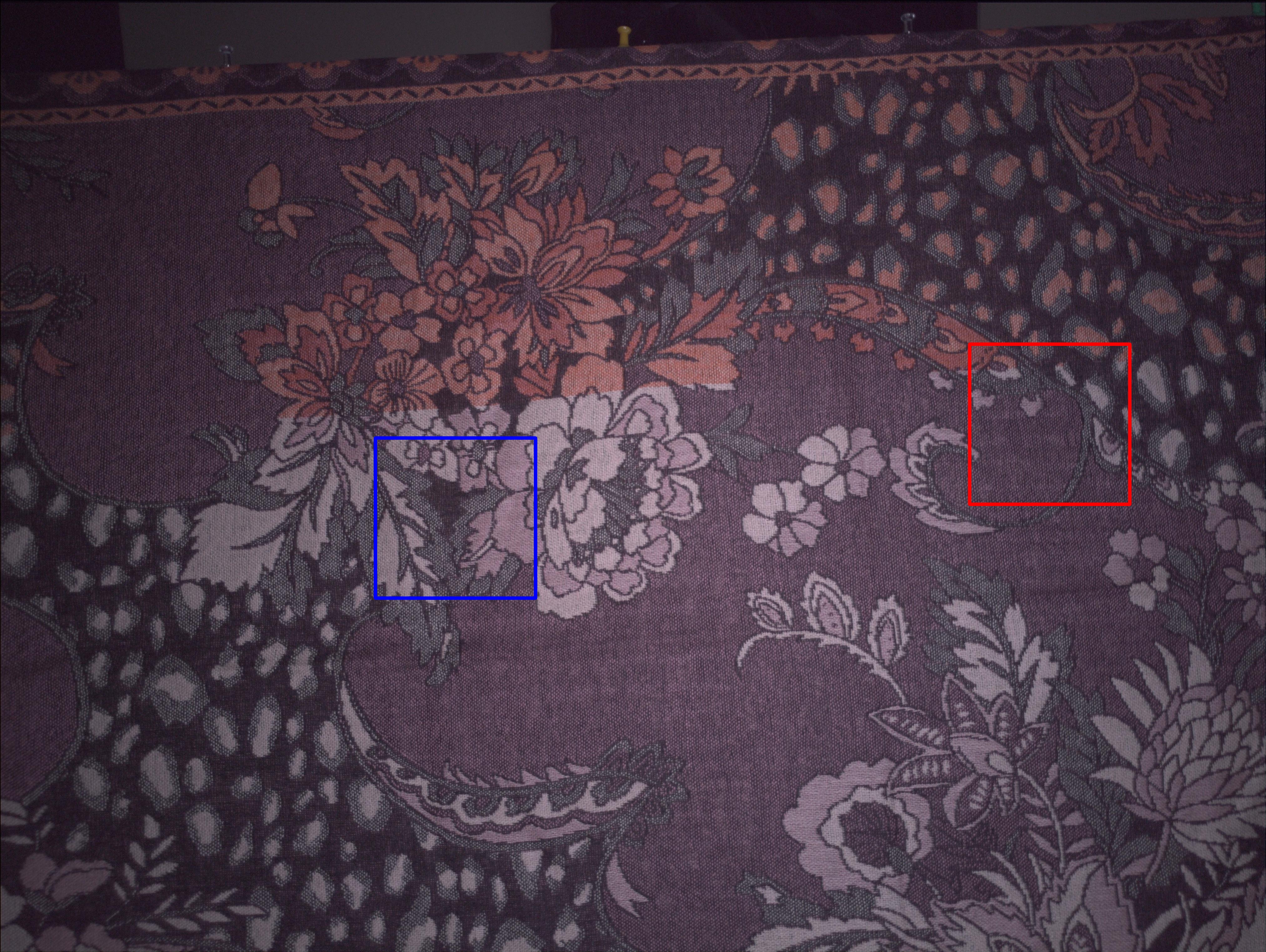}} \\
    \multicolumn{4}{c}{Reference Clean Image} \\
    
    \includegraphics[width=\figwidthsuppnr\columnwidth]{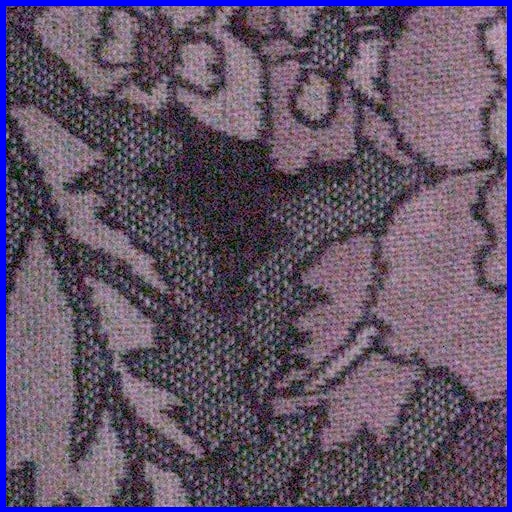} &
    \includegraphics[width=\figwidthsuppnr\columnwidth]{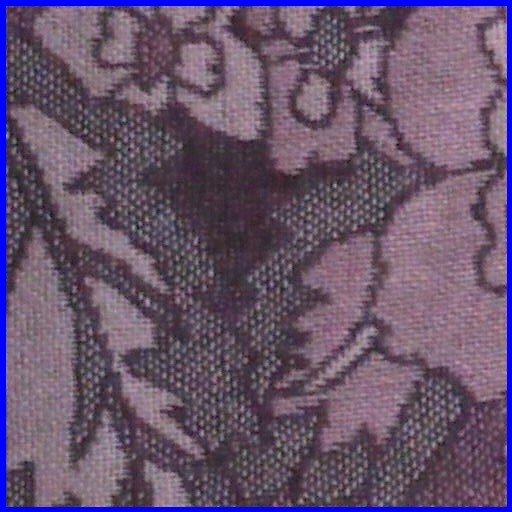} &
    \includegraphics[width=\figwidthsuppnr\columnwidth]{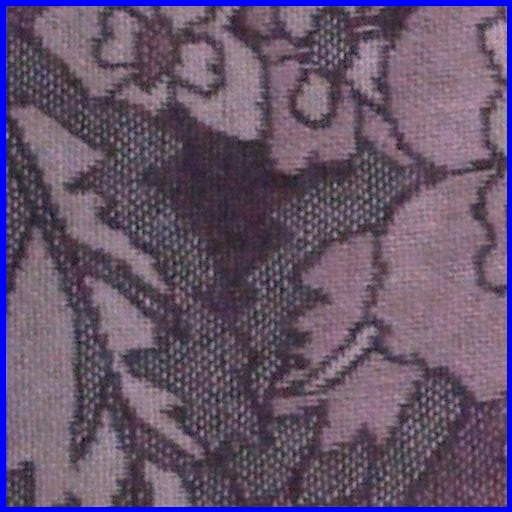} &
    \includegraphics[width=\figwidthsuppnr\columnwidth]{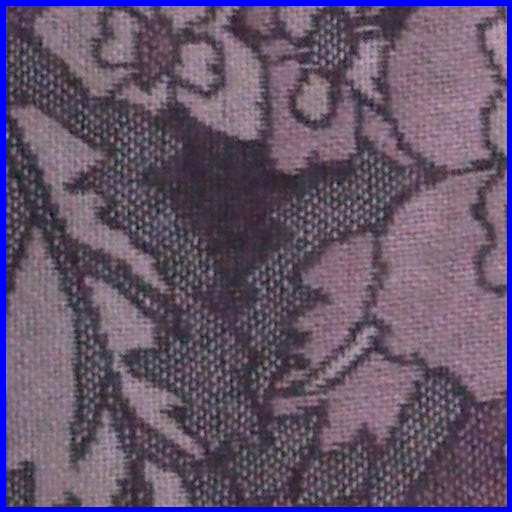} \\
    \includegraphics[width=\figwidthsuppnr\columnwidth]{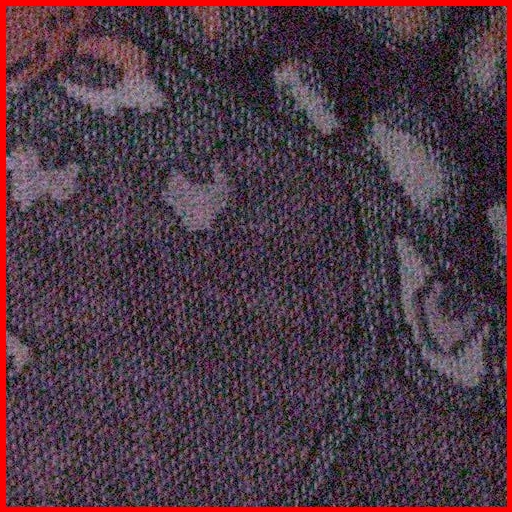} &
    \includegraphics[width=\figwidthsuppnr\columnwidth]{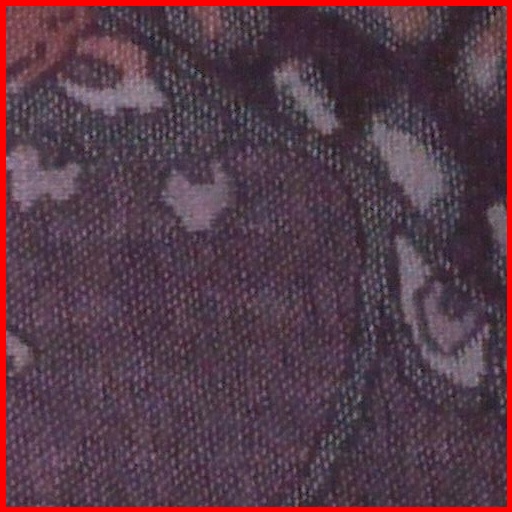} &
    \includegraphics[width=\figwidthsuppnr\columnwidth]{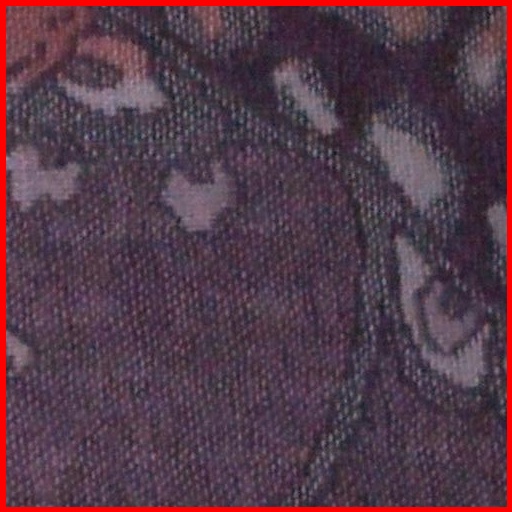} &
    \includegraphics[width=\figwidthsuppnr\columnwidth]{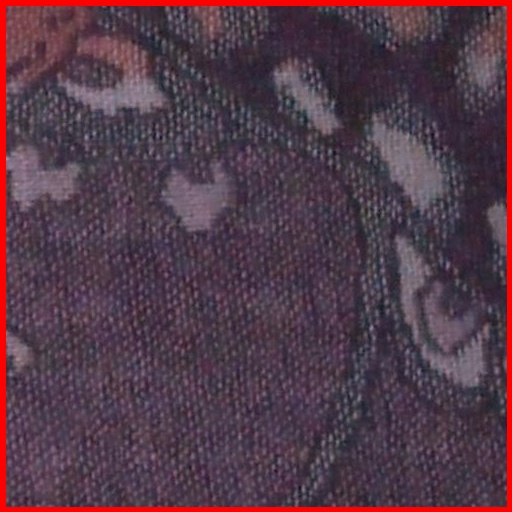} \\
    Noisy & Gaussian & Poisson-Gauss. & Real Only \\
    \includegraphics[width=\figwidthsuppnr\columnwidth]{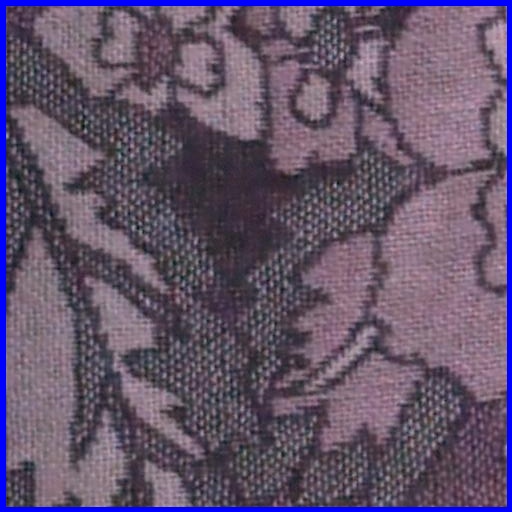} &
    \includegraphics[width=\figwidthsuppnr\columnwidth]{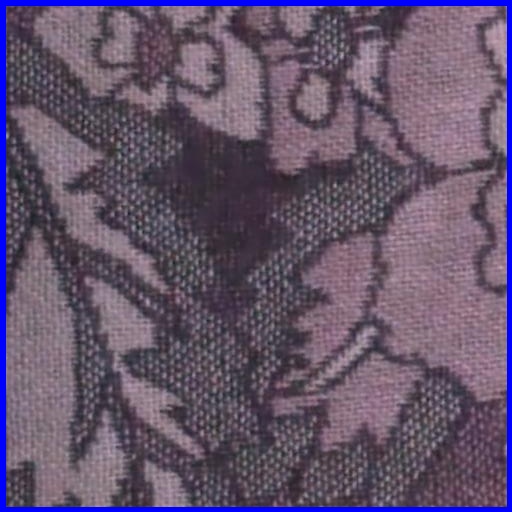} &
    \includegraphics[width=\figwidthsuppnr\columnwidth]{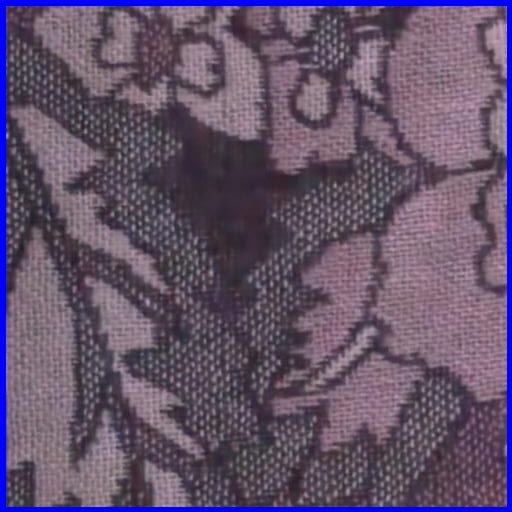} &
    \includegraphics[width=\figwidthsuppnr\columnwidth]{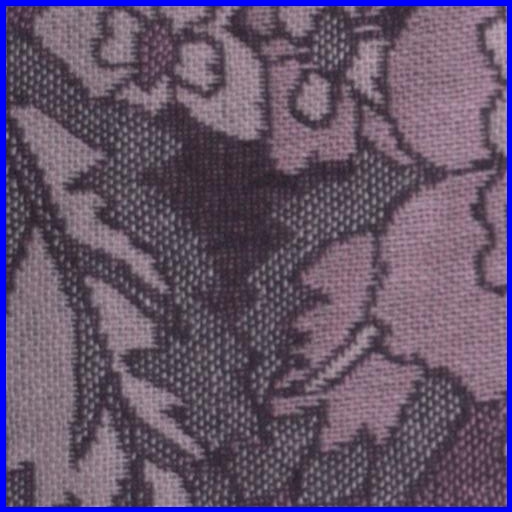} \\
    \includegraphics[width=\figwidthsuppnr\columnwidth]{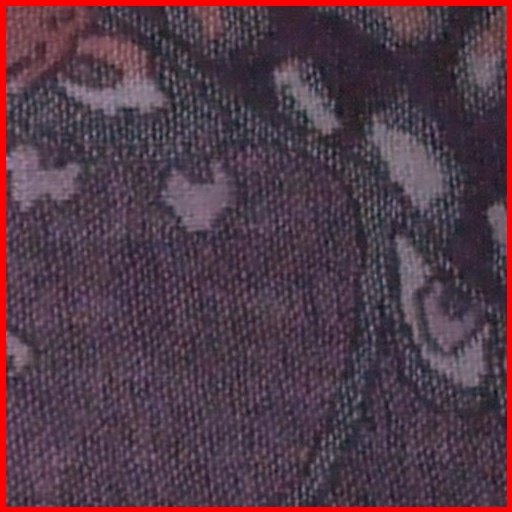} &
    \includegraphics[width=\figwidthsuppnr\columnwidth]{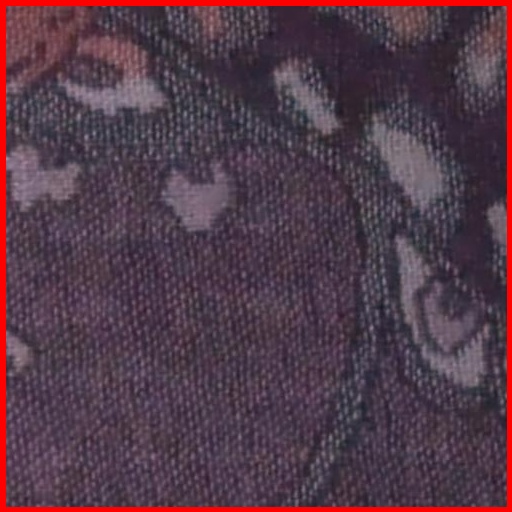} &
    \includegraphics[width=\figwidthsuppnr\columnwidth]{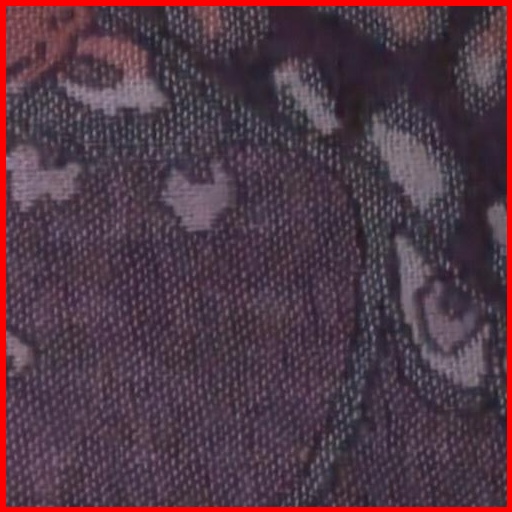} & 
    \includegraphics[width=\figwidthsuppnr\columnwidth]{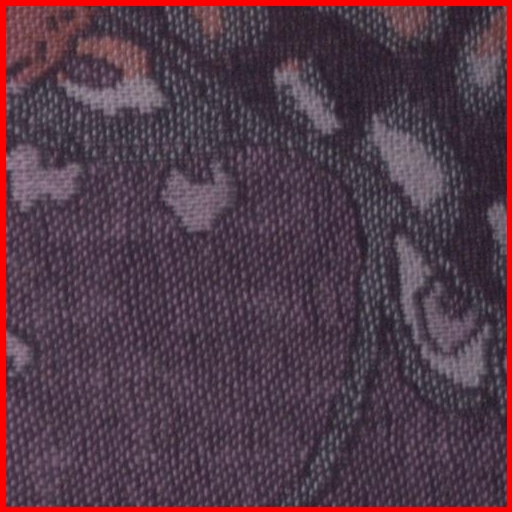} \\
    Noise Flow & Ours & Ours + Real & Clean \\
    \end{tabular}
    \caption{\textbf{More qualitative results of real image denoising (scene 8)}}
    \label{fig:r3}
\end{figure}

More synthesized noise samples as well as the corresponding noisy images are shown in Figs.~\ref{fig:m1}--\ref{fig:m3}.
More qualitative results of real image denoising are shown in Figs.~\ref{fig:r1}--\ref{fig:r3}.

\def\fo{21}
\setlength{\tabcolsep}{1.4pt}

\def\fo{5}
\setlength{\tabcolsep}{1.4pt}

\def\fo{14}
\setlength{\tabcolsep}{1.4pt}

\def\fo{78}
\setlength{\tabcolsep}{1.4pt}

\def\fo{15}
\setlength{\tabcolsep}{1.4pt}

\def\fo{21}
\setlength{\tabcolsep}{1.4pt}

\end{document}